\documentclass{article}
\usepackage[margin=1in]{geometry}
\usepackage{amsmath,amsfonts}
\usepackage{bm,dsfont}
\usepackage{graphicx}
\usepackage{lipsum}
\usepackage{booktabs,multirow}
\usepackage[colorlinks=true,linkcolor=blue,citecolor=blue]{hyperref}
\usepackage{comment}
\usepackage{caption}
\usepackage{subcaption}
\usepackage{array}

\usepackage[sorting=none,maxbibnames=99]{biblatex}
\newcommand{\R}{\mathbb{R}}			% Real Numbers
\newcommand{\set}[1]{\left\lbrace #1 \right\rbrace}

\newcommand{\norm}[1]{\left\Vert #1 \right\Vert}

\title{Assessing the performance of correlation-based multi-fidelity neural emulators}
\author{Cristian J. Villatoro$^{1}$, Gianluca Geraci$^{2}$ and Daniele E. Schiavazzi$^{1,*}$}
\date{}

\begin{document}

\maketitle

\begin{center}
    \begin{minipage}{0.8\linewidth}
        \begin{center}
            \vspace{-20pt}
            $^{1}$ Department of Applied and Computational Mathematics and Statistics\\ University of Notre Dame, Notre Dame, IN, USA\\
            $^{2}$ Sandia National Laboratories, Albuquerque, NM, USA\\
            $^{*}$ Corresponding author: dschiavazzi@nd.edu 
        \end{center}
    \end{minipage}
\end{center}

% Image widths
\def\campaignWidth{0.325}
\def\graphicWidth{0.2425}
\def\graphicTwoDWidth{0.19}

\newcolumntype{P}[1]{>{\raggedright\arraybackslash}p{#1}}

\maketitle

\begin{abstract}

\noindent Outer loop tasks such as optimization, uncertainty quantification or inference can easily become intractable when the underlying \emph{high-fidelity} model is computationally expensive. 
Similarly, data-driven architectures typically require large datasets to perform predictive tasks with sufficient accuracy.
A possible approach to mitigate these challenges is the development of \emph{multi-fidelity} emulators, leveraging potentially biased, inexpensive \emph{low-fidelity} information while correcting and refining predictions using scarce, accurate \emph{high-fidelity} data.
This study investigates the performance of multi-fidelity \emph{neural} emulators, neural networks designed to learn the input-to-output mapping by integrating limited high-fidelity data with abundant low-fidelity model solutions.
We investigate the performance of such emulators for low and high-dimensional functions, with oscillatory character, in the presence of discontinuities, for collections of models with equal and dissimilar parametrization, and for a possibly large number of potentially corrupted low-fidelity sources.
In doing so, we consider a large number of architectural, hyperparameter, and dataset configurations including networks with a different amount of spectral bias (Multi-Layered Perceptron, Siren and Kolmogorov Arnold Network), various mechanisms for coordinate encoding, exact or learnable low-fidelity information, and for varying training dataset size.
We further analyze the added value of the multi-fidelity approach by conducting equivalent single-fidelity tests for each case, quantifying the performance gains achieved through fusing multiple sources of information.
\end{abstract}

% ====================
\section{Introduction}
% ====================

Recent advances in numerical methods and algorithms, combined with the increasing availability of high-performance computing hardware, have substantially improved the realism and accuracy of simulations across a broad range of applications, including cardiovascular modeling~\cite{dinh2019}, wind power plants~\cite{Rybchuk2023}, and climate forecasting~\cite{CHKEIR2023106548}, among many others. 
However, the computational cost of such \emph{high-fidelity} (HF) simulations has also grown considerably, restricting the number of realizations that can be computed. As a result, only limited datasets of HF solutions are typically available for training emulators, which are often used for tasks such as uncertainty quantification and inference.

To address this limitation, \emph{multi-fidelity} (MF) surrogate modeling leverages large datasets from computationally inexpensive \emph{low-fidelity} (LF) approximations together with smaller sets of costly HF data to construct accurate predictive models. By combining information across different fidelity levels, MF approaches reduce computational expense without sacrificing predictive accuracy. A wide range of techniques for multi-fidelity information fusion has been proposed in the literature. Classical approaches include discrepancy-based fusion (additive or multiplicative corrections), co-Kriging or Gaussian process regression, and polynomial chaos expansions. The seminal Kennedy–O’Hagan autoregressive model expresses the HF response as an affine transformation of the LF prediction plus an input-dependent discrepancy term~\cite{kennedy2001bayesian}. Co-Kriging extends Gaussian process (GP) regression to multiple fidelities by treating LF and HF data as jointly Gaussian and estimating their cross-correlation, thereby enriching HF predictions with LF information. Recursive and nested co-Kriging formulations further generalize this idea to fuse multiple fidelity levels for enhanced accuracy~\cite{le2014recursive,perdikaris2017nonlinear}. Polynomial Chaos Expansion (PCE~\cite{xiu2002wiener}) offers a probabilistic surrogate framework by expanding model outputs in orthogonal polynomial bases tailored to the input distribution. MF-PCE variants use LF models to guide the basis expansion and then apply sparse HF corrections~\cite{palar2018global,salehi2018efficient}. Similarly, radial basis function (RBF) interpolation~\cite{song2019radial} and moving-least-squares (MLS) schemes~\cite{wang2021multi} have been extended to incorporate fidelity-dependent correction terms. Other strategies include space-mapping methods that establish transformations between the input spaces of LF and HF models~\cite{RAUL2024102213}, correction-based approaches that learn direct correlations between LF and HF outputs~\cite{conti2023multi}, autoregressive methods that enhance HF predictions using LF data as auxiliary inputs~\cite{doi:10.1137/22M1503956}, and Bayesian formulations designed to predict and quantify the statistics of extreme events~\cite{Gong2022MultiFidelityBE}.

In recent years, deep learning has emerged as a powerful paradigm for multi-fidelity surrogate modeling. Neural networks (NNs) can capture complex nonlinear correlations between LF and HF data through shared architectures, transfer learning, and residual-correction networks~\cite{meng2020composite}. A common training strategy is hierarchical transfer learning, in which a neural network is first trained on abundant LF samples and subsequently fine-tuned using limited HF data~\cite{song2022transfer}. Other architectures explicitly represent fidelity levels through specialized subnetworks that exchange information via shared latent layers. For example, multi-fidelity hierarchical neural processes (MF-HNP) use latent variables to capture correlations across fidelities by learning a shared latent representation, such that, conditioned on these variables, predictions at different fidelities become approximately independent~\cite{wu2022multi}. Beyond standard feedforward networks, neural operator architectures such as the Deep Operator Network (DeepONet)~\cite{lu2021learning} and the Fourier Neural Operator (FNO)~\cite{li2020fourier} have been extended to multi-fidelity learning, where coarse-resolution solutions are incorporated to reduce reliance on expensive fine-scale data~\cite{lu2022multifidelity,howard2023multifidelity}. In addition, physics-informed surrogates embed governing equations directly into the training process. By constraining the network correction using physical laws, these models require significantly fewer data and achieve higher accuracy than unconstrained alternatives~\cite{10.1115}.
A recent review on various approaches for multi-fidelity modeling can be also found in~\cite{fernandez2016review_comp_corr}.

Despite their promise, MF models face several challenges. In many applications, models are parameterized using physically meaningful inputs, which often differ between fidelity levels and lead to reduced linear correlations. In deep learning–based MF frameworks, the ability to capture such nonlinear correlations depends strongly on the expressivity of the chosen neural network architecture. This motivates a systematic investigation into the effectiveness of different network architectures when applied to multi-fidelity frameworks. Furthermore, because discrepancies in input parameterization can degrade correlation between LF and HF responses, see, e.g.,~\cite{Zeng2023,Zeng2025}, coordinate transformations offer a potential solution. By mapping the problem to a common latent or encoded coordinate space, these transformations can enhance correlation, simplify network complexity, and improve predictive accuracy. Importantly, such coordinate encoding strategies are independent of the network architecture and can therefore complement the investigation of different architectural designs~\cite{Villatoro2023}.

In this work, we compare the performance of three neural network architectures—multilayer perceptrons (MLPs), sinusoidal representation networks (SIRENs), and kernel-based approximation networks (KANs)—for constructing accurate multi-fidelity surrogates across a range of test problems. We also examine how coordinate encoding can enhance model performance by simplifying correlations between low- and high-fidelity outputs. To assess the benefits of multi-fidelity learning, we perform systematic comparisons with single-fidelity neural network emulators trained solely on HF data. This analysis quantifies the performance gains achieved by incorporating LF information across varying dataset sizes, network architectures, and problem types. The results demonstrate that multi-fidelity approaches consistently outperform their single-fidelity counterparts, particularly in data-scarce scenarios, although the magnitude of improvement depends on both the problem characteristics and the chosen architecture. The remainder of this paper is organized as follows. Section~\ref{sec:Multi-fidelity surrogates} presents the base MF framework, Section~\ref{sec:Coordinate encoding} introduces coordinate encoding for differently parameterized models, and Section~\ref{sec:Network architectures} describes the neural architectures considered in this study. Sections~\ref{sec:Loss function} and~\ref{sec:Hyperparameter optimization strategy} define the loss functions and hyperparameter optimization strategy, respectively. Section~\ref{sec:results} presents numerical results comparing all architectures with and without coordinate encoding, Section~\ref{sec:Discussion} discusses the findings, and Section~\ref{sec:Conclusion and Future Work} concludes the paper with final remarks and directions for future research.

% ===============
\section{Methods}\label{sec:Methods}
% ===============

\subsection{Multi-fidelity emulators}
\label{sec:Multi-fidelity surrogates}

% Defining MF setting
Consider the situation where we would like to generate an emulator for a computationally expensive HF map $\bm{y}_H : \bm{\mathcal{X}}_H \to \mathcal{Y}$ from a small HF dataset $\{\bm{x}_H^{(j)}, y_H^{(j)} \}_{j=1}^{N_H}$, focusing on the situation where $\mathcal{Y}\subset\mathbb{R}$.
Also assume $n$ additional data sources $\{\bm{x}_{L_i}^{(j)}, y_{L_i}^{(j)}\}_{j=1}^{N_{L_i}}$ to be available, evaluated from comparably cheaper LF functions $y_{L,i} : \bm{\mathcal{X}}_{L,i} \to \mathcal{Y}$, such that for each $i=1,\ldots,n$, the resulting datasets are larger, i.e., $N_{L_i} \gg N_{H}$, and each $y_{L,i}$ is \emph{correlated} to $y_H$.
The dimensionality of $\bm{\mathcal{X}}_{L,i}$ is, in general, different for each $i$ and different from $\bm{\mathcal{X}}_{H}$. 
In addition, we use $\boldsymbol{X}_{H}=\{\bm{x}_{H}^{(j)}\}_{j=1}^{N_{H}}$ and $\boldsymbol{X}_{L,i}=\{\bm{x}_{L,i}^{(j)}\}_{j=1}^{N_{L,i}}$, $i=1,\dots,n$ to indicate the available samples in the HF and LF domain, respectively. 

In this context, discovering the relation between HF and LF model outputs allows us to \emph{fuse information} from multiple datasets, resulting in an enriched combined \emph{multi-fidelity} dataset.
Several approaches for this problem have been proposed in the literature, e.g., ~\cite{kennedy2000predicting,fernandez2016review_comp_corr,perdikaris2017nonlinear_multi_corr_to_non_lin,motamed2020multifidelity,475649};
in this work, we follow the formulation developed in~\cite{meng2020composite}.
Consider, for simplicity, the situation where $\bm{\mathcal{X}}_{L,i}\equiv\bm{\mathcal{X}}_{H}\equiv\bm{\mathcal{X}},\,i=1,\dots,n$. The desired HF surrogate is constructed using LF surrogates $\widehat{y}_{L,i}: \bm{\mathcal{X}} \times \bm{\Theta}_{L,i} \to \mathcal{Y}$ for each $i$, and a \textit{correlation function} $\mathcal{F} : \bm{\mathcal{X}} \times \mathcal{Y}^{n} \times \bm{\Theta}_{H} \to \mathcal{Y}$, expressed as
\begin{equation}
    \label{eq:standard_mf_form}
    \widehat{y}_H (\bm{x};\bm{\theta})
    = \mathcal{F}(\bm{x}, \widehat{y}_{L,1}(\bm{x};\bm{\theta}_{L,1}), \cdots, \widehat{y}_{L,n}(\bm{x};\bm{\theta}_{L,n}); \bm{\theta}_H).
\end{equation}
Here, the network parameters are denoted as $\bm{\theta} = \bigcup_{i} \bm{\theta}_{L_i} \cup \bm{\theta}_{H}$.
Following \cite{meng2020composite}, we further write
\begin{equation}
    \mathcal{F}(\bm{x}, \widehat{y}_{L,1}, \ldots, \widehat{y}_{L,n}; \bm{\theta}_H)
    =\mathcal{F}_{l}\left(\bm{x}, \widehat{y}_{L,1}, \cdots, \widehat{y}_{L,n}; \bm{\theta}_H^{l} \right)
    +
    \mathcal{F}_{nl}\left(\bm{x}, \widehat{y}_{L,1}, \cdots, \widehat{y}_{L,n}; \bm{\theta}_H^{nl} \right),
\end{equation}
and $\bm{\theta}_H = \bm{\theta}_H^{l} \cup \bm{\theta}_H^{nl}$ to denote a decomposition of the low- to high-fidelity map and its network parameters into a linear and a nonlinear component, respectively.
For simplicity, we define $\widehat{y}_{H}^{l} = \mathcal{F}_{l}$ as the linear correlation and $\widehat{y}_{H}^{nl} = \mathcal{F}_{nl}$ as the nonlinear correlation so that high-fidelity prediction is written compactly as $\widehat{y}_H = \widehat{y}_H^{l} + \widehat{y}_H^{nl}$.
For the purposes of this work, we will refer to this network configuration as the ``standard network" and compare its performance against other configurations as discussed in Section~\ref{sec:Coordinate encoding}.

\begin{figure}[ht!]
\begin{subfigure}{0.48\textwidth}
    \centering
    \includegraphics[width=\textwidth]{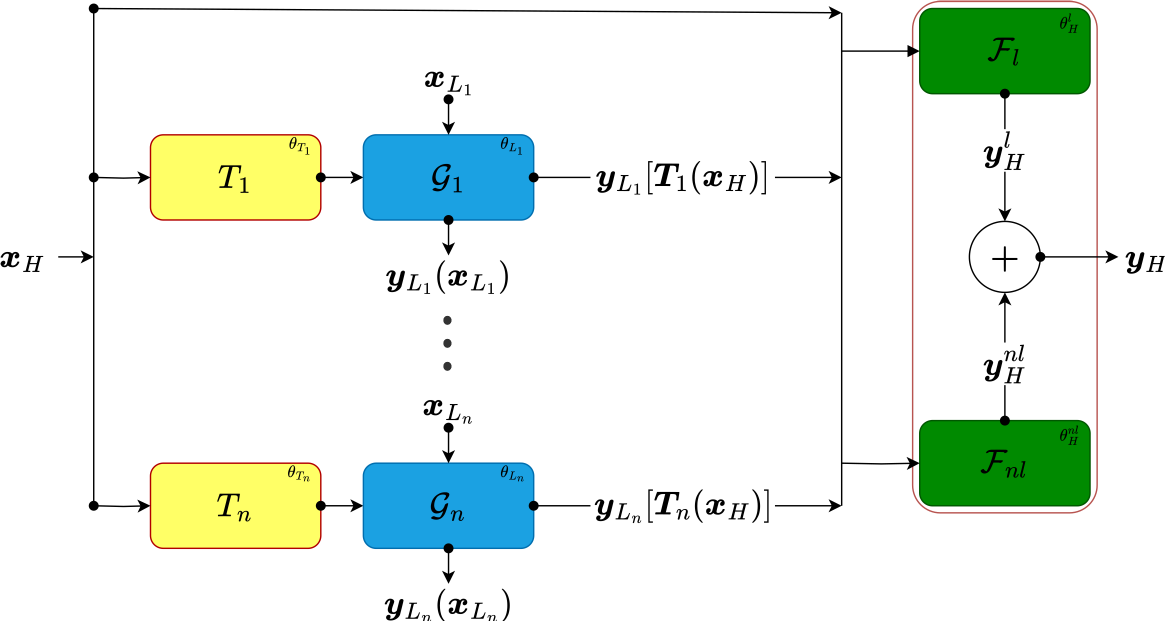}
    \caption{Schematic representation of a standard multi-fidelity neural network with $n$ LF surrogates.}
    \label{fig:architecture_diagram}
\end{subfigure}
\hfill
\begin{subfigure}{0.48\textwidth}
    \centering
    \includegraphics[width=\textwidth]{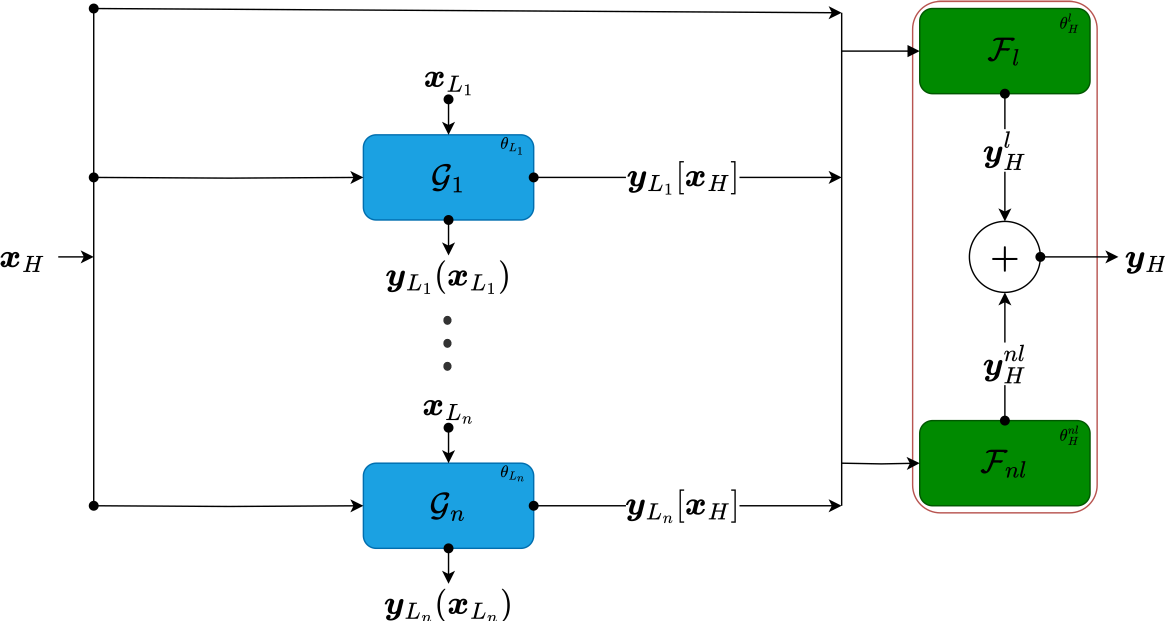}
    \caption{Schematic representation of an \emph{encoded} multi-fidelity neural network with $n$ LF surrogates.}
    \label{fig:architecture_diagram_enc}
\end{subfigure}
\vspace{3pt}
\caption{Schematic of standard and encoded multi-fidelity neural networks.}
\end{figure}

A schematic representation of a standard multi-fidelity network architecture is shown in Figure~\ref{fig:architecture_diagram}.
For this architecture, two training strategies are possible. Either the LF surrogates are separately pre-trained to a desired accuracy and then kept fixed while learning the HF correlation function, or the entire system can be trained end-to-end, learning LF and HF models at the same time.

Without loss of generality, the nonlinear correlation network does not have bias term while the linear correlation network does. Our numerical experiments suggest that this choice speeds up training the HF surrogate and prevents effort being spent on transferring the bias from the linear to the nonlinear network.

% =============================================================
\subsection{Coordinate encoding}\label{sec:Coordinate encoding}
% =============================================================

One of the limitations of the standard architecture is to require a shared parameterization for LF and HF models. Additionally, even under an identical parameterization, a coordinate transformation can be effective in maximizing the linear correlation between LF and HF models, reducing surrogate complexity.
For these reasons, a modification of the network architecture, the so-called \emph{coordinate encoded MF network} (or simply encoded MF network) is introduced which supports dissimilar parameterization between low- and high-fidelity models.

Specifically, we consider the more general case where $\bm{\mathcal{X}}_{L,i} \not\equiv \bm{\mathcal{X}}_{H},\,i=1,\dots,n$ and, for every LF network, we add a \emph{coordinate map} $T_{i} : \bm{\mathcal{X}}_{H} \times \bm{\mathcal{\theta}}_{T,i} \to \bm{\mathcal{X}}_{L,i}$, defining a HF surrogate as
\begin{equation}
    \widehat{y}_H
    = \widehat{y}_{H}^{l}\left(\bm{x}_H, \widehat{y}_{L,1} \circ T_1, \ldots, \widehat{y}_{L,n} \circ T_n; \bm{\theta}_H^{l} \right)
    + \widehat{y}_{H}^{nl}\left(\bm{x}_H, \widehat{y}_{L,1} \circ T_1, \ldots, \widehat{y}_{L,n} \circ T_n; \bm{\theta}_H^{nl} \right),
\end{equation}
with parameters $\bm{\theta}_{enc} = \{\bigcup_{i=1}^{n} (\bm{\theta}_{L_i} \cup \bm{\theta}_{T_i})\} \cup\bm{\theta}_H$.
In this study, the coordinate map $T_i$ represents a general transformation between the HF input space $\bm{\mathcal{X}}_H$ and the $i$-th LF input space $\bm{\mathcal{X}}_{L_i}$, with an implementation which considers a \emph{linear} or \emph{nonlinear} $T_i$, implemented as an MLP without and with hidden layers (plus activations), respectively.
A schematic representation of the proposed architecture is shown in Figure~\ref{fig:architecture_diagram_enc}, which is similar to that proposed in~\cite{meng2020composite} but includes one coordinate transformation per LF.

Coordinate encodings are initialized to an identity, whether the transformation is linear or nonlinear. 
Formally, given a coordinate encoding $T_i$, where $k_{L_i} = \dim(\bm{\mathcal{X}}_{L_i})$, and $k_H = \dim(\bm{\mathcal{X}}_{H})$,
\begin{itemize}
    \item if $k_{L_i} = k_H$, then $T_i$ is initialized as the identify $\mathds{I}_{k_H}\in\R^{k_H\times k_H}$.
    \item if $k_{L_i} \neq k_H$, then $T_i$ is initialized as a binary matrix $\mathds{I}_{k_{L_i}, k_H} \in \R^{k_{L_i} \times k_H}$ where the diagonal entries are 1 and the other entries are 0.
\end{itemize}

For nonlinear encoding, the identity initialization is achieved by modifying the standard Xavier-initialized MLP~\cite{glorot2010understanding}. In particular, all layers but the last are Xavier initialized and the last layer's weights are initialized to 0, ensuring the network initially represents the identity mapping.
We found this initialization improved performance as the encoding networks are initialized as an identity map, while still allowing flexible adaptation during training.

% =================================================================
\subsection{Network architectures}\label{sec:Network architectures}
% =================================================================

We compare the performance of multiple types of neural networks from the literature (i.e., MLP, Siren and KAN) as candidates for multi-fidelity emulators. In addition, we test the performance of each of these networks with and without coordinate encoding. We further discuss their setup and initialization strategy in Subsections~\ref{sec:MLP}, \ref{sec:Siren}, and \ref{sec:KAN}.

% =================================================================
\subsubsection{Multilayer perceptron (MLP) networks}\label{sec:MLP}
% =================================================================

We use MLPs with hyperbolic tangent activation and Xavier initialization as our architectural baseline. 
For the sake of completeness, we briefly discuss an $L$ layer MLP with hidden layers of size $[n_1, \cdots, n_{L-1}]$ as a function $\text{MLP}: \R^{n_0}\to \R^{n_L}$ of the form
\[
\begin{cases}
    \operatorname{MLP}(x)
    =
    \left(
    F_{L-1} \circ (\sigma \circ F_{L-2}) \circ \cdots \circ (\sigma \circ F_{0})
    \right) (x)
    \\
    F_i(x) = W_i x + b_i, & i = 0, \cdots, L-1
\end{cases}
\]
where $\sigma$ is a nonlinear activation function, and $F_i$ is the $i$-th linear layer with weights and biases equal to $W_i \in \R^{n_{i+1}\times n_i}$ and $b_i \in \R^{n_{i+1}}$, respectively.

% =============================================
\subsubsection{Siren networks}\label{sec:Siren}
% =============================================

A Siren network combines a feed forward neural network architecture with sinusoidal activation~\cite{sitzmann2020implicitneuralrepresentationsperiodic} and a careful initialization strategy.
This interesting choice of activation is such that the derivative of a Siren network is also a Siren network, leading to improved accuracy when reconstructing signal/images from their spatial gradient, when reconstructing geometries through their distance function representation (i.e., equation-informed reconstruction based on the Eikonal equation), or when using PINN-based loss augmentation~\cite{karniadakis2021physics} for the solution of PDEs~\cite{sitzmann2020implicitneuralrepresentationsperiodic}.
A Siren network with $L$ layers is written as
\[
\begin{cases}
 \operatorname{Siren}(x) = W_{L-1} (\phi_{L-2} \circ \cdots \circ \phi_{0})(x) + b_{L-1},
 \\
 \phi_0(x) = \sin(\omega_0 \cdot W_{0}\,x + b_{0}),
\\
\phi_l(x) = \sin(W_{l} x + b_{l}),\,\,\text{for}\,\,l = 1, \cdots, L-2
\end{cases}
\]
where $\omega_0$ is a fixed frequency factor, and $W_l \in \R^{n_{l+1}\times n_l}$ and $b_l \in \R^{n_{l+1}}$ are the weight matrix and bias vector for the $l$-th layer, respectively.
Note that we found discrepancies between the official Siren Python implementation's initialization of the weights of their network~\cite{OfficialSirenImplementation2020} and the initialization scheme described in~\cite{sitzmann2020implicitneuralrepresentationsperiodic}.
Since the network is sensitive to how it is initialized, we feel comfortable using a different, more streamlined implementation of a Siren network that also initializes the network weights differently~\cite{LucidrainsSirenImplementation2023}.
In particular, we set $c = 6$, $\omega_0 = 30$, $W_0, b_0 \sim \mathcal{U}\left(-1/n_{l} , 1/n_{l} \right)$ and $W_l, b_l \sim \mathcal{U}\left( - \sqrt{c/\omega_0^2 n_{l}} , \sqrt{c/\omega_0^2 n_{l}} \right)$ for $l=1,\cdots,L-1$. After numerical experiments testing $\omega_0 \in \{5, 10, 20\}$, we found no meaningful difference in predictive performance as $\omega_0$ varied for the test problems we investigated.

% =============================================================
\subsubsection{Kolmogorov-Arnold Networks (KAN)}\label{sec:KAN}
% =============================================================

A Kolmogorov-Arnold Network, or KAN, is a feedforward neural network architecture that leverages the Kolmogorov–Arnold representation theorem~\cite{liu2024kankolmogorovarnoldnetworks}.
The theorem states that any real multi-variate continuous function $f$ can be written as a superposition of a finite number of univariate continuous functions.
Although there are many examples of functions whose exact Kolmogorov–Arnold representations involves non-smooth and possibly fractal univariate functions, the rationale for using this representation lies in the possibility of stacking Kolmogorov–Arnold representations in successive layers (KAN layers, similar to layers in an MLP), which may provide accurate representations from smoother univariate functions.

In terms of implementation, we use the representation used in~\cite{liu2024kankolmogorovarnoldnetworks}, but many other formulations for KAN are discussed in the literature, e.g., ~\cite{347e3b2c-ef29-31f0-80ae-455ba2685cfa,bams/1183527229,doi:10.1080/00029890.1962.11989915,SCHMIDTHIEBER2021119}. 
When approximating a function $f : \R^{n_0} \to \R^{n_L}$, a KAN with $L$ layers and hidden layer size $[n_1, \cdots, n_{L-1}]$, can be expressed in the form
\[
\text{KAN}(x) = (\Phi_{L-1} \circ \Phi_{L-2} \circ \cdots \circ \Phi_{0})(x),
\]
such that for each $\Phi_l(x) = \Phi_l(x_1, \cdots, x_{n_l}) = [\phi_{l,i,j}(x_j)]\in \R^{n_{l+1}, n_{l}},$ for $l = 0, 1, \cdots, L-1$, where each \emph{basis} $\phi_{l,i,j}$ represents a univariate function.
In general, these basis functions can take many forms, e.g. Chebyshev polynominals~\cite{ss2024chebyshevpolynomialbasedkolmogorovarnoldnetworks}, wavelets~\cite{bozorgasl2024wavkanwaveletkolmogorovarnoldnetworks}, radially symmetric functions \cite{li2024kolmogorovarnoldnetworksradialbasis}, and others.

In this work, we use an implementation of the KAN network~\cite{EfficientKAN2024}, designed to improve efficiency with respect to the version proposed in the original paper~\cite{liu2024kankolmogorovarnoldnetworks}.
Specifically, the implementation we use contains the following changes. First, the $\ell_{1}$ norm for regularization is applied to the network weights as opposed to the activation responses, as originally proposed in~\cite{liu2024kankolmogorovarnoldnetworks}.
Second, the univariate functions are simplified by removing the addition scaling applied to each spline, as this scaling can be absorbed into the B-spline weights.
Finally, activations, basis function scaling coefficients and B-splines weights, are initialized using Kaiming Initialization rather than the original scheme.
Finally, the form of the individual univariate functions, $\phi$, is selected as described in~\cite{liu2024kankolmogorovarnoldnetworks} such that
\[
\phi(x) = w_b\,\,\text{SiLU}(x) + \operatorname{spline}(x),\,\,\text{and}\,\,\text{spline}(x) = \sum_{k} c_k B_{k}(x),
\]
where $\operatorname{SiLU}$ is the  Sigmoid Linear Unit function and $B_{k}$ are learned B-splines.

% =================================================
\subsection{Loss function}\label{sec:Loss function}
% =================================================

% Define individual loss components
The numerical results presented in Section~\ref{sec:results} are generated using a loss function containing the terms
\[
  \begin{split}
    \mathcal{L}_{\text {Err}} &=
    \norm{ \widehat{y}_H-y_H }_2^2 + \sum_{i=1}^{n} \norm{ \widehat{y}_{L_i}-y_{L_i} }_2^2,\\
    \mathcal{L}_{\text {Reg}} &= \lambda_H^{n l} \norm{ \boldsymbol{\theta}_H^{nl} }_2 + \lambda_L \sum_{i=1}^{n} \norm{ \boldsymbol{\theta}_{L_i} }_2,\\
    % \item $\mathcal{L}_{\text {Spar}} =
    % \alpha_H^{nl} \norm{ \theta_H^{nl}}_1 + \alpha_{L} \sum_{i=1}^{n} \norm{\theta_{L_i}}_1$.
    \mathcal{L}_{\text {Enc}} &=
    \lambda_T \norm{ \widehat{y}_H^{l} - \widehat{y}_H }_2^2 + \lambda_D \sum_{i=1}^{n} \operatorname{IS}\left(T_{i}\left(\boldsymbol{X}_{H}\right) ; \boldsymbol{X}_{L_{i}}\right),
    \end{split}
\]
where $\mathcal{L}_{\mathrm{Err}}$ measures both the HF and LF mean-squared errors (MSEs) incurred from fitting the surrogates to data, $\mathcal{L}_{\mathrm{Reg}}$ measures the $\ell_{2}$ regularization loss from the HF nonlinear correlation sub-network and from the LF surrogates, whereas $\mathcal{L}_{\text {Enc}}$ measures the losses directly resulting from the selected coordinate encoding.
Specifically, it quantifies the sum of the discrepancy between the HF prediction and the HF linear correlation, thus helping to maximize the amount of linear correlation induced by coordinate encoding, and limits the amount of extrapolation produced by such encoding. 
Here, the interval score ($\text{IS}$) term measures the average $\ell_1$ distance of a point cloud $S = \set{s^{(j)}}_{j=1}^{m} \subset \mathbb{R}^{p}$ from a hyper-rectangle $\mathcal{R} = \prod_{i=1}^{p} [a_i,b_i]$, and it is defined as~\cite{gneiting2007strictly}
\[
\text{IS}\left( S; \mathcal{R} \right) := \frac{1}{m}\sum_{j=1}^{m}\text{d}_{1}(s^{(j)}, \mathcal{R})=\frac{1}{m}\sum_{j=1}^{m}\sum_{i=1}^{p}\left(\text{ReLU}(a_i - s^{(j)}) + \text{ReLU}(s^{(j)} - b_i)\right).
\]
% Particular loss functions
Therefore we consider the loss $\displaystyle \inf _{\boldsymbol{\theta}_L, \boldsymbol{\theta}_H} \mathcal{L}_{\mathrm{Err}} + \mathcal{L}_{\mathrm{Reg}}$ when learning $(\widehat{y}_{L,1}, \ldots, \widehat{y}_{L,n}, \mathcal{F})$ using a standard MF architecture and the loss $\displaystyle \inf _{\boldsymbol{\theta}_L, \boldsymbol{\theta}_H, \boldsymbol{\theta}_T} \mathcal{L}_{\mathrm{Err}} + \mathcal{L}_{\mathrm{Reg}} + \mathcal{L}_{\mathrm{Enc}}$ when instead learning $(\widehat{y}_{L,1}, \ldots, \widehat{y}_{L,n}, T_{1}, \ldots, T_{n}, \mathcal{F})$ from an encoded MF architecture.

% =======================================================
\subsection{Numerical results setup}\label{sec:exp_setup}
% =======================================================

The numerical results in Section~\ref{sec:results} report normalized HF MSEs across 27 network configurations, obtained by combining three levels of network complexities, three sizes HF dataset, and three types of coordinate encoding.
For each case, the test MSEs from the 4 independent realizations are normalized using the squared $\ell_2$ norm of each HF function over its corresponding input domain.
This normalization allows us to compare the performance of different network configurations across test cases of different scales.
From these repetitions, we report both the mean normalized MSE and its coefficient of variation (CoV).
Each configuration has its results averaged over four independent repetitions, using independent training and testing sets to ensure reproducibility.

All network architectures used include at most three components: the HF nonlinear correlation sub-network, the nonlinear encoder network (if activated), and the LF predictor networks.
The three complexity configurations are: (1) nonlinear networks with three hidden layers with 16 neurons each, combined with exact LF function(s), (2) nonlinear networks with a single hidden layers with 8 neurons each, combined with exact LF function(s), and (3) nonlinear networks with a single hidden layers with 8 neurons each, but using learned LF surrogate(s).
For the KANs, we treat the number of neurons as the basis functions in each hidden layer. For example, a three layer KAN using 16 neurons would be a $[x, 16, 16, 16, y]$ KAN, where $x$ and $y$ are the input and output dimensions, respectively.
When the use of exact LF functions are available, they provide an upper bound to achievable accuracy of the HF surrogates, corresponding to the limit case of perfectly accurate LF surrogates.

% Number of HF and LF samples
To explore sensitivity to HF sample sizes, we generate HF data from Sobol' samples~\cite{SOBOL196786} of size \{8, 16, 32\}, augmented with boundary points to avoid extrapolation error dominating the results. Repeated Sobol' sequences are extracted using randomized power-of-two offsets.
For cases where the input dimension is greater than one, the training set sizes are increased to \{64, 128, 256\} HF samples. 
Testing sets are drawn independently from a uniform distribution of equal size to the training sets.
When one or more LF networks are also trained, we adopt an LF-to-HF oversampling ratio of 8.
Each test case is evaluated under three encoding choices: no encoding, linear encoding, and nonlinear encoding.

Finally, we repeated the same numerical experiments using a single-fidelity HF network in order to quantify the effect of using a MF network versus not using one. These experiments involved 3 different network complexities and 3 different HF dataset sizes, resulting in 9 single-fidelity network configurations. The single-fidelity networks use identical architectures to their multi-fidelity counterparts but are trained exclusively on the high-fidelity data without access to low-fidelity information. This direct comparison enables us to assess the performance benefits achieved through multi-fidelity fusion across different test cases and network configurations.

% ===============================================
\subsection{Hyperparameter optimization}\label{sec:Hyperparameter optimization strategy}
% ===============================================

To ensure a systematic and fair comparison of the performance between the different architectures, we use Hyperopt~\cite{bergstra2013making} to determine the best set of hyperparameters for each test case.
The optimized hyperparameters include the learning rate, an LF $\ell_2$ regularization penalty $\lambda_L$, a penalty $\lambda^{nl}_{H}$ for the HF nonlinear correlation network weight $\ell_2$ norm, a penalty associated with the difference between $\widehat{y}_H$ and $\widehat{y}_H^{l}$, and one associated with the IS term.
The exact distribution and range of each hyperparameter are provided in Table~\ref{table:hyperopt}. This hyperparameter space was used for all networks, regardless of whether they included an encoding block.
Finally, all results reported in Section~\ref{sec:results} are computed using the Adam optimizer~\cite{adam2014}. 

Separate studies have also been conducted to investigate the effects of using a different optimizer, a different learning rate decay factor, warm solution restarts, and a varying number of epochs.
\begin{table}[htb!]
\centering
% \resizebox{\linewidth}{!}{
% \small
\begin{tabular}{c c c} 
\toprule
{\bf Hyperparameter} & {\bf Type} & {\bf Range}\\
\midrule
lr & loguniform  &  $\log([10^{-5}, 10^{-3}])$
\\
$\lambda_{L}$ & Choice & $[0, 10^{-5}, 10^{-3}, 1]$
\\
$\lambda^{nl}_{H}$ & Choice & $[0, 10^{-5}, 10^{-3}, 1]$
\\
$\lambda_{T}$ & Choice & $[0, 10^{-5}, 10^{-3}, 1]$
\\
$\lambda_{D}$ & Choice & $[0, 10^{-5}, 10^{-3}, 1]$
\\
\bottomrule
\end{tabular}
% }

\caption{Hyperparameter space definition in Hyperopt.}\label{table:hyperopt}
\end{table}

% ============================================
\section{Numerical results}\label{sec:results}
% ============================================

Next, we present a series of test cases that demonstrate the performance of the multi-fidelity network configurations discussed above in approximating various models. These models are characterized by low- and high-dimensional inputs, involve one or more low-fidelity predictors, and are either available in closed form or derived from the solution of a PDE.
In Table \ref{table:test_problems}, we also provide a reference for each test case.

\begin{table}[h!]
\centering
\begin{tabular}{l c c c c} % Use p{} to control column width
\toprule
\textbf{Section} & \textbf{Dimensionality} & \textbf{Number of LF} & \textbf{Solution} & \textbf{Reference} \\ 
\midrule
\S 3.1.1-5 & 1 & 1 & Analytical & 
%A composite neural network that learns from multi-fidelity data: Application to function approximation and inverse PDE problems
\cite{meng2020composite} \\
\S 3.2.1-3 & 1 & 2 & Analytical & 
%A composite neural network that learns from multi-fidelity data: Application to function approximation and inverse PDE problems
\cite{meng2020composite} \\
\S 3.3.1 & 2 & 1 & Analytical &
%multi-fidelity methods for uncertainty quantification in cardiovascular hemodynamics~
\cite{fleeterThesis} \\
\S 3.3.2 & 2/3 & 1 & Analytical &
%multi-fidelity methods for uncertainty quantification in cardiovascular hemodynamics~
\cite{fleeterThesis} \\
\S 3.3.3 & 2 & 1 & PDE & 
%Mathematical models of threshold phenomena in the nerve membrane~
\cite{fitzhugh1955mathematical} \\ 
\S 3.3.4 & 2 & 8 & Analytical &
%MFNets: data efficient all-at-once learning of multi-fidelity surrogates as directed networks of information sources~
\cite{Gorodetsky2021} \\
\S 3.4.1 & 20 & 1 & Analytical & 
%A composite neural network that learns from multi-fidelity data: Application to function approximation and inverse PDE problems
\cite{meng2020composite} \\
\bottomrule
\end{tabular}
\caption{List of test problems with relevant features and references.}
\label{table:test_problems}
\end{table}

% =====================================
\subsection{One dimensional test cases with single low-fidelity}
% =====================================

% ================================================
\subsubsection{1D linearly correlated models (K1)}
\label{K1 example}
% ================================================

\begin{figure}[ht!]
\centering
\includegraphics[width=0.9\linewidth]{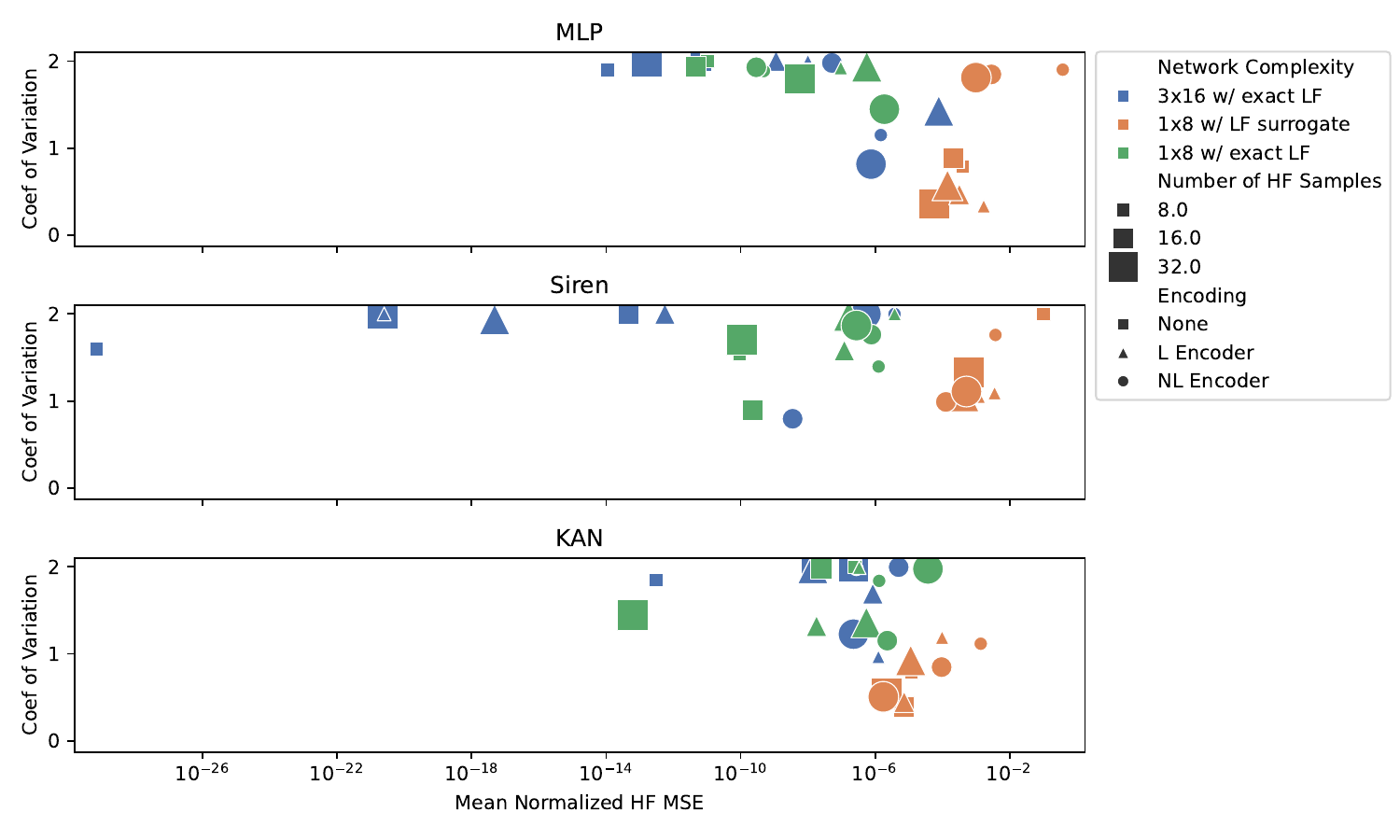}
\caption{Mean normalized HF MSE results for example K1.}
\label{fig:K1_scatter}
\end{figure}

Consider the model pair
\begin{equation}
\begin{aligned}
    y_L(x) =& 0.5(6 x-2)^2 \sin (12 x-4)+10x-10, & x \in [0,1]
    , \\
    y_H(x) =& (6 x-2)^2 \sin (12 x-4), & x \in [0,1]
    .\\
\end{aligned}
\end{equation}
where $y_H$ can be expressed as a \emph{linear} function of $y_L$, i.e., $y_H = 2 y_L - 20 x + 20$. 
The standard MF framework without coordinate encoding has been shown to be able to recover the HF function and the correlation between the HF and LF models well using MLPs~\cite{meng2020composite}.
As such, this example serves as a verification case to show the other network architectures can perform similarly for this simple example.
In Figure~\ref{fig:K1_scatter}, we show the resulting mean normalized MSE versus its CoV over 4 independent repetitions. 
At the same time, the figure show results for networks of varying complexity, different coordinate encoding, and trained using a different number HF samples. 
From the figure, we can see all the configurations are able to produce accurate HF models with most networks able to reach normalized MSE under $10^{-4}$.
When the LF response is also learned during training, the relative MSE is unable to drop below $10^{-6}$ and, in such a case, KAN performs best most often, followed by Siren and MLP. 
% Difference between number of samples
In addition an increasing number of samples leads to better results for Siren and MLP, but results appear to be less sensitive for KAN. 
% Difference between encoding 
Finally, best results are achieved without an encoder, meaning that, for this simple problem, the addition of an encoder only leads to a more complicated training. 

\begin{figure}[ht!]
\centering
\includegraphics[width=0.32\linewidth]{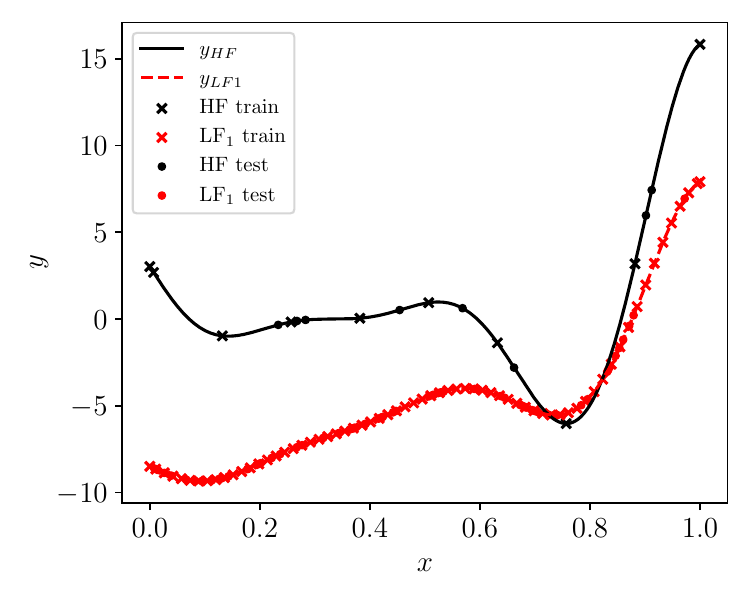}
\includegraphics[width=0.32\linewidth]{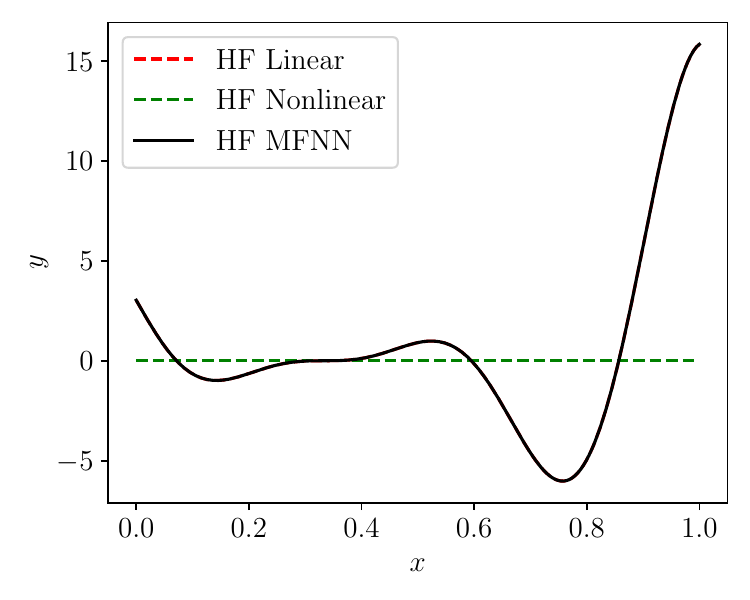}

\caption{Best HF and LF model fitting results plotted against using 8/64 HF/LF training samples and an exact LF model for a Siren network with no encoding (left) with the learned linear/nonlinear HF correlation (right).}
\label{fig:K1_fit}
\end{figure}

Figure~\ref{fig:K1_fit} shows the multi-fidelity neural network framework applied to a simple test case using 8 high-fidelity and 64 low-fidelity training samples with a Siren architecture. The left panel shows successful fitting of both fidelity levels, while the right panel displays the learned linear and nonlinear correlation components between them. Although this represents a straightforward function approximation problem where direct high-fidelity training would suffice, it validates the multi-fidelity methodology and demonstrates the framework's ability to capture cross-fidelity relationships.

% ===========================================================
\subsubsection{1D Piece-wise linearly correlated models with colocated discontinuities (K2)}\label{K2 example}
% ===========================================================

\begin{figure}[ht!]
\centering
\includegraphics[width=0.9\linewidth]{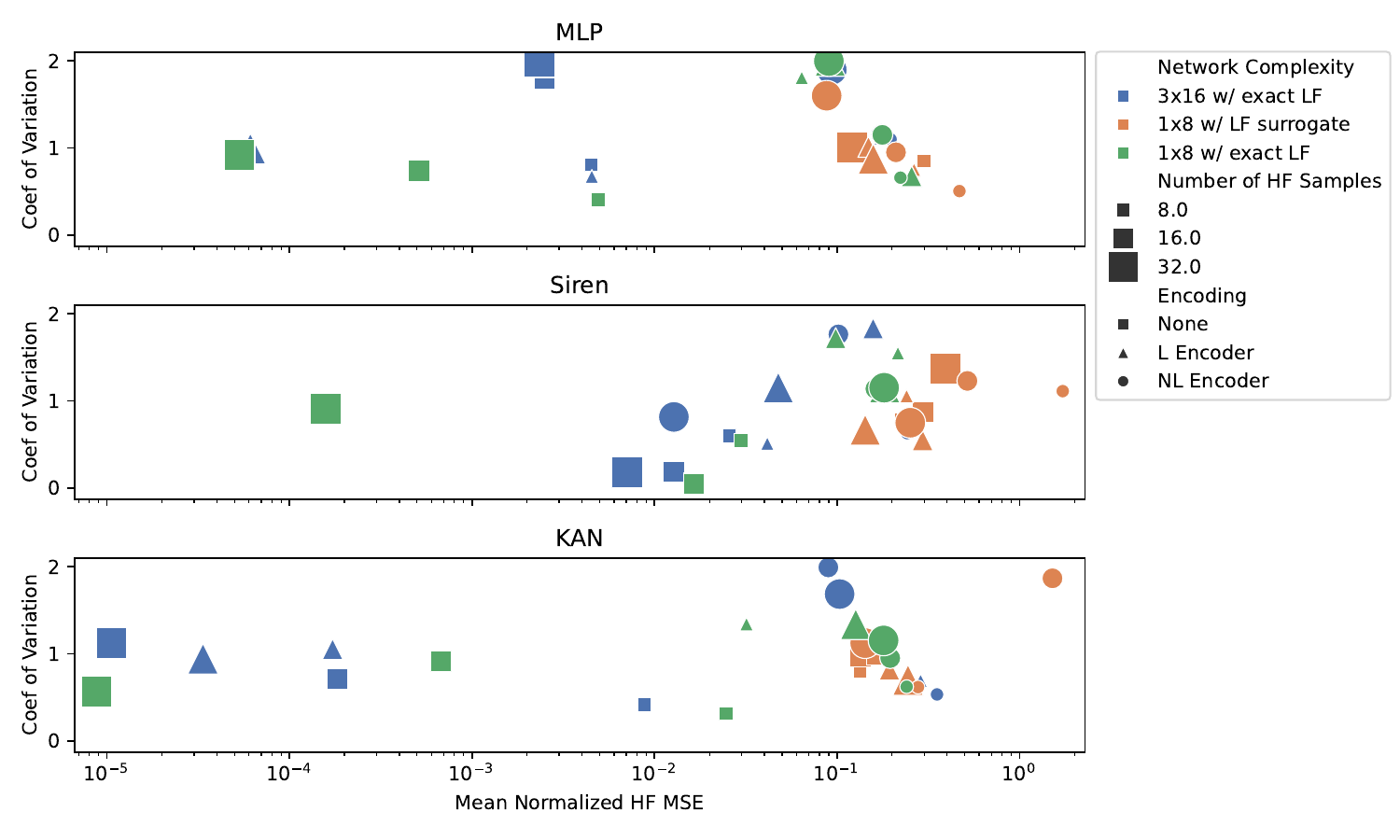}
\caption{Mean normalized HF MSE over 4 repetitions for example K2 using various network configurations.}
\label{fig:K2_scatter}
\end{figure}

Consider the model pair
\begin{equation}
\begin{aligned}
    y_L(x) =&
    3 \cdot \mathds{1}_{(0.5,1]} + 0.5(6 x-2)^2 \sin (12 x-4)+10x-10
     , & x \in [0,1],
    \\
    y_H(x) =&
    4 \cdot \mathds{1}_{(0.5,1]} + 2 y_L(x)-20 x+20
    , & x \in [0,1].
\end{aligned}
\end{equation}
In this test case, $y_H$ is a piece-wise linear function of $y_L$. A discontinuity is also introduced at $x=1/2$ in the LF and HF function, with a magnitude of $3$ and $4$, respectively.
The results in Figure~\ref{fig:K2_scatter} show that best MSE is achieved with the least encoder complexity (no encoder).
The resulting MSE is very sensitive to the ability of the predictor to capture the correct location of the discontinuity, and HF testing samples that are drawn closer to the discontinuity then their training counterpart, produce sensible jumps in the resulting MSE. 
Therefore, for this test case, an identity encoding (or no encoding) represents an optimal choice, since it does not alter the information on the discontinuity location learned from the LF model.
Even for this case, simultaneously learning the LF predictor severely limits the resulting MSE.  

\begin{figure}[ht!]
\centering
\includegraphics[width=0.32\linewidth]{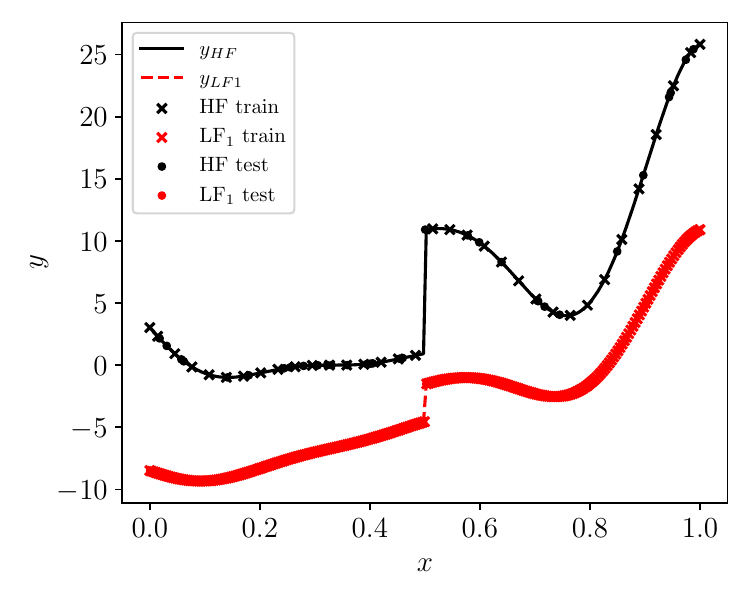}
\includegraphics[width=0.32\linewidth]{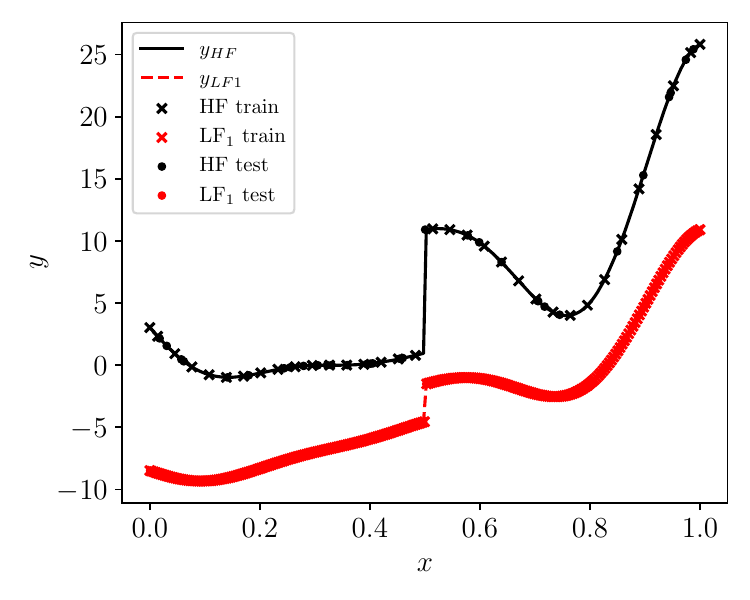}
\includegraphics[width=0.32\linewidth]{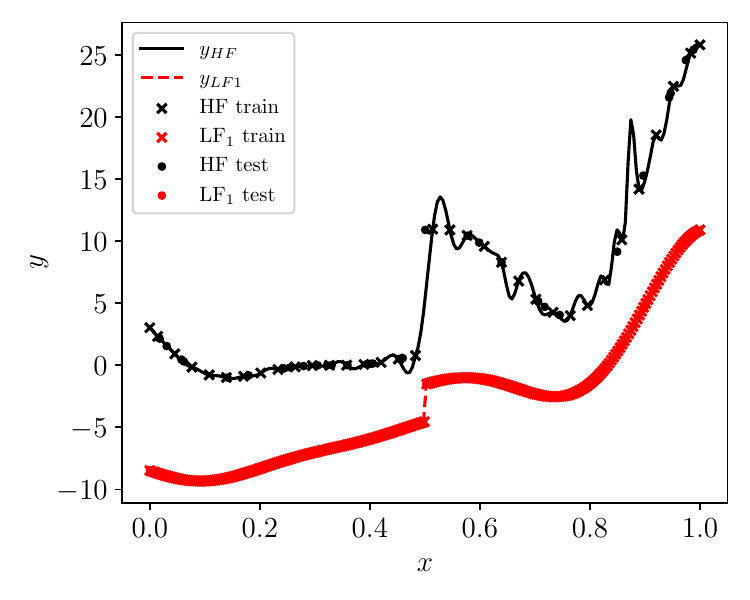}
\caption{HF and LF model fitting results for 32/256 HF/LF training samples, and exact LF predictor. We consider a network with no encoding (left), linear encoding (middle), and nonlinear encoding (right) for example K2.}
\label{fig:K2_fits}
\end{figure}

To better understand the tendencies of each network configuration and encoding strategy, we inspect the HF model fits using 32 HF samples and an exact LF predictor.
Figure \ref{fig:K2_fits} shows a comparison between the best performing KANs using no encoding, using linear encoding , and using nonlinear encoding, respectively.
The figure shows that the nonlinear encoding network moves the discontinuity in the HF function to fit (or overfit) the training data, at the expense of over-fitting on the right side of the discontinuity location. This explains the higher MSE observed for all nonlinear encoding networks for this example.

\begin{figure}[ht!]
\centering
\includegraphics[width=0.32\linewidth]{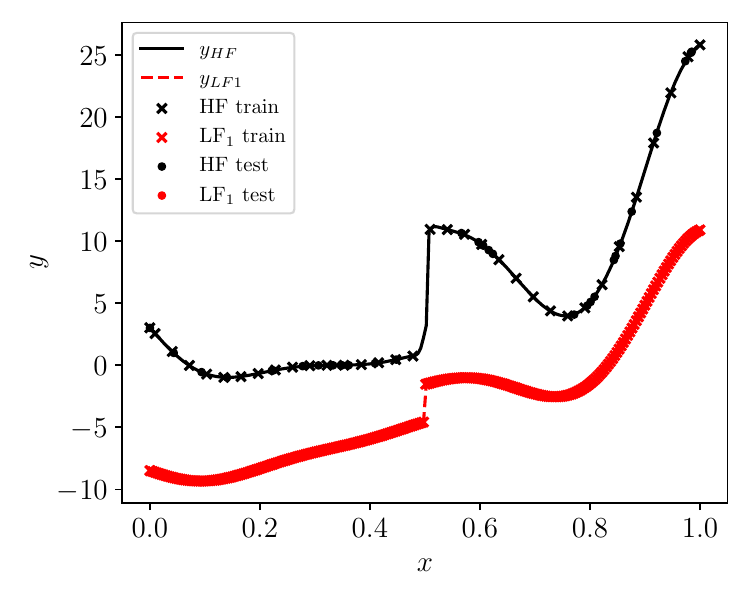}
\includegraphics[width=0.32\linewidth]{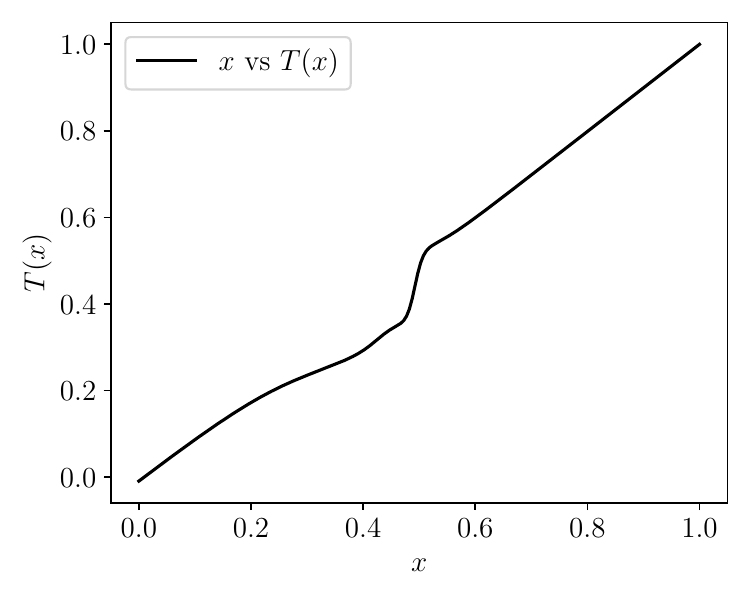}
\includegraphics[width=0.32\linewidth]{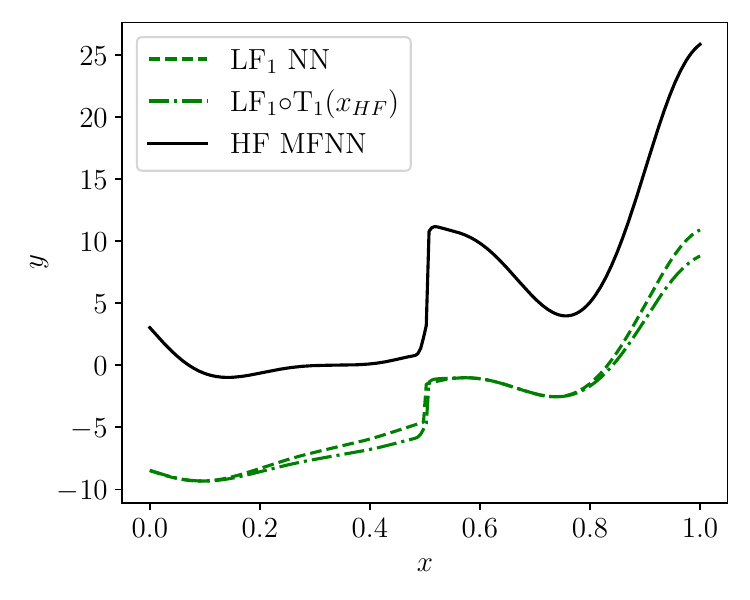}
\caption{Best HF and LF model fitting results plotted against using 32/256 HF/LF training samples and an exact LF model for a network with nonlinear encoding (left), the learned nonlinear coordinate encoding $T_1$ (middle), and the LF function composed with its encoding $y_{L} \circ T$ (right) for example K2.}
\label{fig:K2_NL_fit}
\end{figure}

This is evident when inspecting the best performing nonlinear encoding network in Figure~\ref{fig:K2_NL_fit}, a Siren network.
The figure shows the resulting nonlinear coordinate encoding $T_1$, and how it affects the LF predictors, i.e., $y_{L_1} \circ T_1$. Interestingly, the encoding $T_1$ stretches the domain before and after $x=0.5$ while shrinking a small part of the domain near the discontinuity location.

\begin{figure}
    \centering
    \includegraphics[width=0.7\linewidth]{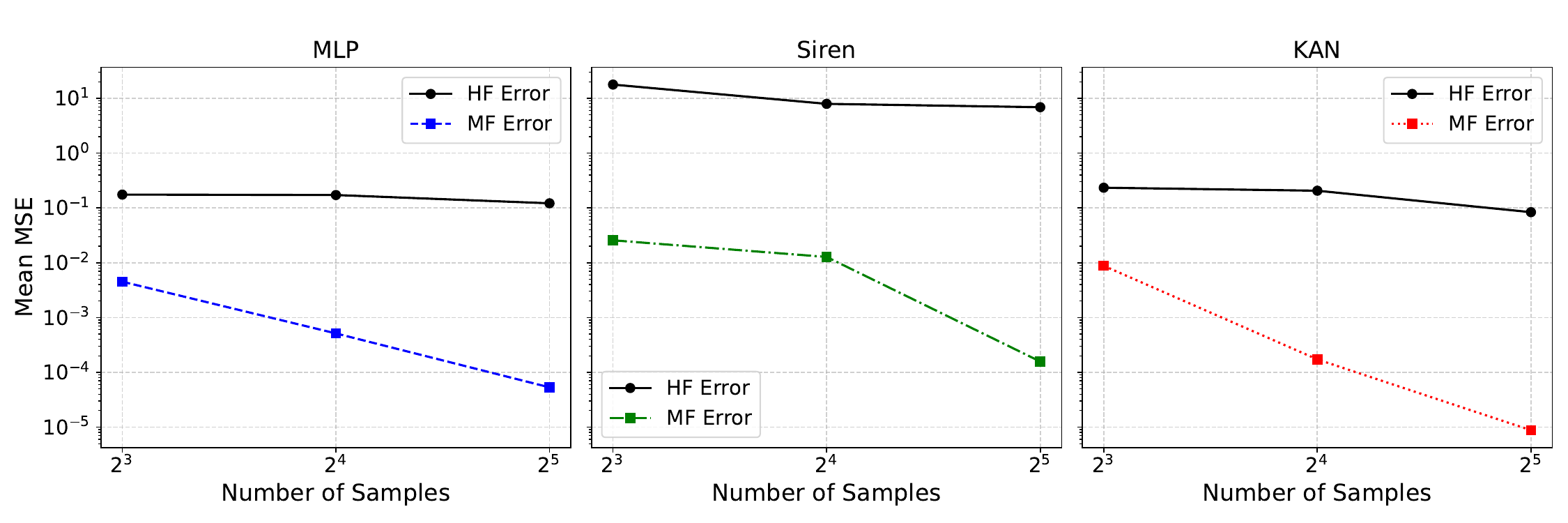}
    \includegraphics[width=0.25\linewidth, height=3.5cm]{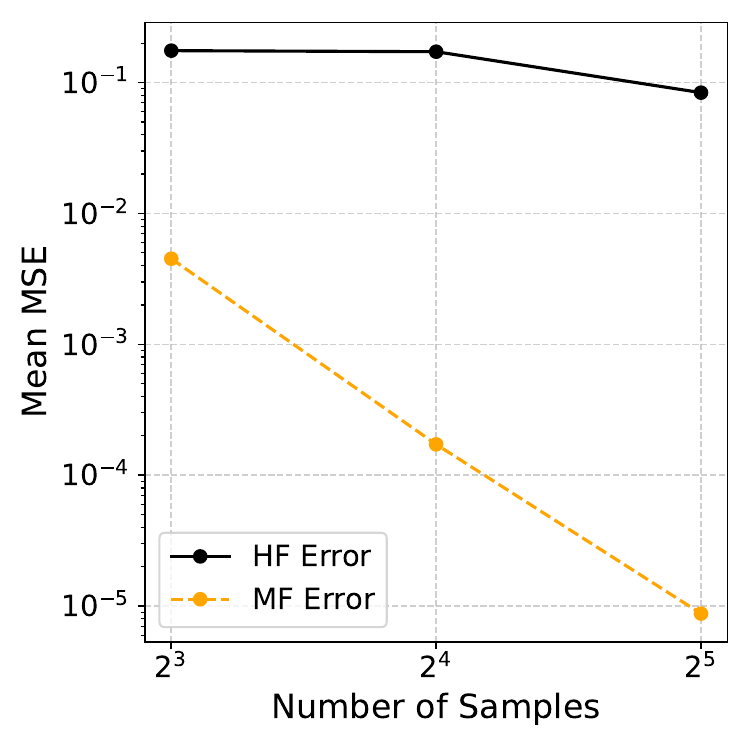}
    \caption{
    Mean MSE results of test case K2 comparing the best-performing single high-fidelity (HF) network and the best multi-fidelity network: results separated by network architecture across different HF sample sizes (left) and the best results at each HF sample size are shown, regardless of architecture type or network complexity (right).}
    \label{fig:MSE_Lines_k2}
\end{figure}

When comparing these multi-fidelity results with equivalent single-fidelity networks trained solely on HF data, the MF approach demonstrates substantial improvements at every sample size for each architecture as evident in Figure~\ref{fig:MSE_Lines_k2}.
Single-fidelity networks trained with the same number of HF samples showed normalized MSE values consistently 1-4 orders of magnitude higher than their MF counterparts.
This performance gap was most pronounced for larger sample sizes (32 HF samples).
The advantage of incorporating LF information is clear in this test case.

% ===========================================================
\subsubsection{1D Piece-wise linearly correlated models with shifted discontinuity (K2 shift)}\label{K2 shift example}
% ===========================================================

\begin{figure}[ht!]
\centering
\includegraphics[width=0.9\linewidth]{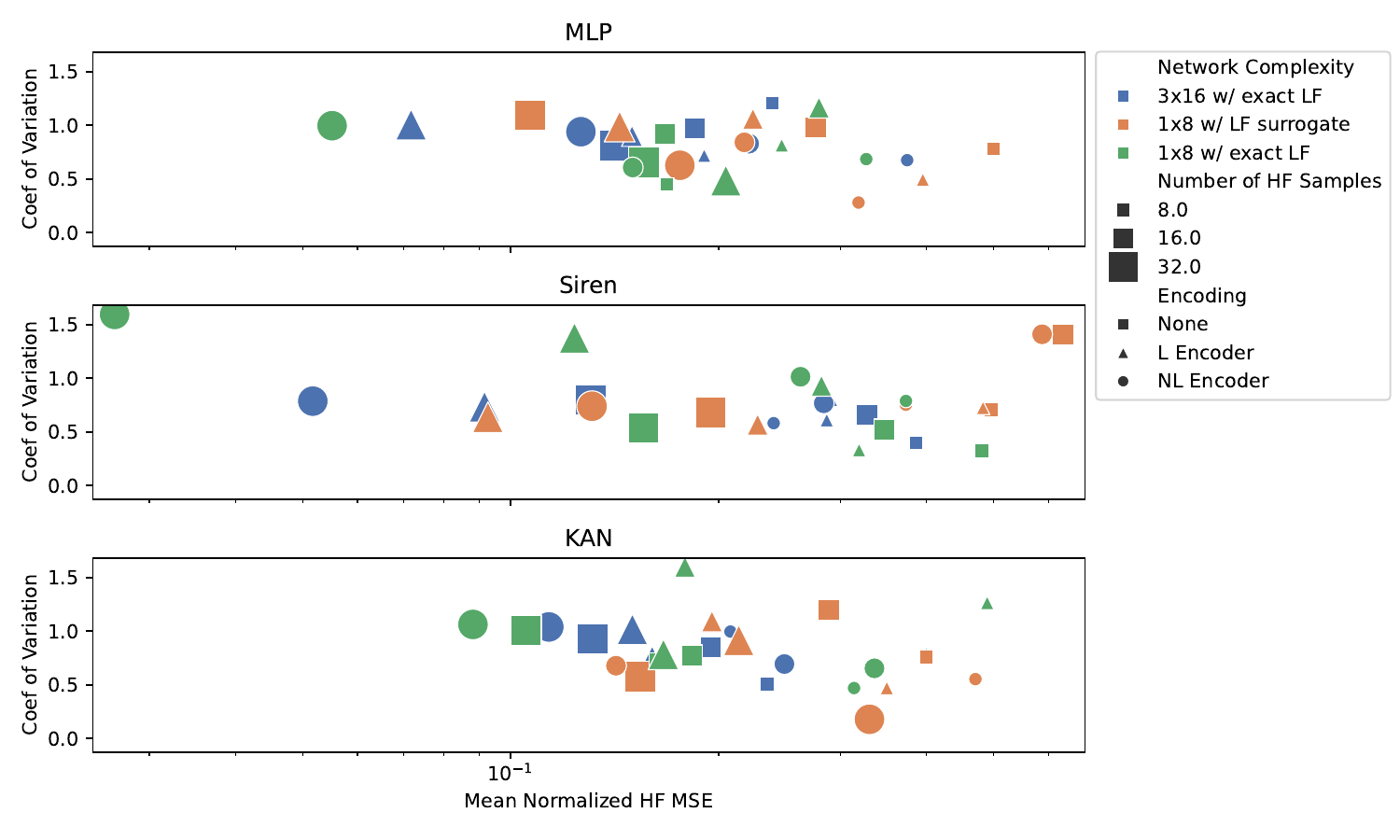}
\caption{Mean normalized HF MSE for example K2 with shifted discontinuity.}
\label{fig:K2s_scatter}
\end{figure}

Consider now the model pair
\begin{equation}
\begin{aligned}
    y_L(x) =&
    3 \cdot \mathds{1}_{(0.6,1]} + 0.5(6 x-2)^2 \sin (12 x-4)+10x-10
     , & x \in [0,1],
    \\
    y_H(x) =&
    4 \cdot \mathds{1}_{(0.5,1]} + 2 y_L(x)-20 x+20
    , & x \in [0,1].
\end{aligned}
\end{equation}
We tested this model pair using the same experimental setup as the previous test case. This allowed us to directly observe how network performance changes when discontinuities appear at different locations in the LF and HF functions.
This test case is very similar the previous one in Section~\ref{K2 example}, however, the discontinuity in the LF and HF are located in a different position, i.e., at $x=0.6$ and $x=0.5$, respectively.
In this case, we expect the standard network to struggle to capture the HF response without a significant nonlinear correlation between models, while the encoded networks should be able to shift the discontinuity appropriately.
As expected, Figure~\ref{fig:K2s_scatter} shows a reversal in effectiveness in coordinate encoding, where the nonlinear encoding is most accurate followed by linear and no encoding, in that order. 
Moreover, the nonlinear encoding seems to perform best for every network architecture, suggesting that networks augmented with an encoder are better capable to handle changes in discontinuity location.
That being said, the normalized MSE for this example is significantly higher than for the previous test case.

\begin{figure}[ht!]
\centering
\includegraphics[width=0.32\linewidth]{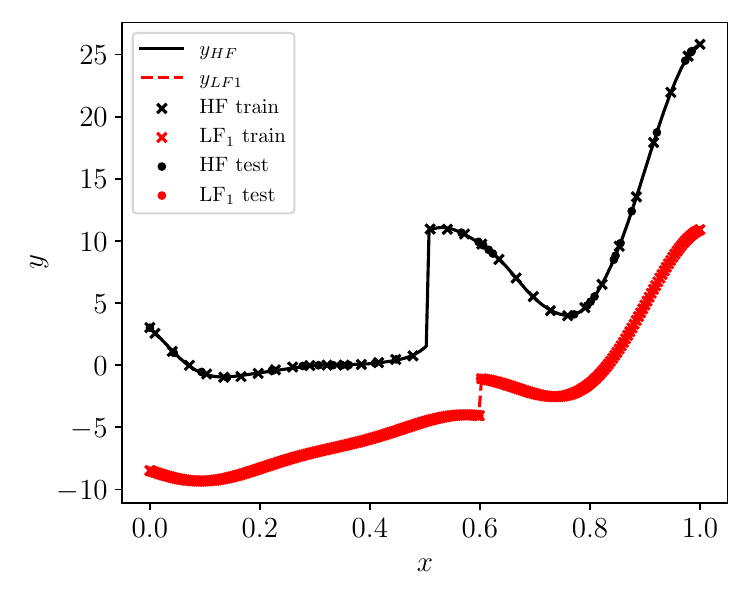}
\includegraphics[width=0.32\linewidth]{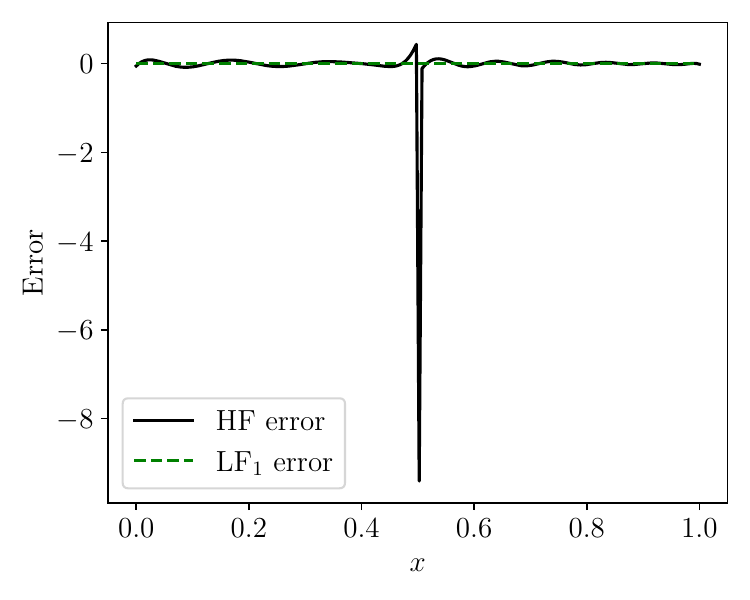}
\includegraphics[width=0.32\linewidth]{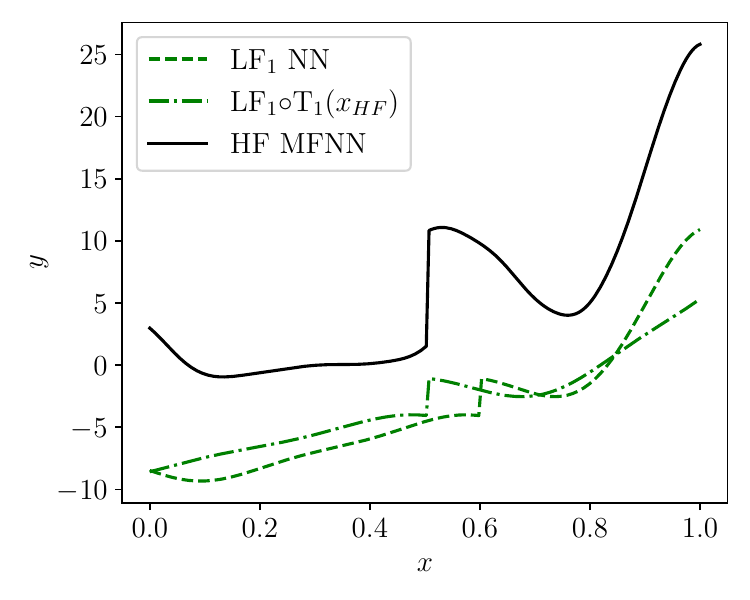}
\caption{Results for 32/256 HF/LF training samples using an exact LF predictor for the K2 test case with shifted discontinuity. Best fits results for nonlinear encoding (left), HF and LF model errors (middle), and encoded LF predictor $y_{L_1} \circ T_1$ (right).}
\label{fig:Best_KS2_NL_fit}
\end{figure}

When inspecting the best performing network, a Siren network with nonlinear encoding, it's clear why all the normalized MSE are so high. 
As shown in Figure~\ref{fig:Best_KS2_NL_fit}, the HF model is very accurate everywhere but a small neighborhood around the discontinuity where the error is large, suggesting that the MSE is still dominated by the HF approximation near the discontinuity.
However, the effectiveness of a coordinate encoding becomes evident in this case, where the encoding ($y_{L_1} \circ T_1$) is successful in shifting the discontinuity from $x=0.6$ into $x=0.5$.

\begin{figure}
    \centering
    \includegraphics[width=0.7\linewidth]{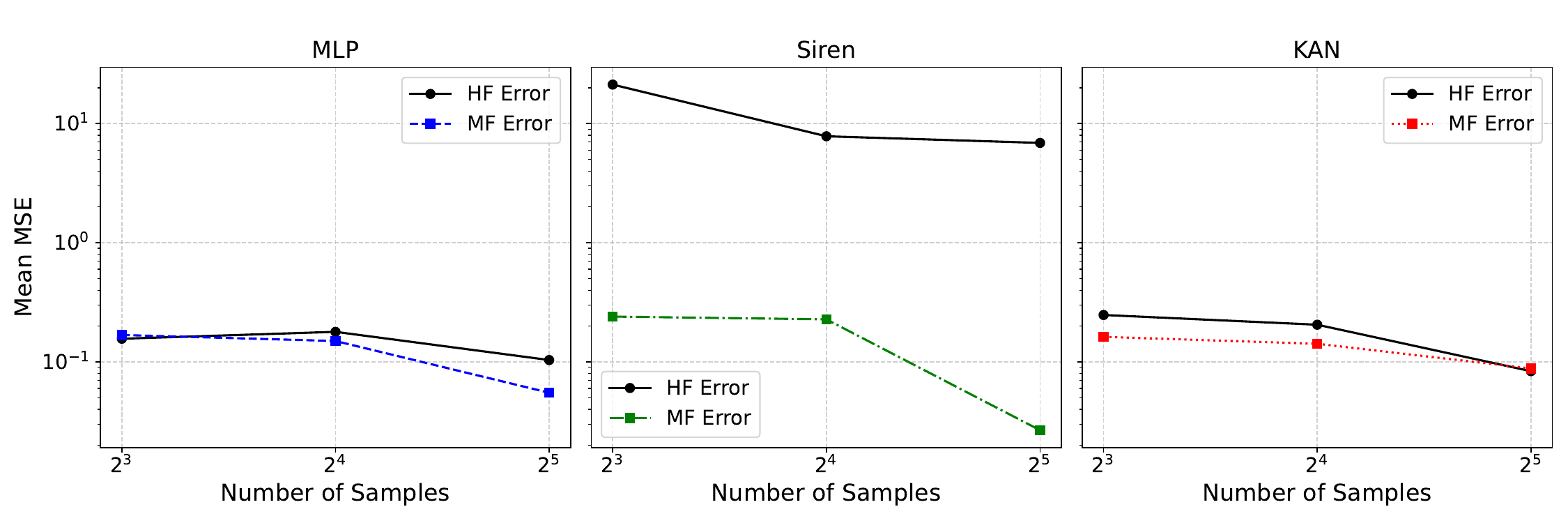}
    \includegraphics[width=0.25\linewidth, height=3.5cm]{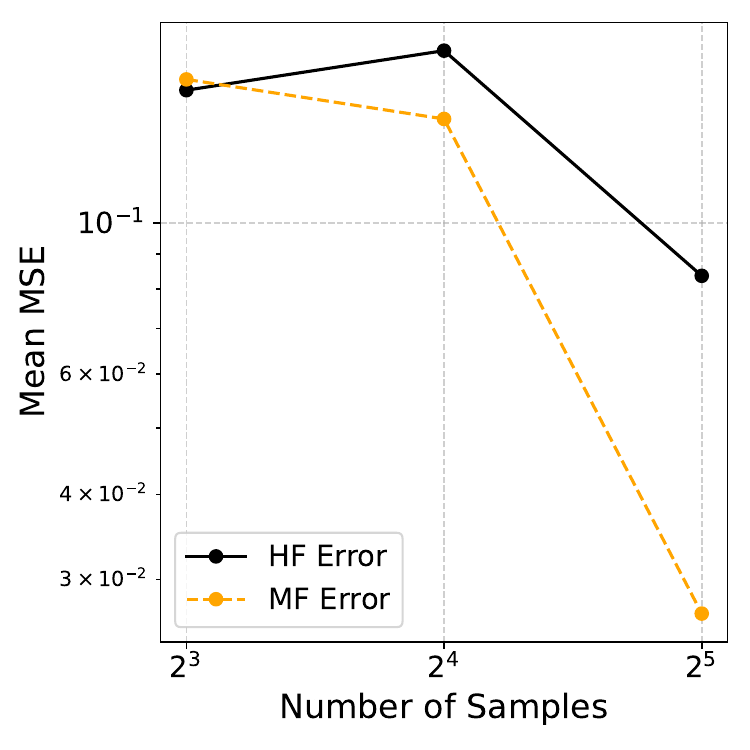}
    \caption{
    Mean MSE results of test case K2 shift comparing the best-performing single high-fidelity (HF) network and the best multi-fidelity network: results separated by network architecture across different HF sample sizes (left) and the best results at each HF sample size are shown, regardless of architecture type or network complexity (right).}
    \label{fig:MSE_Lines_k2s}
\end{figure}

Interestingly, when comparing these multi-fidelity results with equivalent single-fidelity networks trained solely on HF data for this test case in Figure~\ref{fig:MSE_Lines_k2s}, the benefits of the MF approach were not immediately apparent with small sample sizes. Single-fidelity networks with only 8 HF samples performed comparably to their MF counterparts, both struggling to accurately locate the dislocated discontinuity. However, as the number of HF samples increased to 16 and 32, multi-fidelity networks with coordinate encoding began to demonstrate advantages. This suggests that for problems with dislocated features, the multi-fidelity approach requires a minimum threshold of HF samples to effectively leverage the LF information through appropriate coordinate transformations.

% ==============================================================================
\subsubsection{1D nonlinearly correlated oscillatory models (K3)}
% ==============================================================================

\begin{figure}[ht!]
\centering
\includegraphics[width=0.9\linewidth]{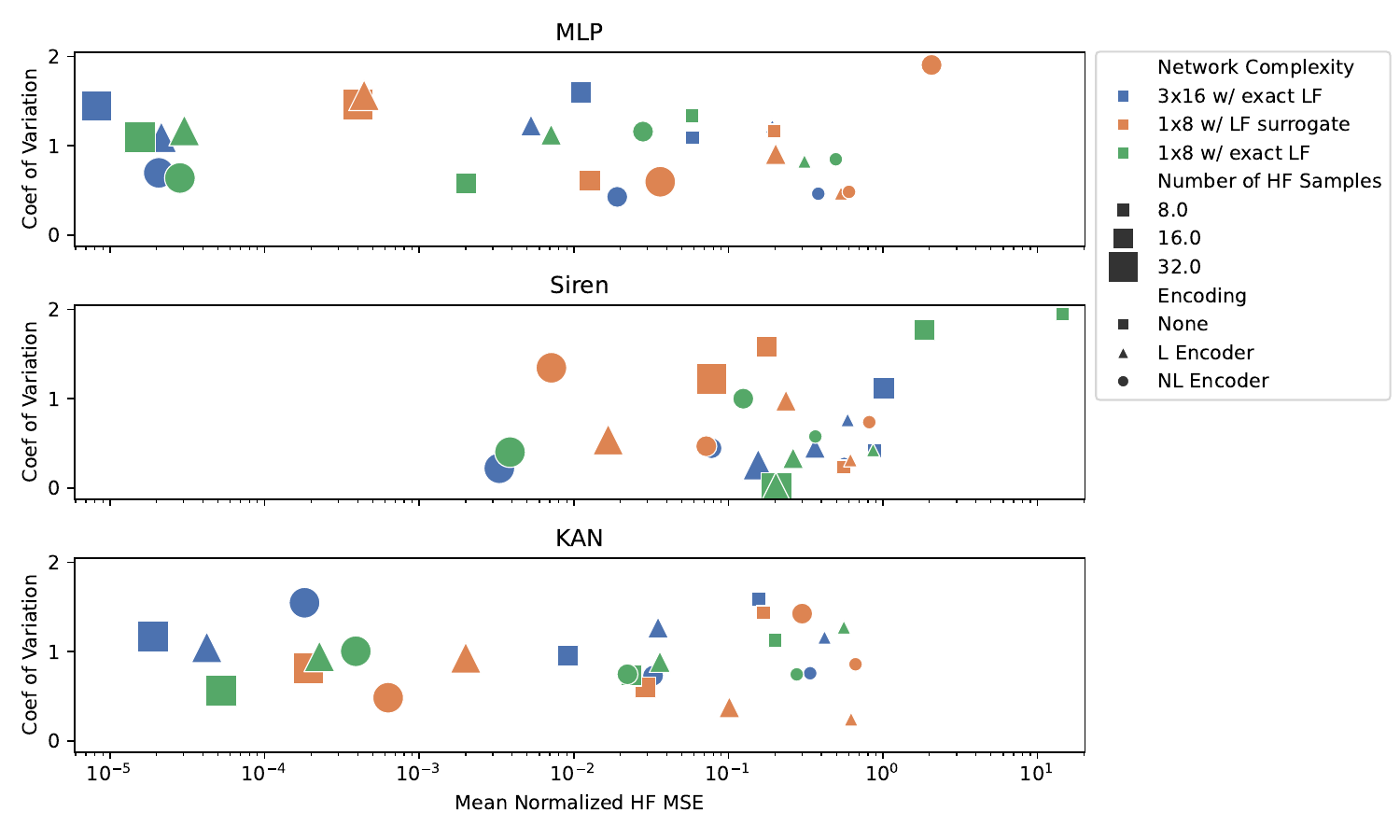}
\caption{Mean normalized HF MSE for test case K3.}
\label{fig:K3_scatter}
\end{figure}

\begin{figure}[ht!]
\centering
\includegraphics[width=0.32\linewidth]{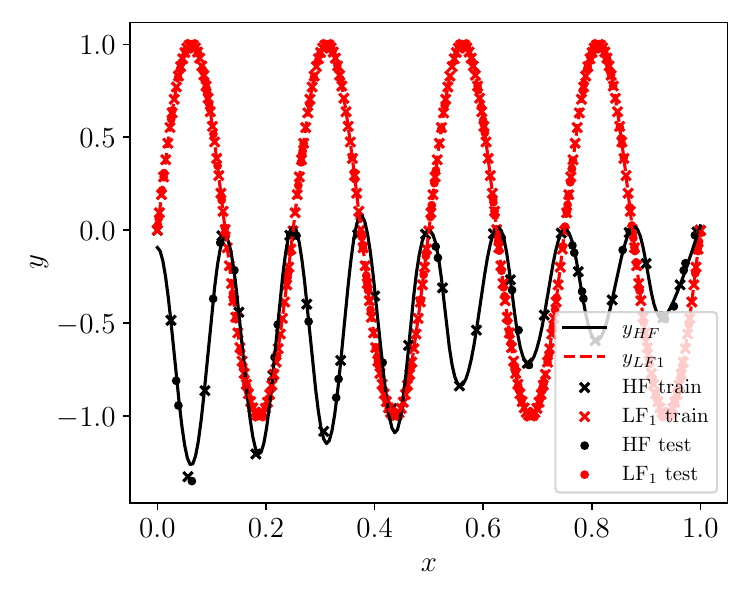}
\includegraphics[width=0.32\linewidth]{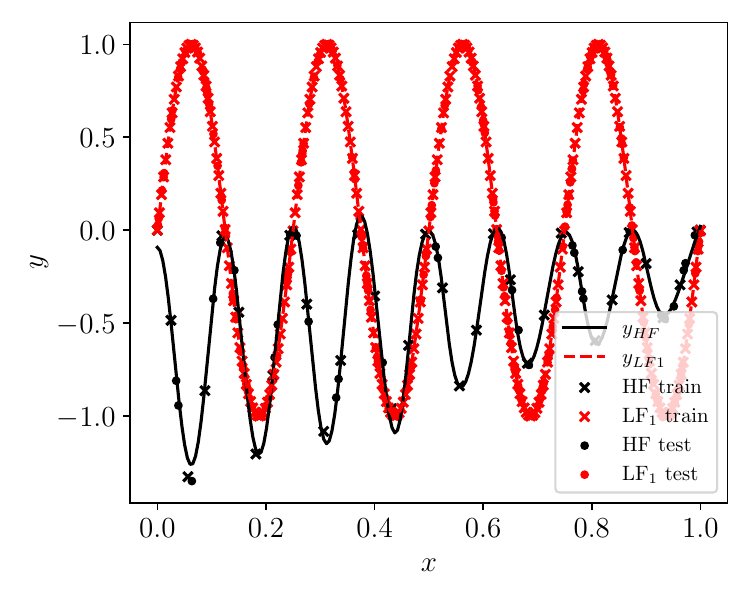}
\includegraphics[width=0.32\linewidth]{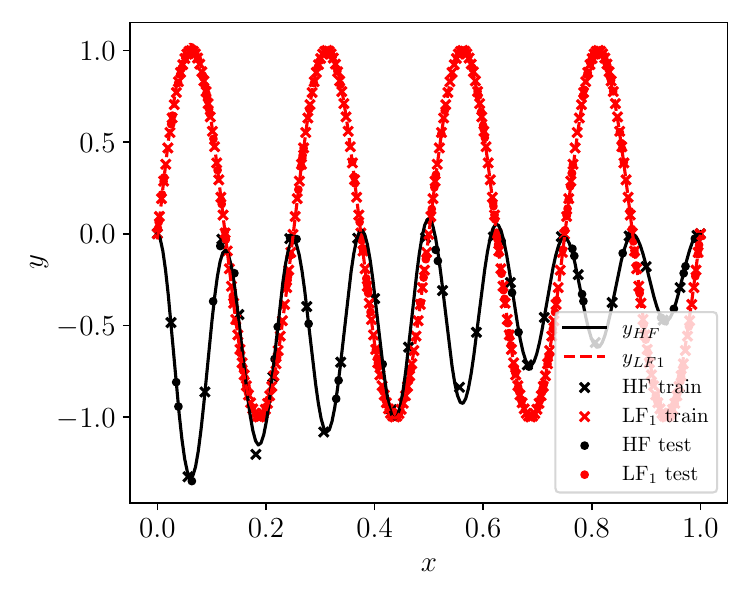}
\\
\includegraphics[width=0.32\linewidth]{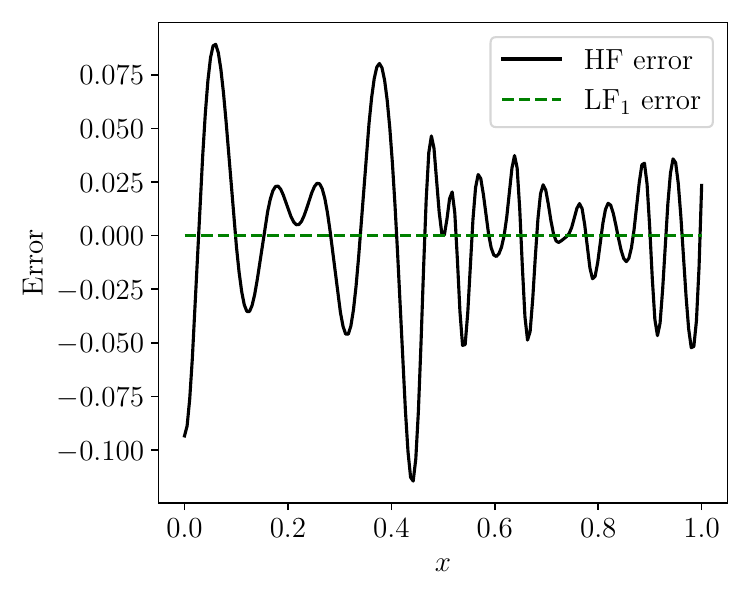}
\includegraphics[width=0.32\linewidth]{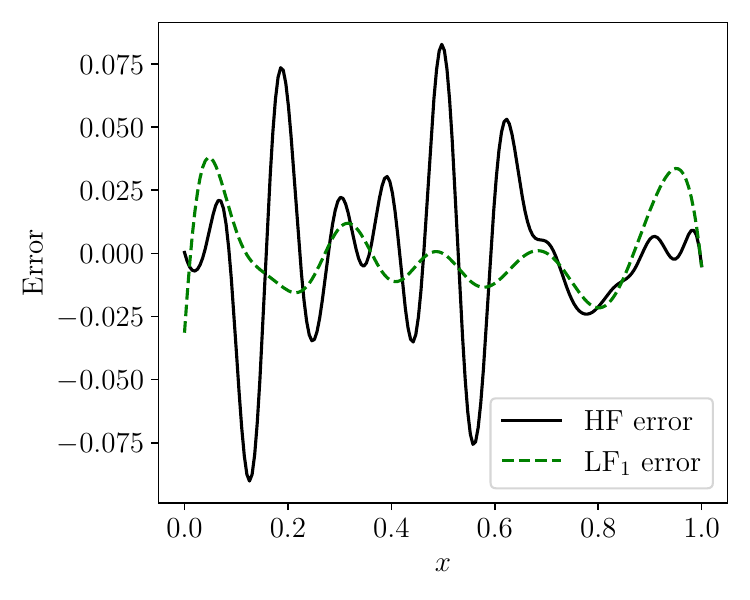}
\includegraphics[width=0.32\linewidth]{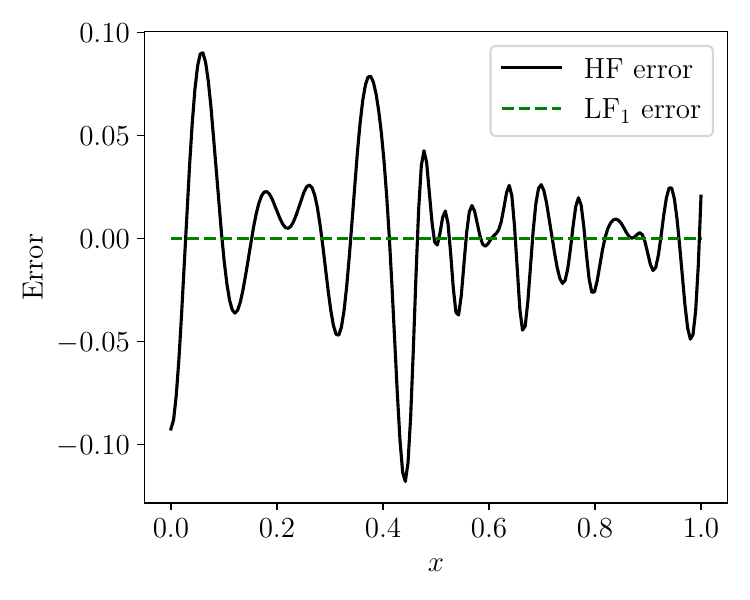}
\caption{Most accurate HF and LF predictors (top row) resulting from 32/256 HF/LF training samples and a Siren architecture, nonlinear encoding with a network complexity of 3x16 with an exact LF model (left), 1x8 with a learned LF surrogate model (middle), and 1x8 with an exact LF model (right) for example K3. Error plots (bottom row) for each type of encoder.}\label{fig:Siren_K3_fits}
\end{figure}

\begin{figure}[ht!]
\centering
\includegraphics[width=0.32\linewidth]{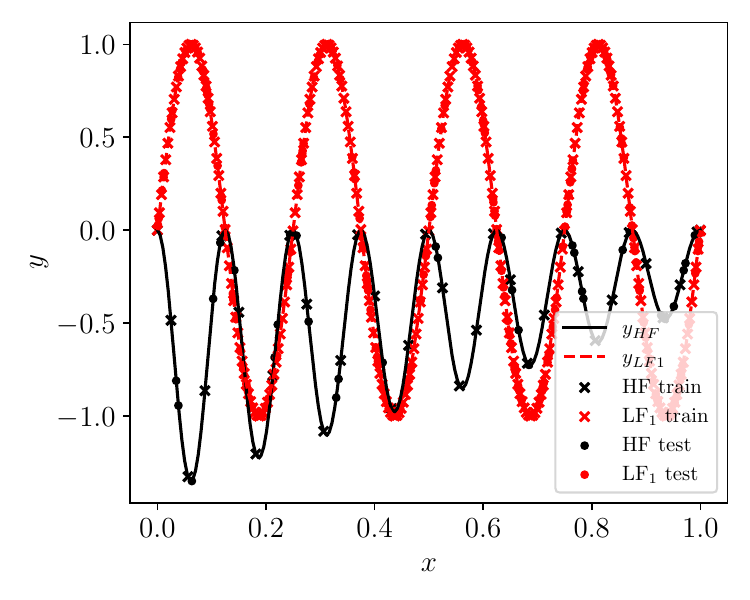}
\includegraphics[width=0.32\linewidth]{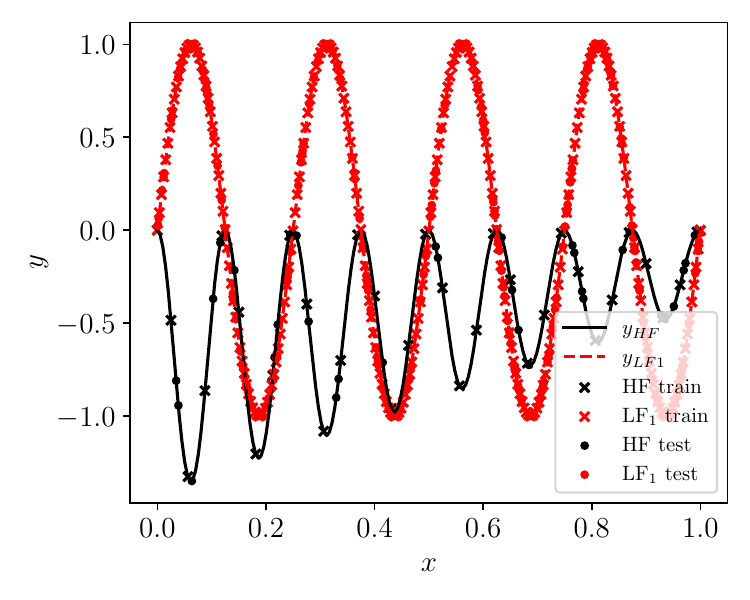}
\includegraphics[width=0.32\linewidth]{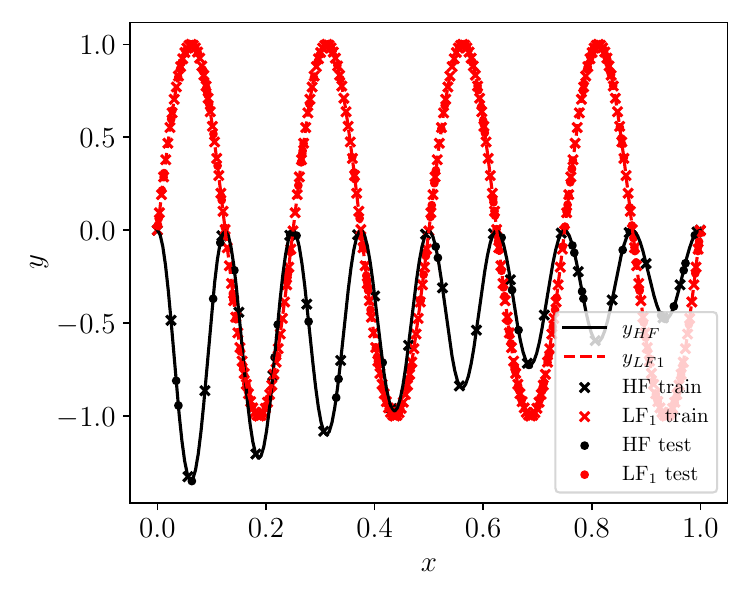}
\\
\includegraphics[width=0.32\linewidth]{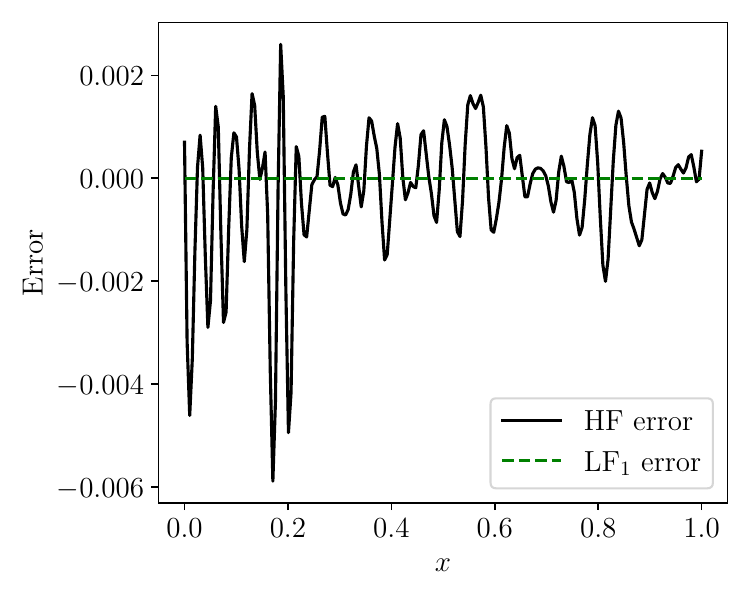}
\includegraphics[width=0.32\linewidth]{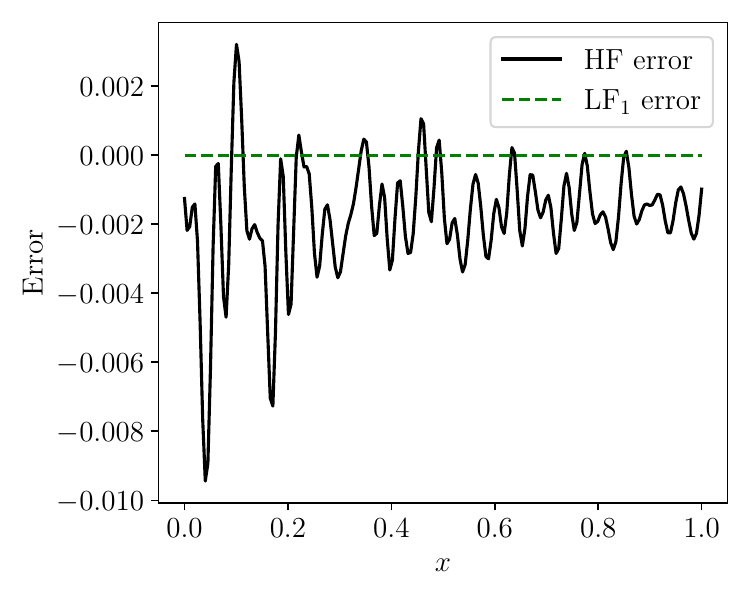}
\includegraphics[width=0.32\linewidth]{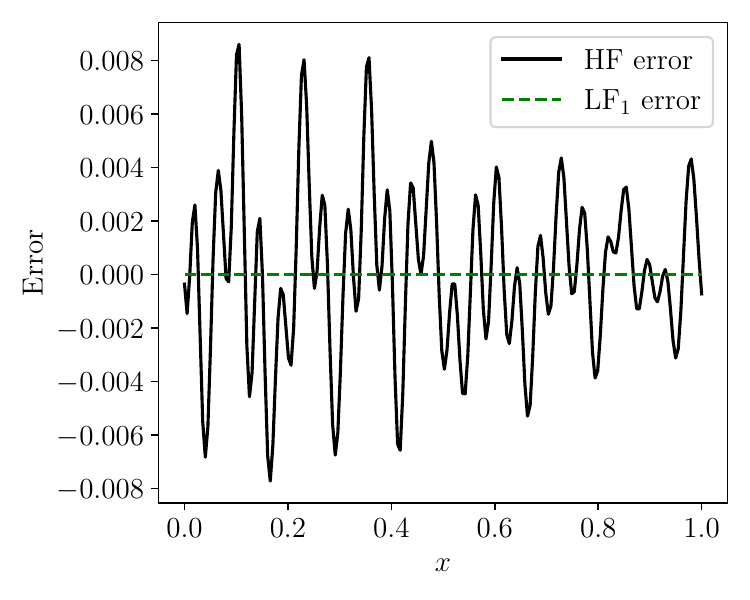}
\caption{Most accurate HF and LF predictors (top row) resulting from 32/256 HF training/testing samples, an exact LF, and a MLP architecture with no encoding (left), linear encoding (middle), and nonlinear encoding (right) for example K3. Error plots (bottom row) for each type of encoder.}\label{fig:Best_MLP_K3_fits}
\end{figure}

Consider the non linearly correlated HF/LF model pair
\begin{equation}
\begin{aligned}
y_L(x) =&
\sin(8\pi x), & x \in [0,1], \\
y_H(x) =&
(x - \sqrt{2}) y_L^2(x), & x \in [0,1].
\end{aligned}
\end{equation}
The LF and HF models are both oscillatory, but differ in frequency.
The results in Figure~\ref{fig:K3_scatter} show that many types of networks are able to learn accurate HF representations, and that no encoding appears to perform best when using MLPs or KANs.
Additionally, networks using LF surrogates do not experience significant differences in MSE magnitudes compared to the network using the exact LF model, but having enough samples appears to be of paramount importance to correctly capture minima and maxima.
Additionally, the normalized HF MSE results from using Siren networks are orders of magnitude worse than the other networks due to underfitting, as shown in Figure~\ref{fig:Siren_K3_fits}. 
Inspecting the best performing $3\times 16$ MLP networks using 32 HF samples and an exact LF, an accurate HF predictor is produced by any encoding mechanism, as shown in Figure~\ref{fig:Best_MLP_K3_fits}. 

\begin{figure}
    \centering
    \includegraphics[width=0.7\linewidth]{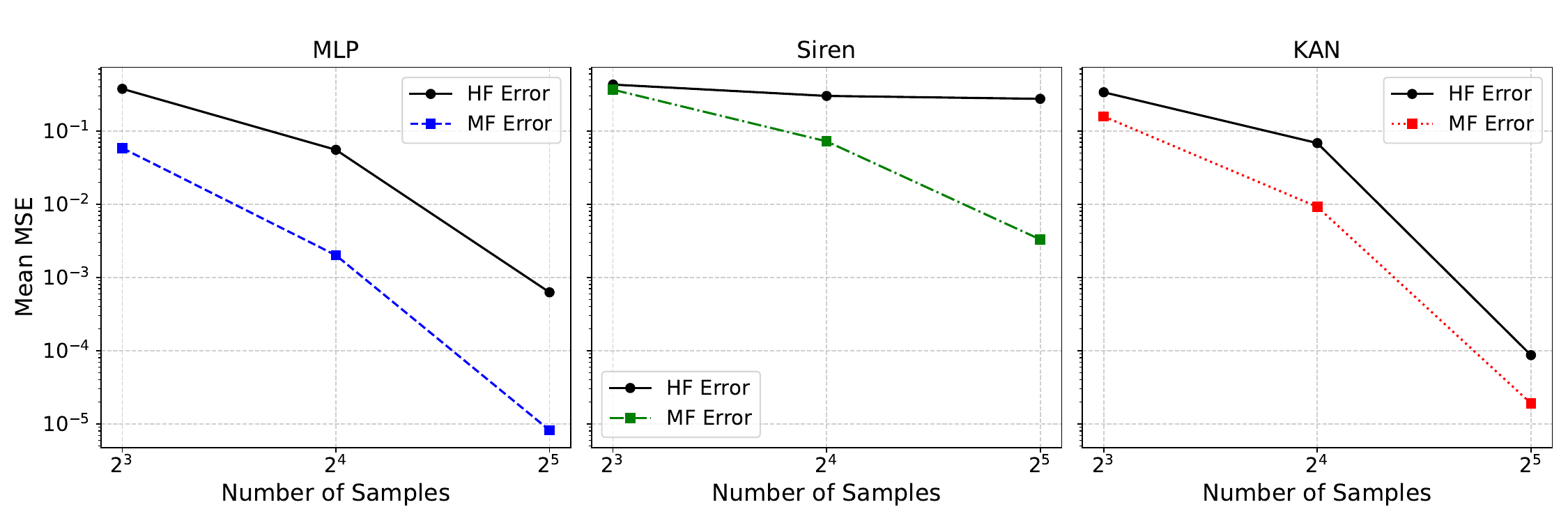}
    \includegraphics[width=0.25\linewidth, height=3.5cm]{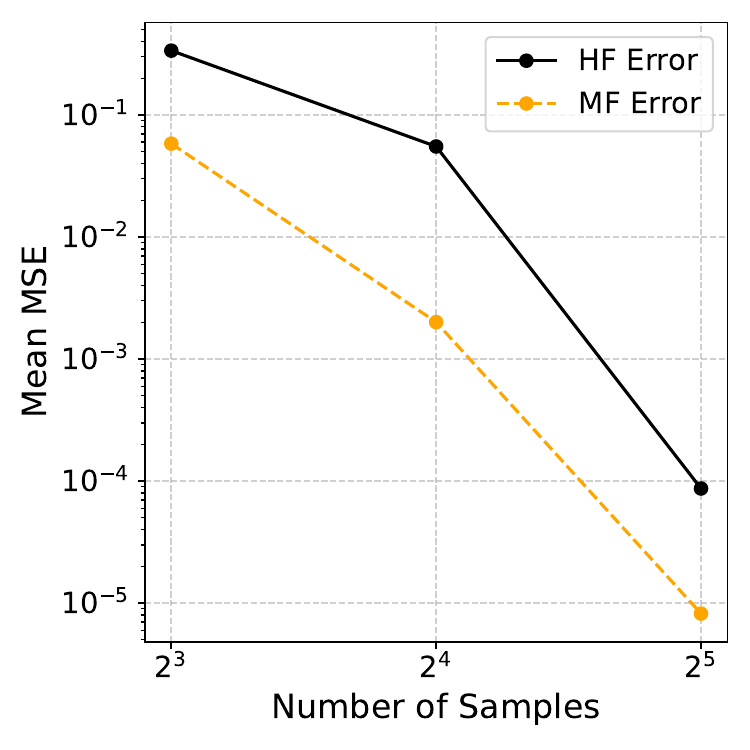}
    \caption{
    Mean MSE results of test case K3 comparing the best-performing single high-fidelity (HF) network and the best multi-fidelity network: results separated by network architecture across different HF sample sizes (left) and the best results at each HF sample size are shown, regardless of architecture type or network complexity (right).}
    \label{fig:MSE_Lines_k3}
\end{figure}

When comparing these multi-fidelity results with equivalent single-fidelity networks trained solely on HF data in Figure~\ref{fig:MSE_Lines_k3}, we observe distinct architecture-dependent behaviors. For MLP architectures, the multi-fidelity approach consistently outperforms single-fidelity networks, with the performance gap widening as sample size increases to 32, resulting in nearly two orders of magnitude lower normalized MSE. Siren networks show a more modest but consistent advantage for multi-fidelity networks across all sample sizes. In contrast, KAN architectures exhibit convergent behavior at higher sample counts, with multi-fidelity and single-fidelity approaches achieving comparable accuracy at 32 samples. This suggests that KANs' expressivity allows them to effectively capture oscillatory functions with sufficient high-fidelity data alone, while MLPs benefit substantially more from the additional information provided by low-fidelity models. These architecture-specific patterns highlight the varying degrees to which different network types can leverage multi-fidelity information for functions characterized by low-frequency oscillations.

% ==============================================================================
\subsubsection{1D Nonlinearly correlated phase shifted oscillatory model pair (K4)}
\label{sec:K4}
% ==============================================================================

\begin{figure}[ht!]
\centering
\includegraphics[width=0.9\linewidth]{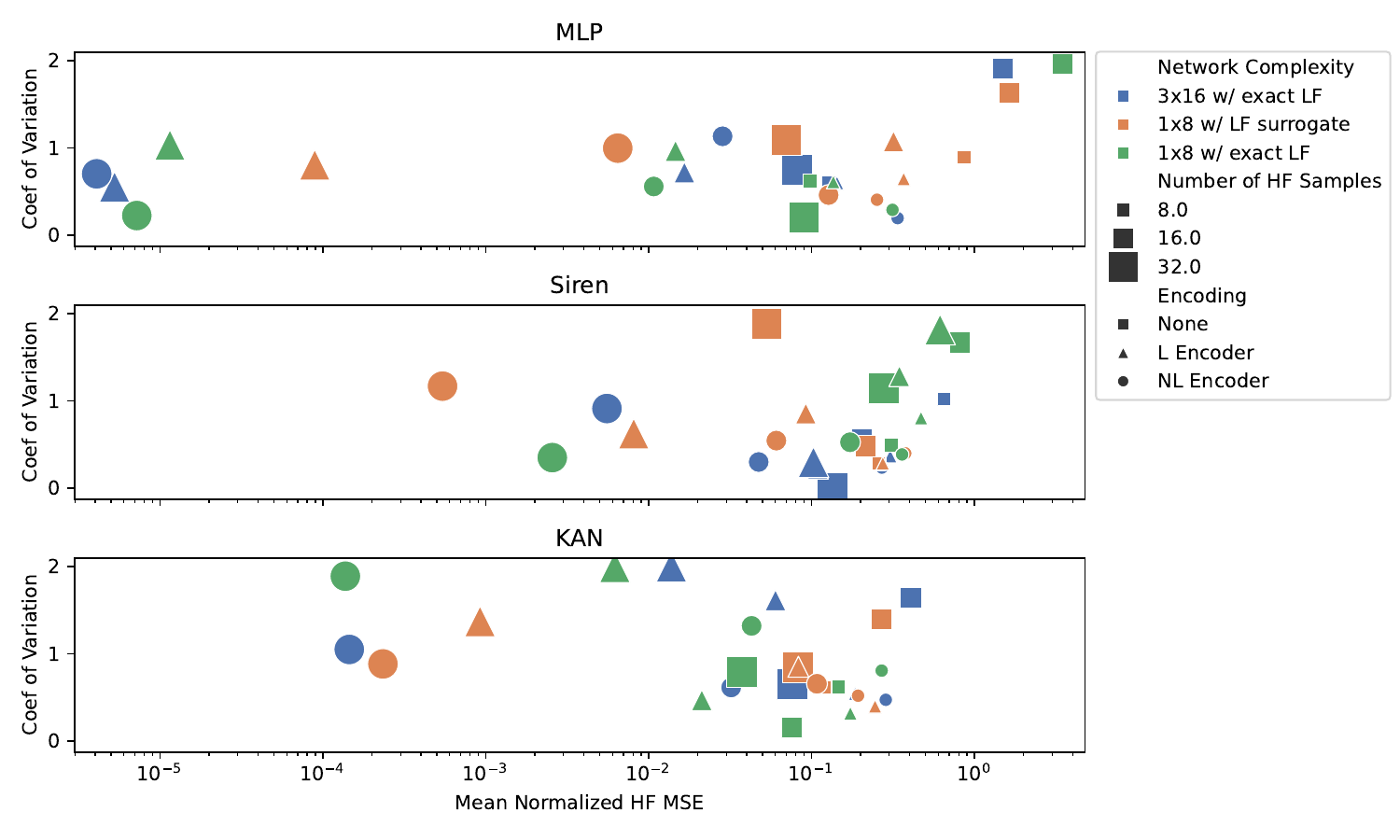}
\caption{Mean normalized HF MSEs of 4 repetitions for example K4 using various network configurations.}
\label{fig:K4_scatter}
\end{figure}

In this section we consider a model pair expressed as
\begin{equation}
\begin{aligned}
    y_L(x) =&
    \sin(8\pi x), & x \in [0,1],
    \\
    y_H(x) =&
    x^2 + y_L^2(x + 1/80), & x \in [0,1].
    % x^2 + \sin^2(8 \pi x + \pi/10)
\end{aligned}
\end{equation}
This example, similar to the previous, is characterized by an entirely nonlinear correlation between LF and HF model output, but also includes a phase shift. 
Without additional information or a sufficient number of HF samples, this slight change is known to make the HF function difficult to learn for MLPs~\cite{meng2020composite}. 
The results in Figure~\ref{fig:K4_scatter} show that using any encoder improves the MSE by several orders of magnitude, independent on the nature of the LF predictor (exact/learnable), with the MLP architectures benefiting particularly from this change.

\begin{figure}[ht!]
\centering
\includegraphics[width=0.32\linewidth]{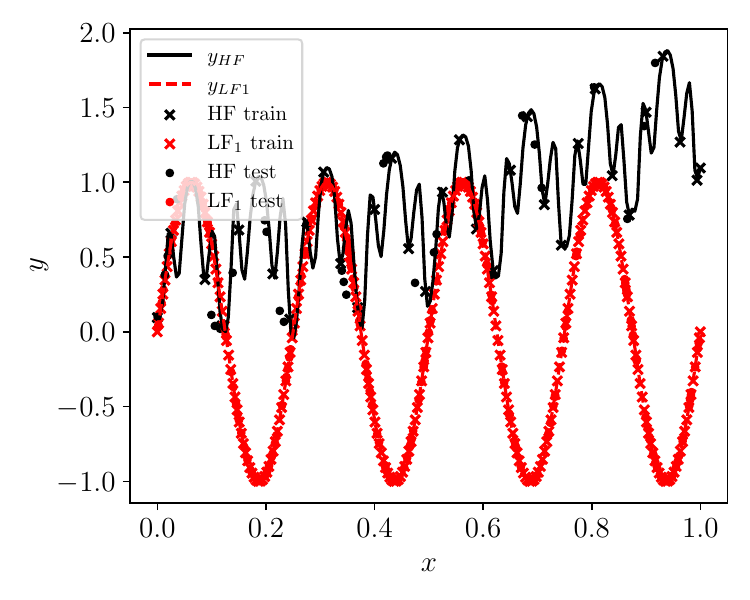}
\includegraphics[width=0.32\linewidth]{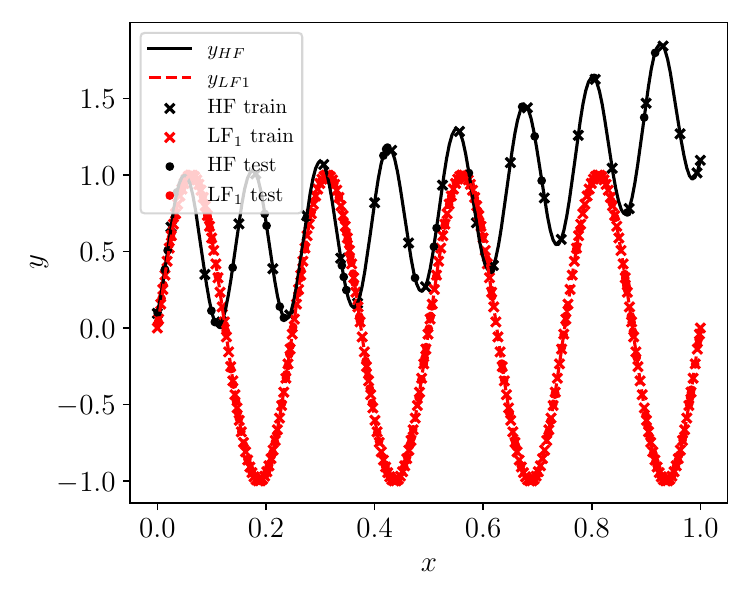}
\includegraphics[width=0.32\linewidth]{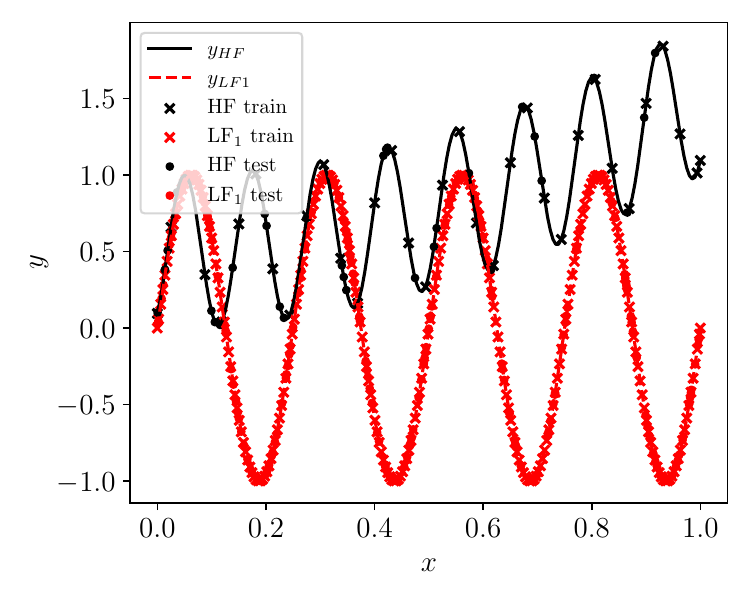}
\\
\includegraphics[width=0.32\linewidth]{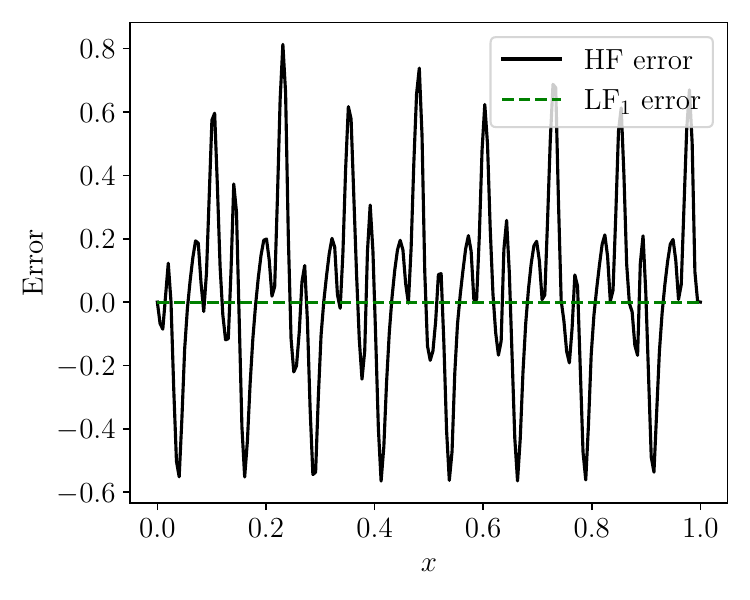}
\includegraphics[width=0.32\linewidth]{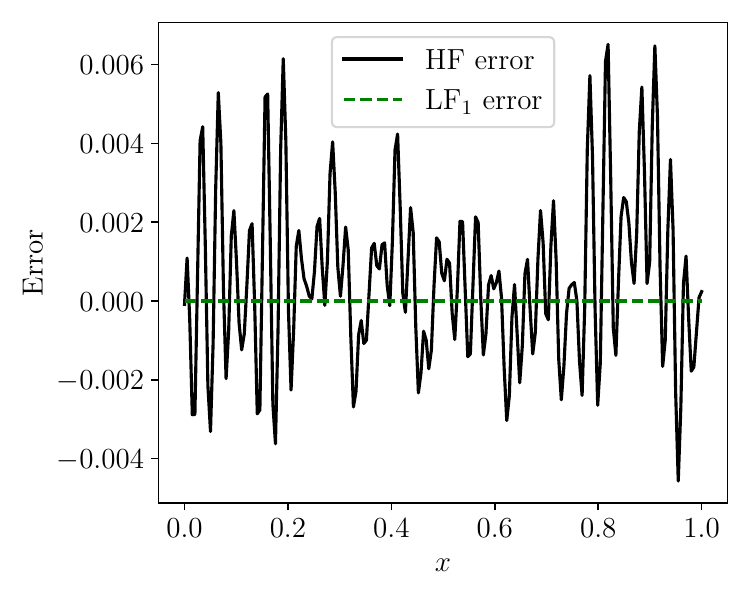}
\includegraphics[width=0.32\linewidth]{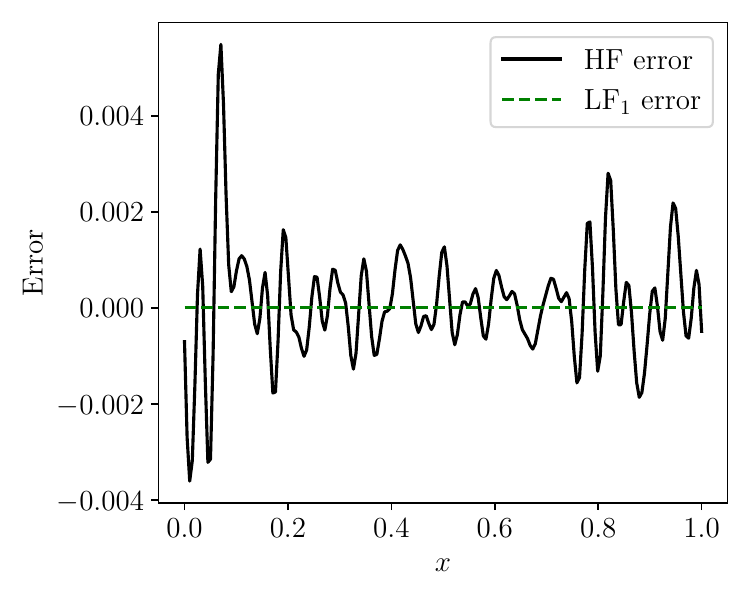}
\caption{Most accurate HF and LF predictors (top row) resulting from 32/256 HF/LF training samples, an exact LF, and a MLP architecture with no encoding (left), linear encoding (middle), and nonlinear encoding (right) for example K4. Error plots (bottom row) for each type of encoder.}
\label{fig:Best_MLP_K4_fits}
\end{figure}

\begin{figure}[ht!]
\centering
\includegraphics[width=0.32\linewidth]{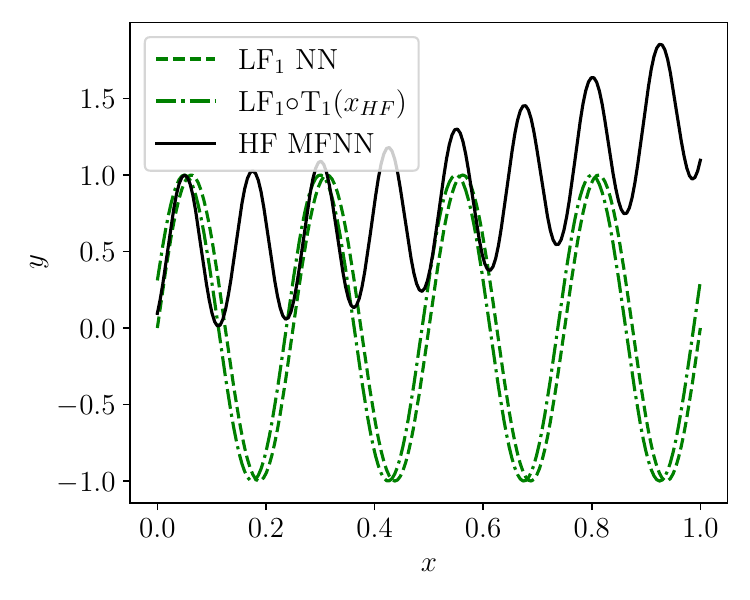}
\includegraphics[width=0.32\linewidth]{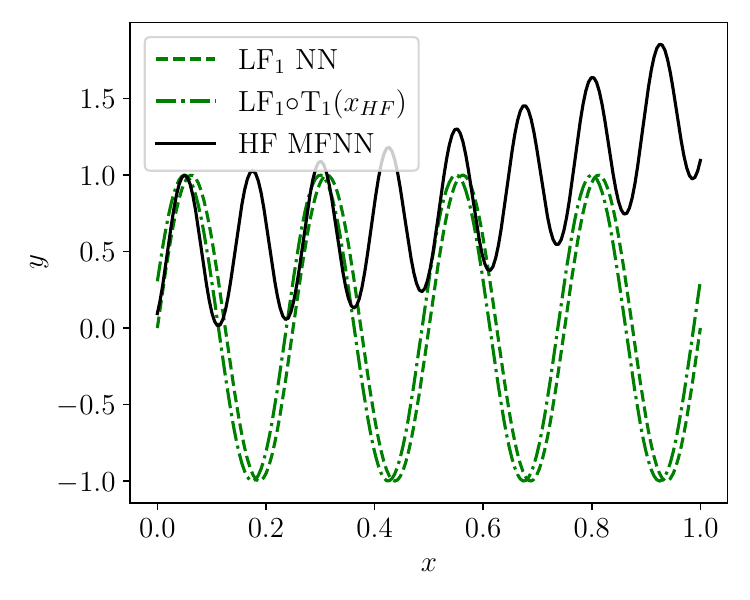}
\caption{LF function $y_{L} \circ T$ composed with linear (left) and nonlinear (right) coordinate encoding for example K4.}
\label{fig:Best_MLP_K4_enc}
\end{figure}

% Owen stated sinusodial activation networks work better with depth, not so much width. Since we're using a 3x16 network, it is a very wide network
Inspecting the best performing $3\times 16$ MLP networks using 32 HF samples and an exact LF, accurate HF model fits are only observed for encoded networks in Figure~\ref{fig:Best_MLP_K4_fits}, with the number of samples remaining an important factor.
Moreover, Figure~\ref{fig:Best_MLP_K4_enc} clearly shows that both the linear and nonlinear encoding networks managed to learn the phase shift discussed above, as expected.

\begin{figure}
    \centering

    \includegraphics[width=0.7\linewidth]{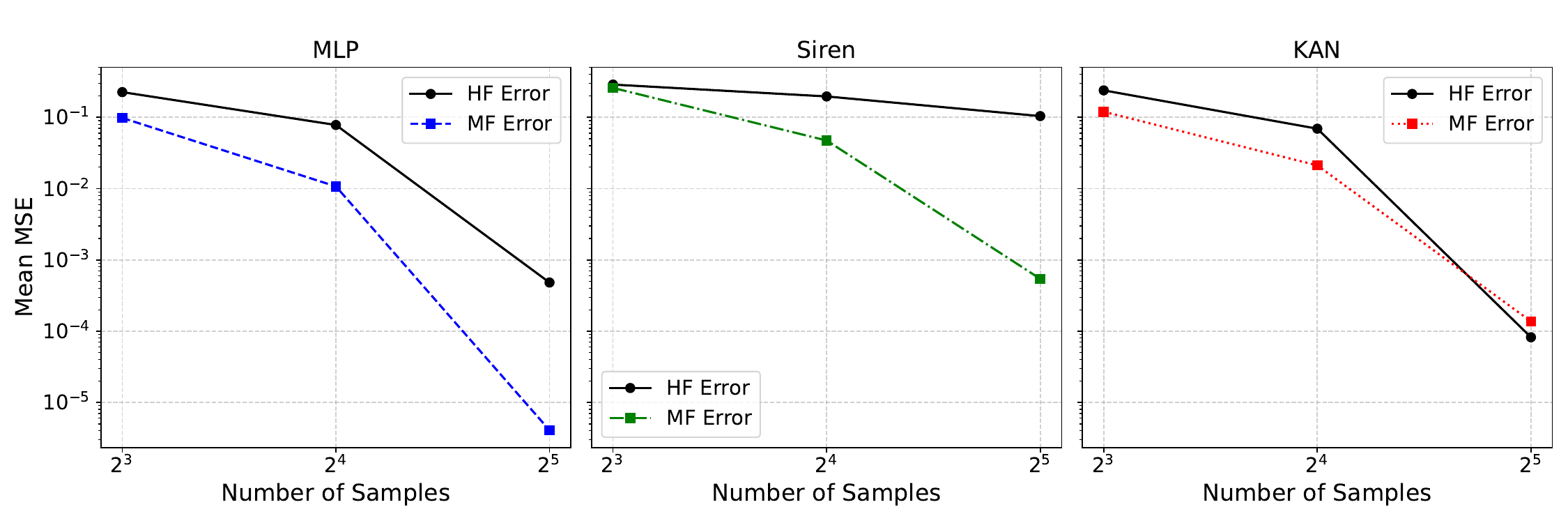}
    \includegraphics[width=0.25\linewidth, height=3.5cm]{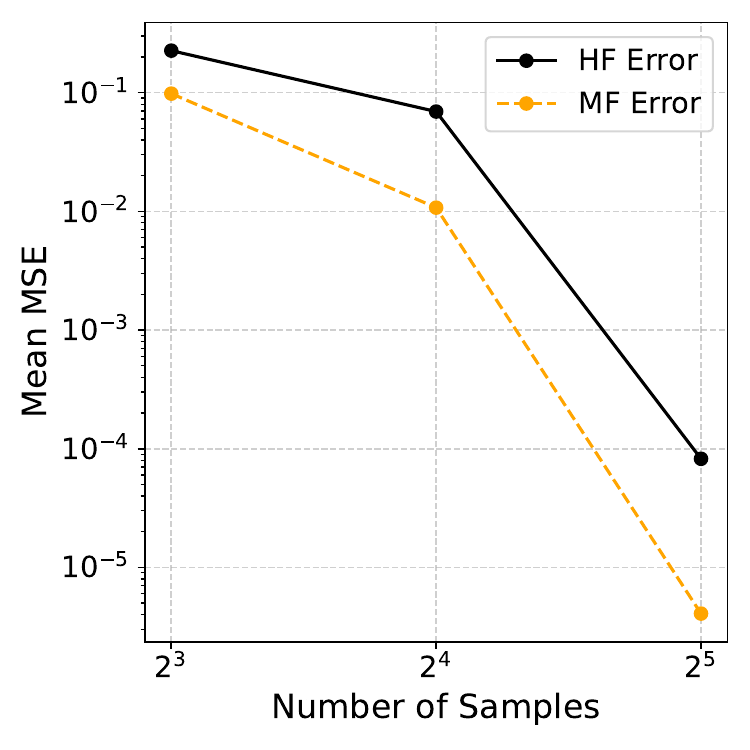}
    \caption{
    Mean MSE results of test case K4 comparing the best-performing single high-fidelity (HF) network and the best multi-fidelity network: results separated by network architecture across different HF sample sizes (left) and the best results at each HF sample size are shown, regardless of architecture type or network complexity (right).}
    \label{fig:MSE_Lines_k4}
\end{figure}

Comparing multi-fidelity results in Figure~\ref{fig:MSE_Lines_k4} of equivalent single-fidelity networks trained solely on HF data reveals architecture-specific patterns similar to those observed in test case K3, but with more pronounced differences. MLP architectures show the most dramatic advantage for multi-fidelity learning, with the gap between single-fidelity and MF approaches widening substantially as sample size increases to 32, resulting in more than two orders of magnitude lower normalized MSE. Siren networks maintain a consistent advantage for the multi-fidelity approach across all sample sizes, with the benefit becoming more pronounced at 32 samples. The KAN architecture again demonstrates convergent performance at higher sample counts, with multi-fidelity and single-fidelity approaches reaching similar accuracy levels at 32 samples. These results further confirm that while MLPs significantly benefit from coordinate encoding to capture the phase-shifted relationship between LF and HF data, KANs possess sufficient flexibility to learn these transformations directly from HF data alone when provided with adequate samples. This advantage of KANs becomes particularly relevant in scenarios where collecting additional data might be challenging or impractical.

% =======================================================================
\subsection{One dimensional test cases with multiple low-fidelity models}
% =======================================================================

In this section, we address the problem of learning the HF function from test case K4 using multiple LF predictors. 
Each subsection below explores a different situation of practical relevance. In the first scenario, both LF models have equal correlation with the HF, and we examine how the MF network discriminates among which information to propagate. 
In the second scenario, only one LF model is correlated with the HF, while the other is uncorrelated. 
Finally, the third scenario considers the case where both LF models are partially correlated, requiring the correlation network to selectively utilize information from both predictors on relevant input sub-domains. 

% ==============================================================================
\subsubsection{Identical nonlinearly correlated LF (K4D)}\label{sec:K4 double}
% ==============================================================================

\begin{figure}[ht!]
\centering
\includegraphics[width=0.9\linewidth]{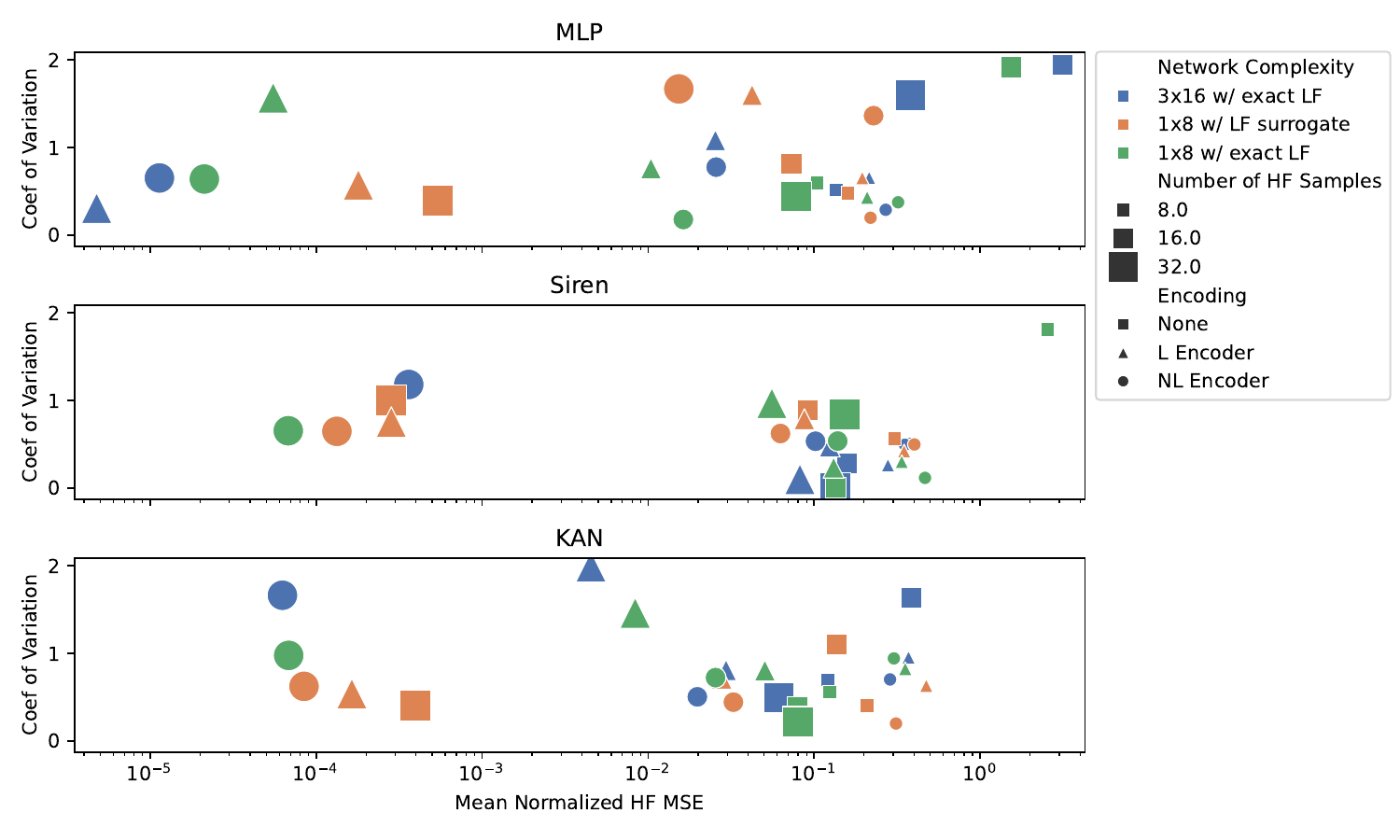}
\caption{Mean normalized HF MSE over 4 repetitions for example K4D, using various network configurations.}
\label{fig:K4_double_scatter}
\end{figure}

Consider the model triplet
\begin{equation}
\begin{aligned}
    y_{L_1}(x) &= \sin(8\pi x), & x \in [0,1],
    \\
    y_{L_2}(x) &= \sin(8\pi x), & x \in [0,1],
    \\
    \text{and}\,
    y_H(x)&=
    % x^2 + y_L^2(x + 1/80), & x \in [0,1].
    x^2 + \sin^2(8 \pi x + \pi/10), & x \in [0,1].
\end{aligned}
\end{equation}
Since the two LF models are identical, we expect a similar accuracy as obtained in Section~\ref{sec:K4} as no new information is being provided by the second LF model.
However, from the results in Figure \ref{fig:K4_double_scatter}, we do see some differences.
Although we see similar results with and without encoding, the addition of a second LF model leads to an increase of accuracy, with more network configurations reaching a  normalized MSE below $10^{-3}$, even when training an LF emulator. 
Using two identical LFs results in a bimodal accuracy distribution independent on the network type (MLP, Siren or KAN), concentrated around $10^{-1}$ and $10^{-4}$, respectively.
The number of samples seems to be the main factor leading to MSE in the order $10^{-4}$, indicating a \emph{positive reinforcement} related to using the same LF twice.

\begin{figure}
    \centering
    \includegraphics[width=0.7\linewidth]{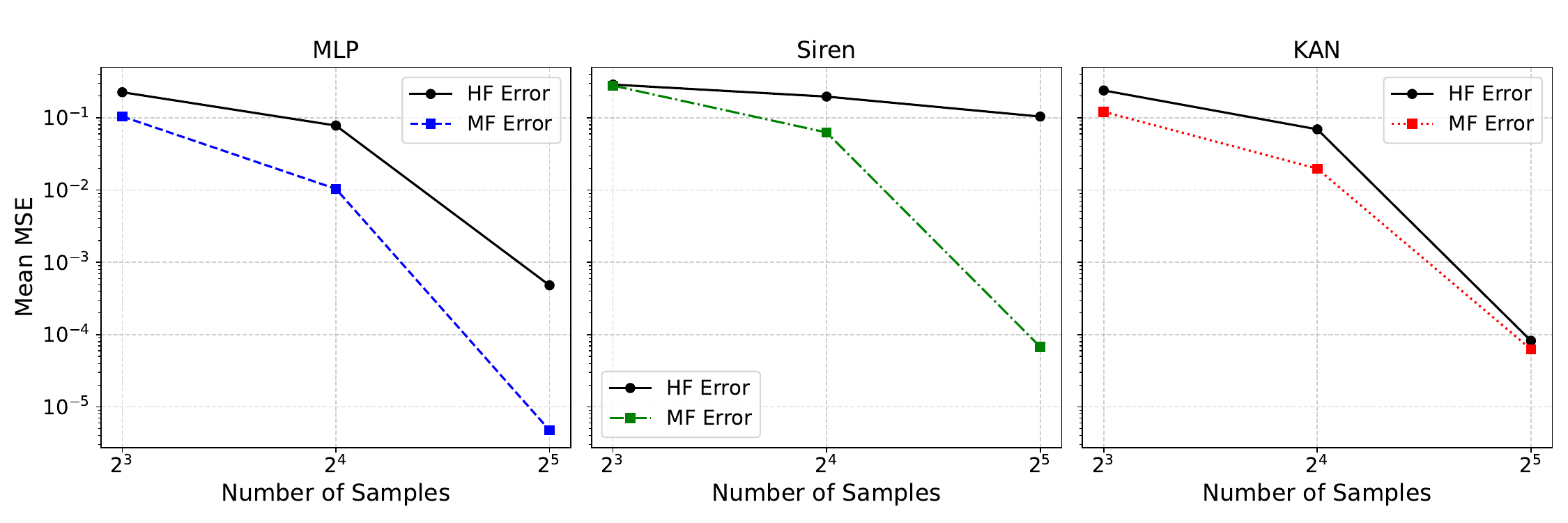}
    \includegraphics[width=0.25\linewidth, height=3.5cm]{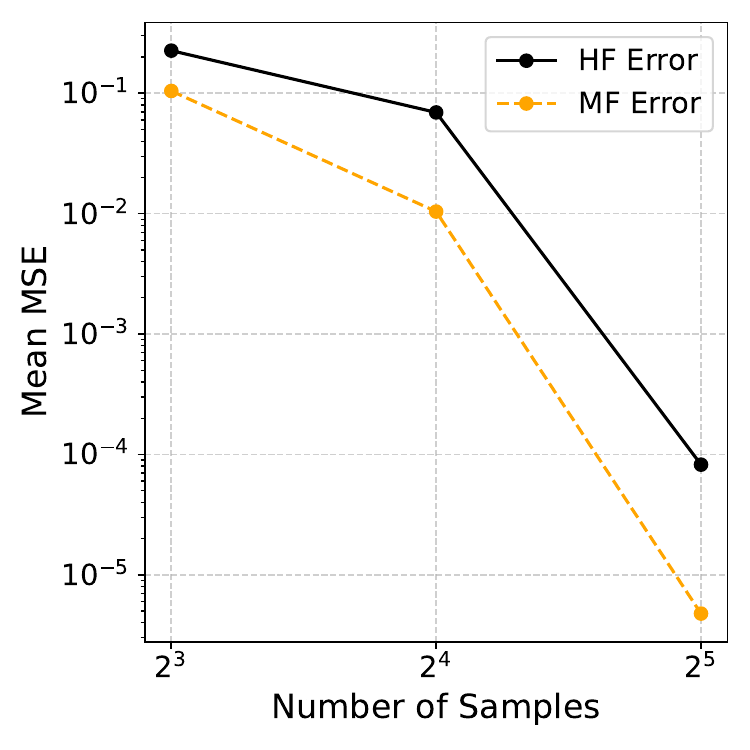}
    \caption{
    Mean MSE results of test case K4D comparing the best-performing single high-fidelity (HF) network and the best multi-fidelity network: results separated by network architecture across different HF sample sizes (top) and the best results at each HF sample size are shown, regardless of architecture type or network complexity (bottom).}
    \label{fig:MSE_Lines_k4_double}
\end{figure}

Figure~\ref{fig:MSE_Lines_k4_double} demonstrates consistent multi-fidelity advantages across all architectures, with Siren networks showing the most dramatic benefit of nearly three orders of magnitude between HF and MF approaches at maximum sampling, while KAN architectures exhibit convergent behavior where both pathways achieve comparable accuracy near $10^{-4}$ at $2^5$ samples.

% ==============================================================================
\subsubsection{One nonlinearly correlated and one uncorrelated LF (K4U)}
% Include at least one uncorrelated LF model
% ==============================================================================

\begin{figure}[ht!]
\centering
\includegraphics[width=\linewidth]{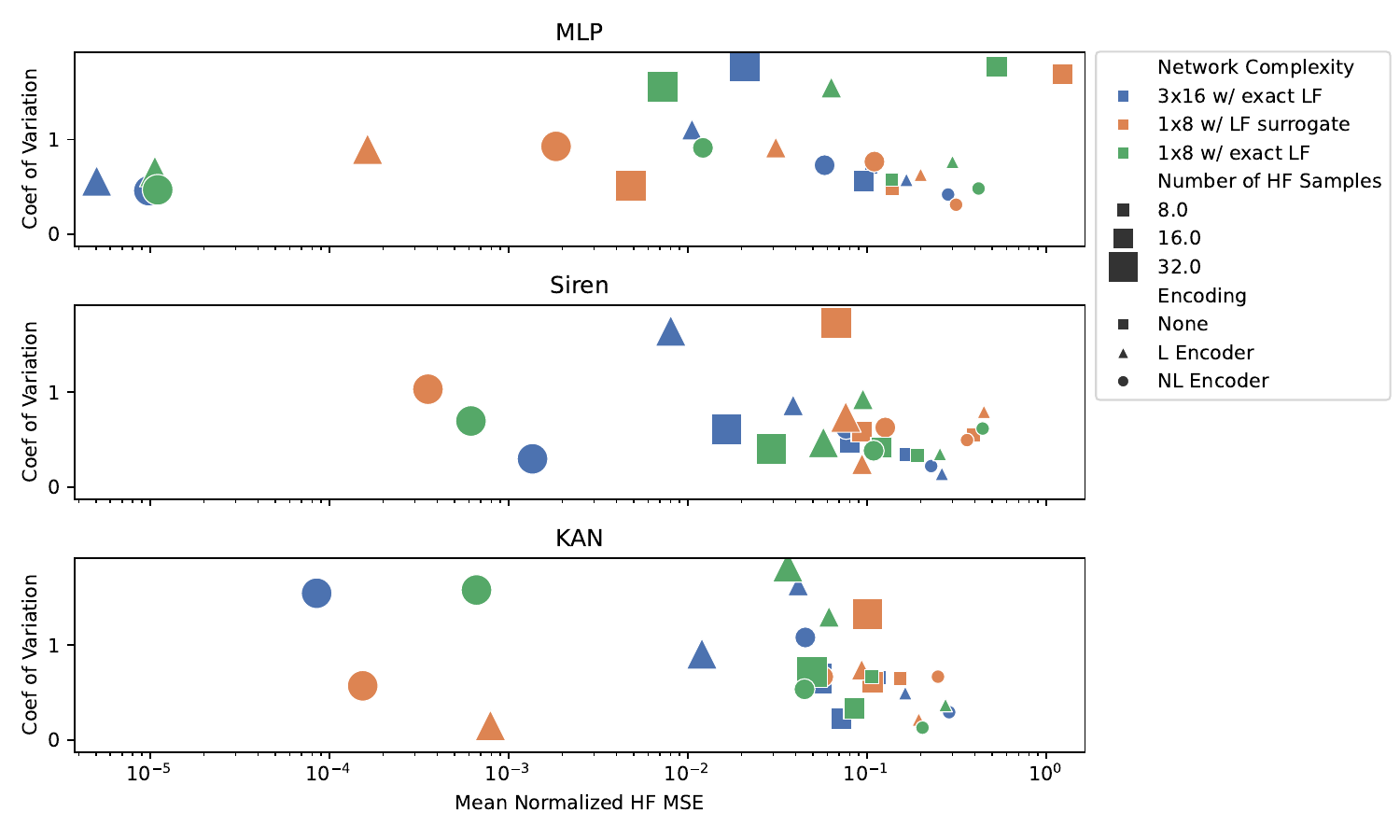}
\caption{Mean normalized HF MSE over 4 repetitions for example K4U, using various network configurations.}
\label{fig:K4_unrelated_scatter}
\end{figure}

We now consider the model triplet
\begin{equation}
\begin{aligned}
    y_{L_1}(x) &= \sin(8\pi x), & x \in [0,1],
    \\
    y_{L_2}(x) &= 0.5(6 x-2)^2 \sin (12 x-4)+10x-10, & x \in[0,1],
    \\
    \text{and}\,
    y_H(x)&=
    % x^2 + y_L^2(x + 1/80), & x \in [0,1].
    x^2 + \sin^2(8 \pi x + \pi/10), & x \in [0,1].
\end{aligned}
\end{equation}
In this example, we use one correlated LF ($y_{L_1}$) and one uncorrelated LF ($y_{L_2}$) to verify the ability of the MF network to discard irrelevant or confusing information.
Comparing the results in Figure~\ref{fig:K4_unrelated_scatter} and Figure~\ref{fig:K4_scatter}, we see an overall similar distribution of normalized MSE, with encoded networks generally performing better than standard networks. 
Addition of an uncorrelated LF model $y_{L_2}$ does not seem to disrupt the predictive capability of a MF emulator.

\begin{figure}
    \centering

    \includegraphics[width=0.7\linewidth]{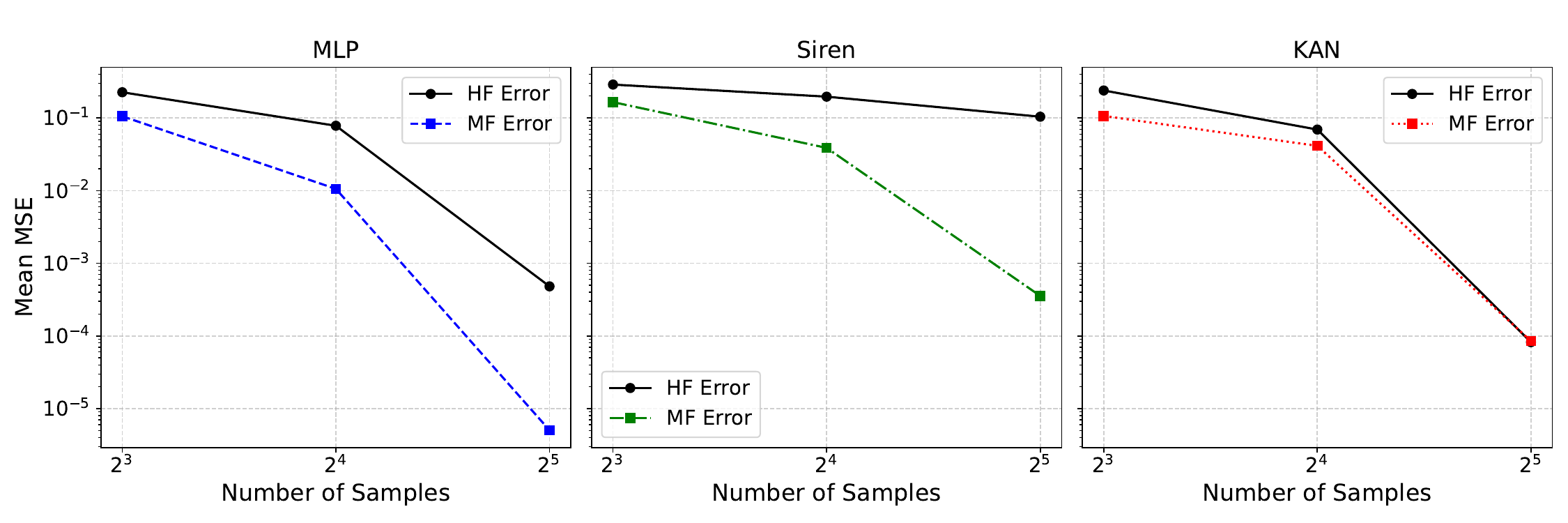}
    \includegraphics[width=0.25\linewidth, height=3.5cm]{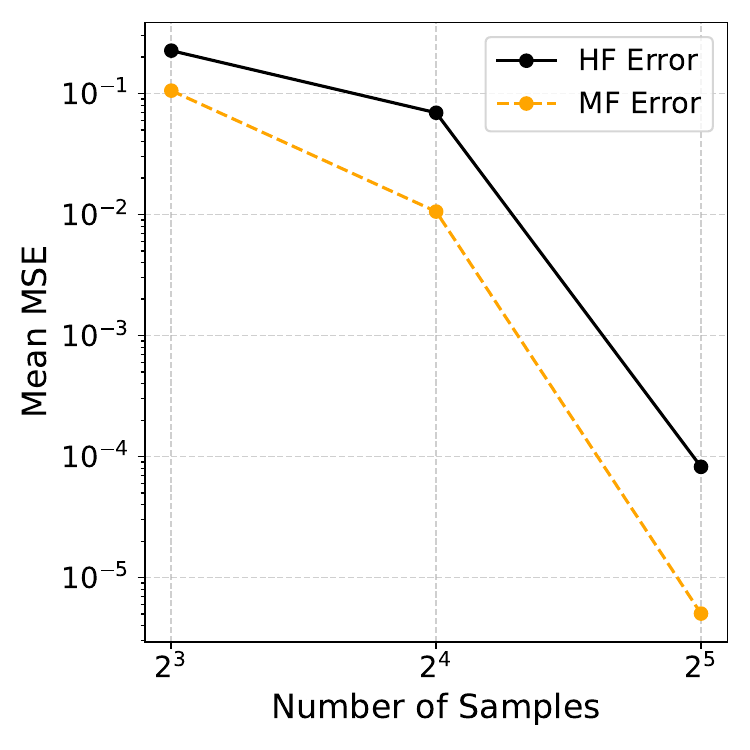}
    \caption{
    Mean MSE results of test case K4U comparing the best-performing single high-fidelity (HF) network and the best multi-fidelity network: results separated by network architecture across different HF sample sizes (top) and the best results at each HF sample size are shown, regardless of architecture type or network complexity (bottom).}
    \label{fig:MSE_Lines_k4_unrelated}
\end{figure}

Results in figure~\ref{fig:MSE_Lines_k4_unrelated} closely mirror the results in figure~\ref{fig:MSE_Lines_k4}, with all architectures achieving quantitatively similar error magnitudes and convergence rates, suggesting that this alternative set of low-fidelity model captures comparable information content for multi-fidelity learning.

% ==================================================
\subsubsection{LFs with piecewise correlation (K4P)}
% ==================================================

\begin{figure}[ht!]
\centering
\includegraphics[width=\linewidth]{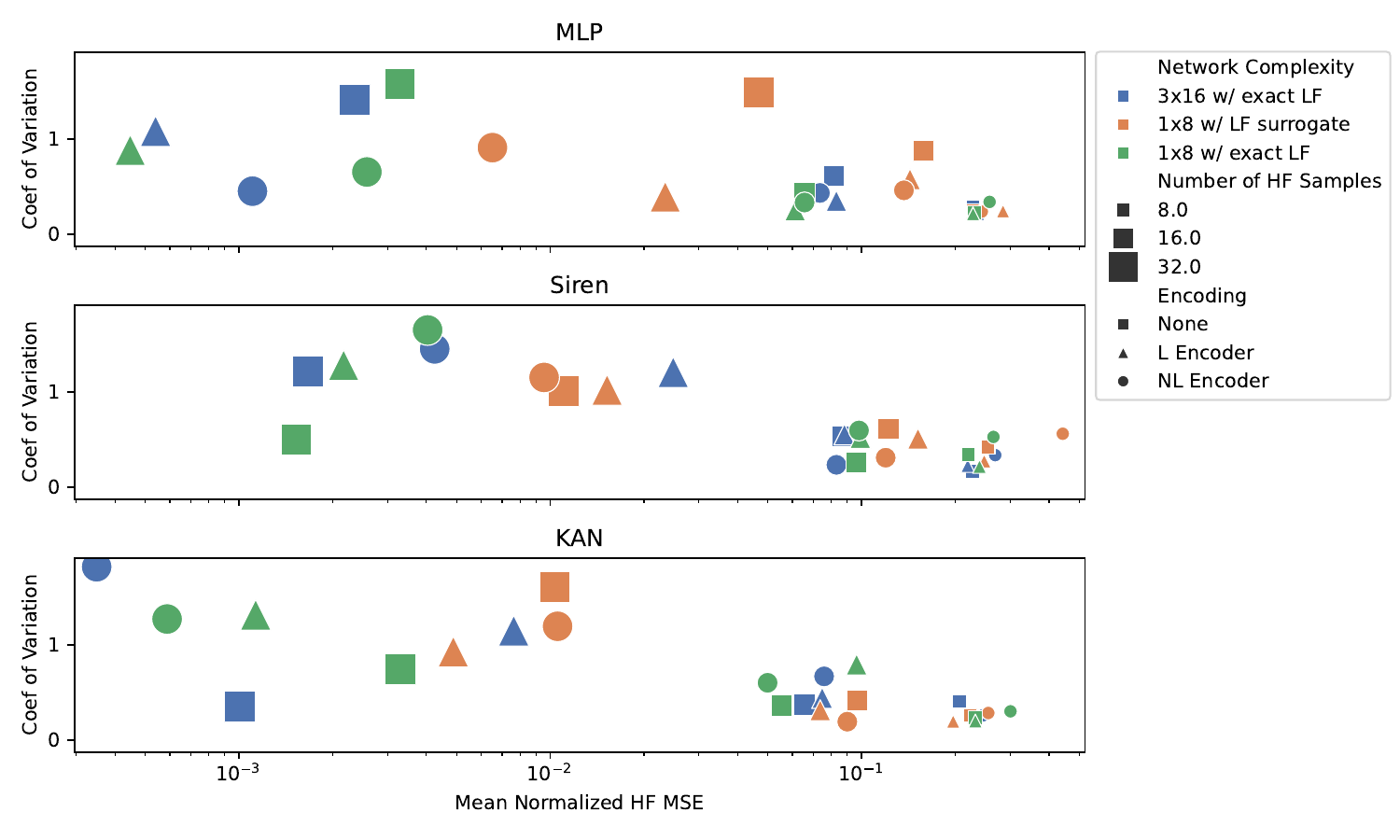}
\caption{Mean normalized HF MSE over 4 repetitions for example K4P, using various network configurations.}
\label{fig:K4_piecewise_scatter}
\end{figure}

Let $K_1(x) = 0.5(6 x-2)^2 \sin (12 x-4)+10x-10$ and $K_3(x) = \sin(8\pi x)$. Then consider the model triplet
\begin{equation}
\begin{aligned}
    y_{L_1}(x) &=
    \begin{cases}
        K_1(x), & x \in [0,0.5], \\
        K_3(x), & x \in (0.5,1],
    \end{cases}
    \\
    y_{L_2}(x) &=
    \begin{cases}
        K_3(x), & x \in [0,0.5], \\
        K_1(x), & x \in (0.5,1],
    \end{cases}
    \\
    \text{and}\,
    y_H(x) &= x^2 + \sin^2(8 \pi x + \pi/10), & x \in [0,1].
\end{aligned}
\end{equation}
In this example, we consider two LF models, both correlated only within a subdomain and uncorrelated otherwise. However, there is at least one subdomain that contains a correlated LF. 
In particular, $y_{L_1}$ is correlated with $y_{H}$ over $(0.5,1]$ and uncorrelated otherwise, while $y_{L_2}$ is correlated with $y_{H}$ over $[0,0.5]$ and uncorrelated otherwise.
Therefore, this example tests the networks' ability to selectively extract information from correlated LFs on input subdomains.
Comparing the results in Figure~\ref{fig:K4_piecewise_scatter} and Figure~\ref{fig:K4_scatter}, we observe slightly less accurate results (relative MSE greater than $10^{-4}$) with stronger dependence on the number of samples.

\begin{figure}
    \centering
    \includegraphics[width=0.7\linewidth]{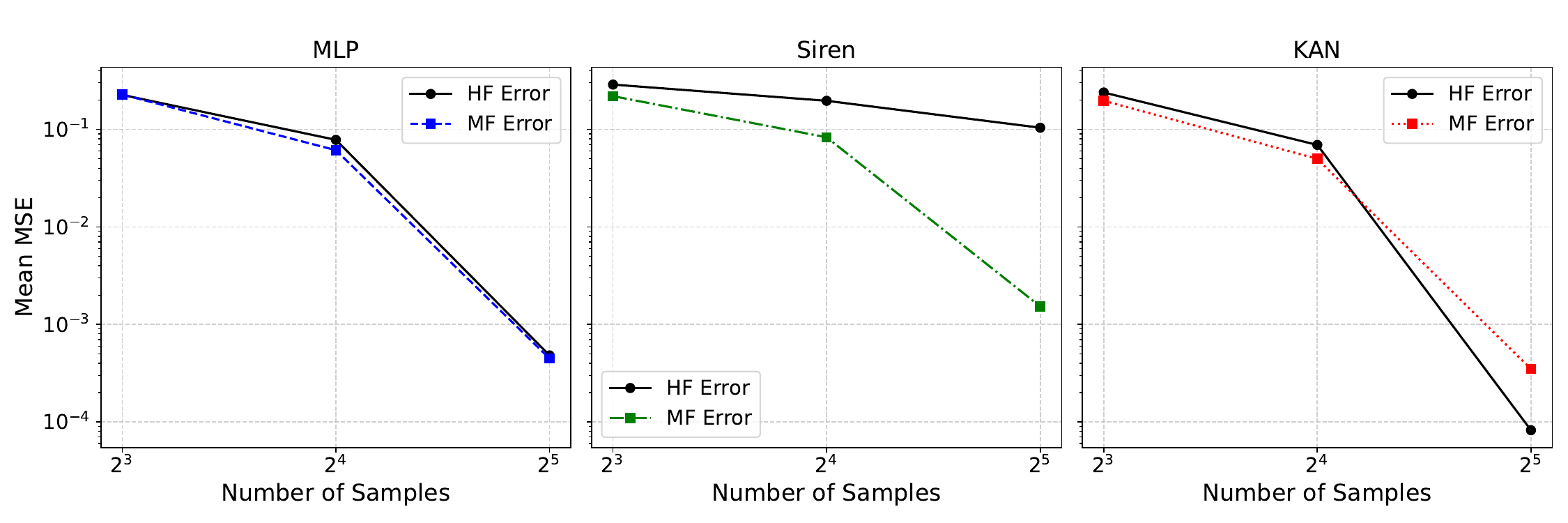}
    \includegraphics[width=0.25\linewidth, height=3.5cm]{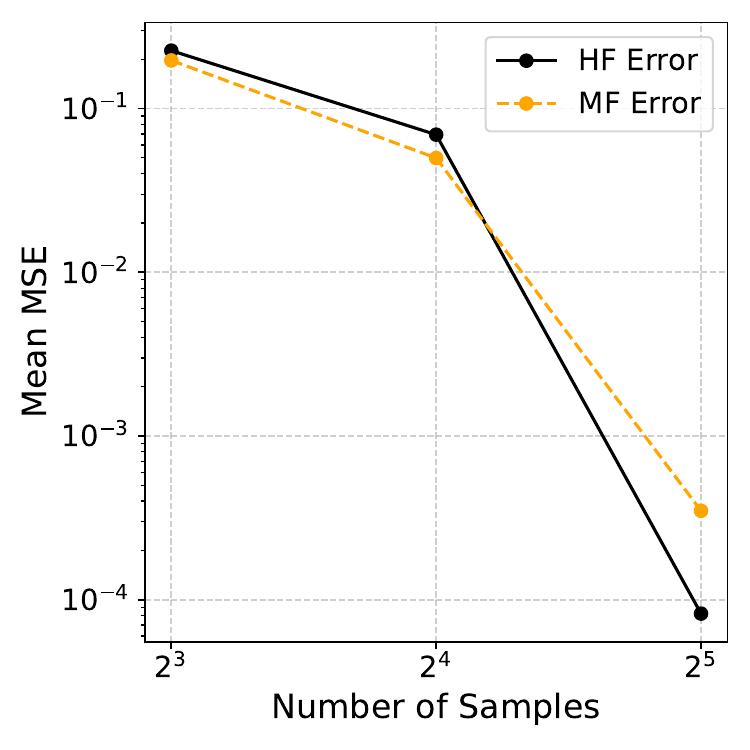}
    \caption{
    Mean MSE results of test case K4P comparing the best-performing single high-fidelity (HF) network and the best multi-fidelity network: results separated by network architecture across different HF sample sizes (top) and the best results at each HF sample size are shown, regardless of architecture type or network complexity (bottom).}
    \label{fig:MSE_Lines_k4_piecewise}
\end{figure}

Figure~\ref{fig:MSE_Lines_k4_piecewise} exhibits distinct behavior from the other cases, with MLP networks showing convergence between MF and HF pathways at equivalent sample counts, suggesting that learning selective subdomain correlations diminishes the multi-fidelity advantage for this architecture. Siren networks continue to benefit substantially from multi-fidelity data, maintaining their characteristic performance gap between approaches. KAN architectures ultimately achieve the lowest absolute errors below $10^{-4}$, with all architectures demonstrating stronger sample dependence than in the K4D and K4U cases, consistent with the increased difficulty of extracting information from partially correlated low-fidelity models. Notably, HF-only networks outperform their MF counterparts at $2^5$ samples for certain architectures, indicating that at sufficient sample sizes, the added complexity of learning piecewise correlations may hinder rather than help performance.

% =====================================
\subsection{Two dimensional test cases}
% =====================================

% =======================================================
\subsubsection{Two-dimensional model pair with equal parameterization (2DE)}
% =======================================================

\begin{figure}[ht!]
\centering
\includegraphics[trim={0 0 0 2.5em},clip,width=0.32\linewidth]{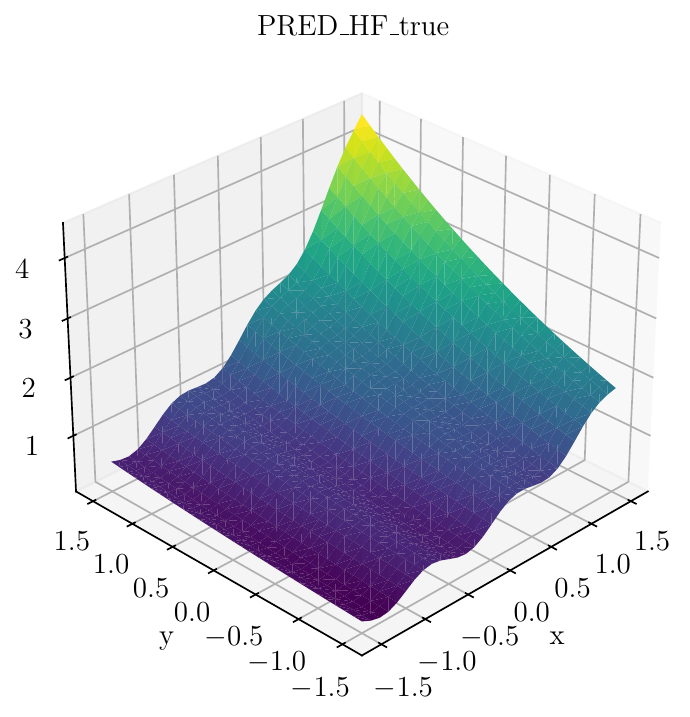}
\includegraphics[trim={0 0 0 2.5em},clip,width=0.32\linewidth]{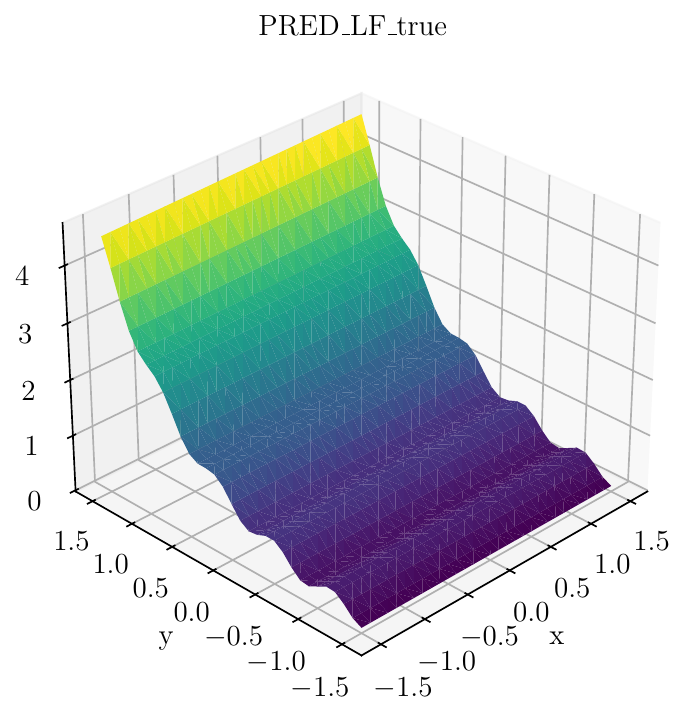}
\caption{Exact target HF (left) and LF (right) models.}
\label{fig:2d_equal_exact_func}
 \end{figure}

We now consider a pair of two-dimensional models expressed as
\begin{equation}
\label{eq:2D_equal}
\begin{aligned}
    y_L(x_1,x_2) =& \exp(0.01 x_1 + 0.99 x_2) + 0.15\sin(3\pi x_2), & x_1, x_2 \in [-1.5,1.5]
    ,\\
    y_H(x_1,x_2) =& \exp(0.7 x_1 + 0.3 x_2) + 0.15\sin(3\pi x_1), & x_1, x_2 \in [-1.5,1.5].
\end{aligned}
\end{equation}
A surface plot of the two models is also shown in Figure~\ref{fig:2d_equal_exact_func}.
These models are both characterized by a an increasing trend, albeit in two orthogonal directions, and by secondary oscillations at different frequencies. In addition, they are related through a linear change of variable $y_H = y_L \circ T^*$, where
\begin{equation}
    T^*(x_1, x_2)
    =
    \begin{pmatrix}
        0 & 1.5
        \\
        1/30 & -0.2
    \end{pmatrix}.
\end{equation} 
As a result, we expect a linear encoding to induce a linear correlation between the two models (actually to make them identical), leading to an accurate MF emulator with minimal nonlinear correlation.

\begin{figure}[ht!]
\centering
 \includegraphics[width=\linewidth]{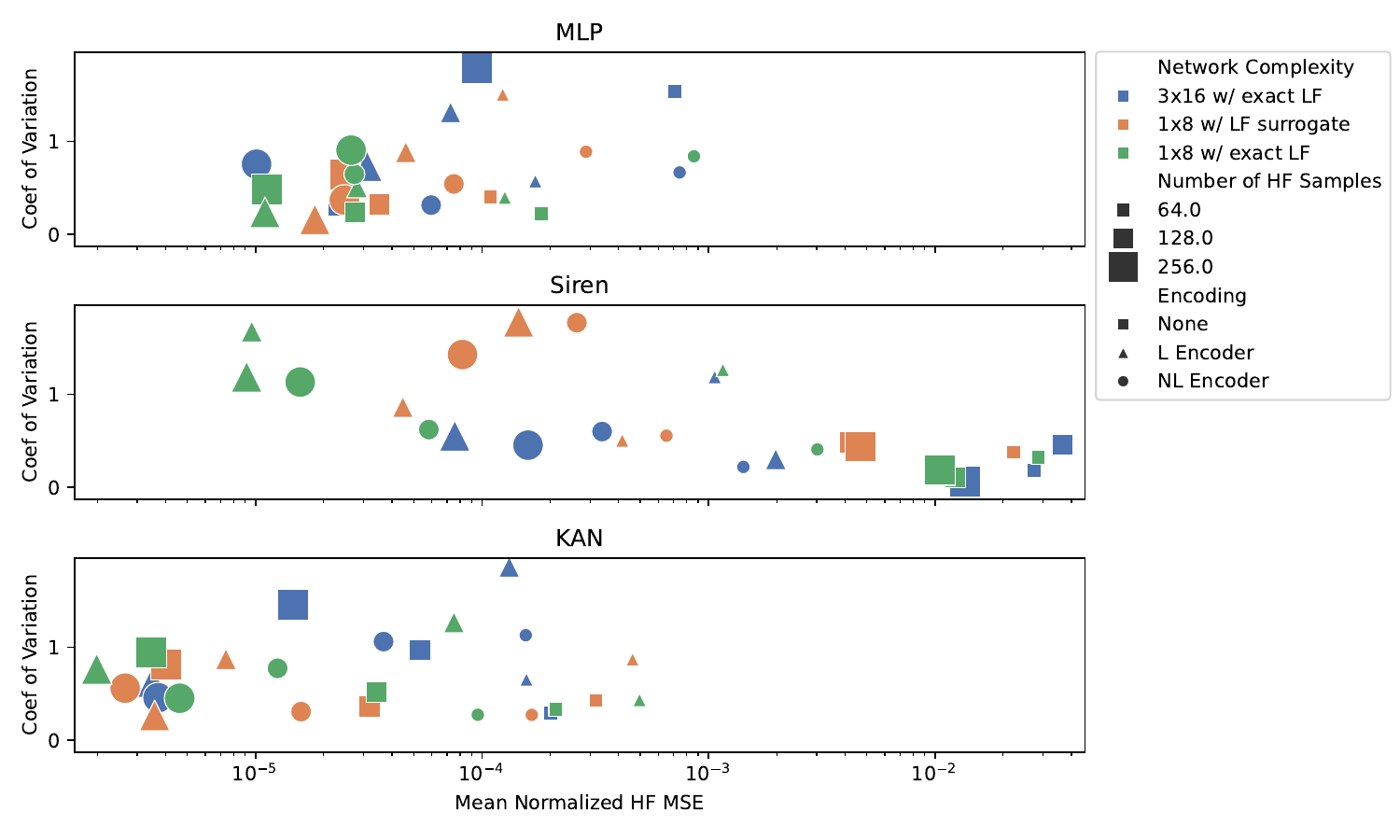}
 \caption{Mean normalized HF MSE over 4 repetitions for example 2DE, using various network configurations.}
\label{fig:2d_equal_scatter_plot}
 \end{figure}

Figure~\ref{fig:2d_equal_scatter_plot} shows that all network instances are able to train accurate HF emulators, particularly KANs, followed by MLPs and Siren. Moreover, satisfactory accuracy can be achieved independent of model complexity and both for an exact and trainable LF response.

\begin{figure}[ht!]
\centering
\includegraphics[trim={0 0 0 2.5em},clip,width=0.24\linewidth]{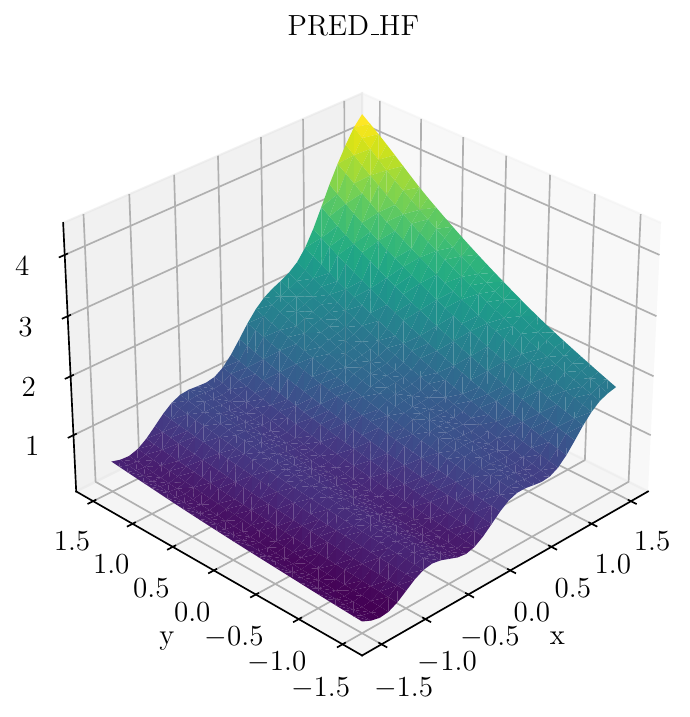}
\includegraphics[trim={0 0 0 2.5em},clip,width=0.24\linewidth]{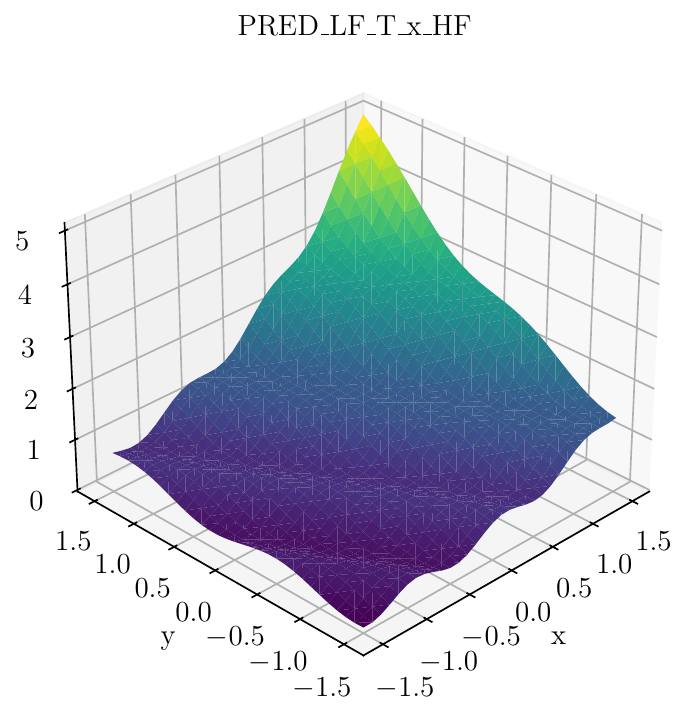}
\includegraphics[trim={0 0 0 2.5em},clip,width=0.24\linewidth]{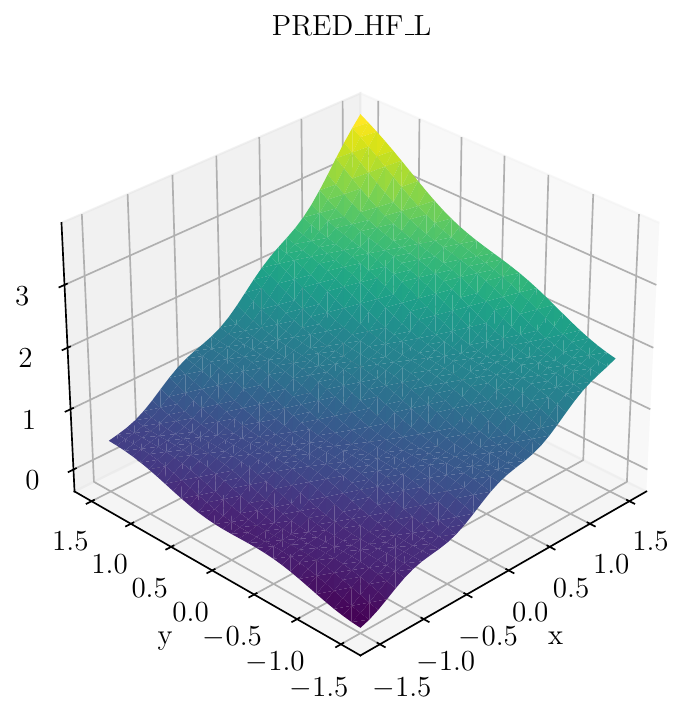}
\includegraphics[trim={0 0 0 2.5em},clip,width=0.24\linewidth]{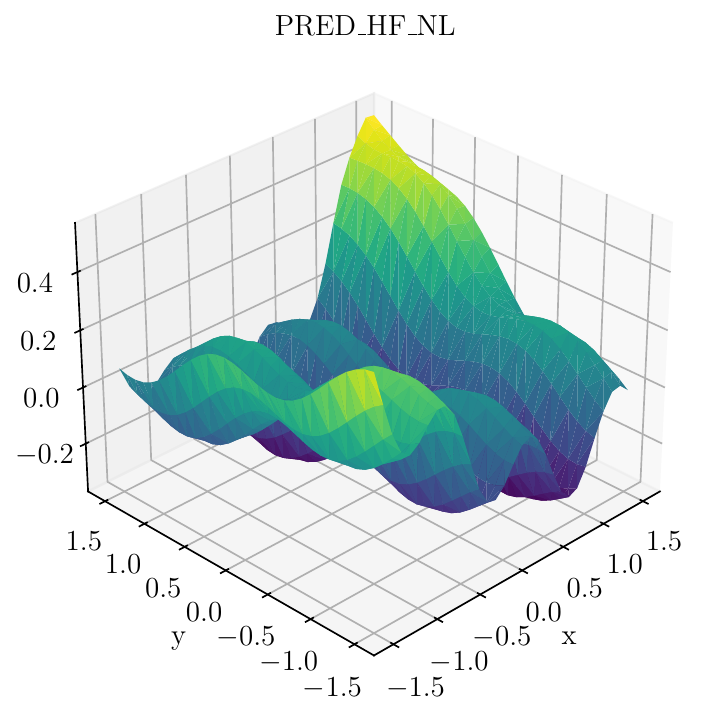}
\caption{Best model fit for test case 2DE, resulting from 256/2048 HF/LF training samples, and an exact LF model for a KAN architecture with linear encoding. The figure shows the HF model (left), the LF function composed with its encoding $\widehat{y}_{L} \circ T$ (left center), the HF linear correlation $\widehat{y}_{H}^{l}$ (right center), and the HF nonlinear correlation $\widehat{y}_{H}^{nl}$ (right).}
\label{fig:2d_equal_l_en_fit}
\end{figure}

Inspecting the best-performing network in Figure~\ref{fig:2d_equal_l_en_fit}, a KAN using linear encoding, reveals that, although the HF and LF models have different axes of variation, the encoder learns to rotate the LF function to better align with the target HF function. Upon further inspection of the correlation learned, we observe that the linear correlation captures the general shape of the HF function, while the nonlinear correlation needs only to learn small corrective features. The sum of these components results in a very accurate representation of the HF function.

\begin{figure}
    \centering
    \includegraphics[width=0.7\linewidth]{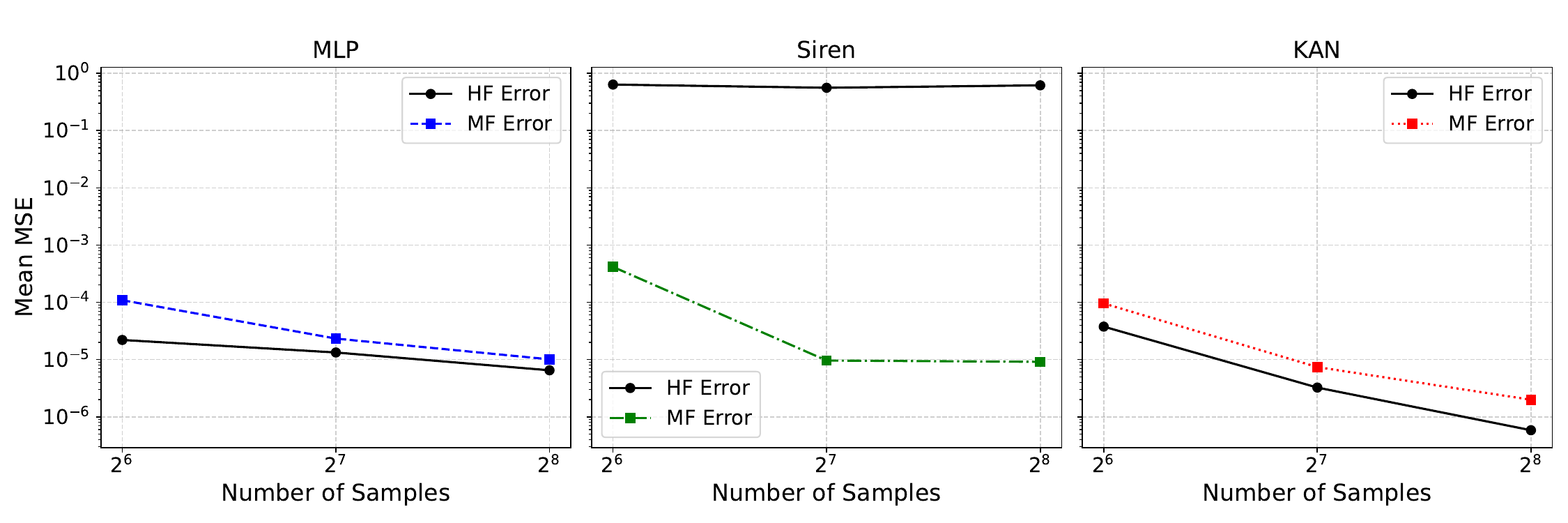}
    \includegraphics[width=0.25\linewidth, height=3.5cm]{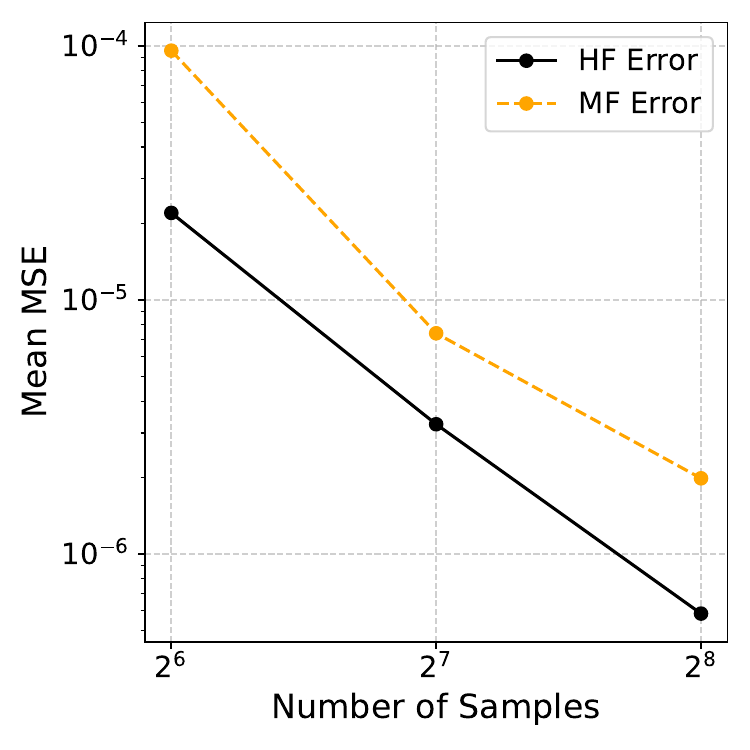}
    \caption{
    Mean MSE results of test case 2DE comparing the best-performing single high-fidelity (HF) network and the best multi-fidelity network: results separated by network architecture across different HF sample sizes (left) and the best results at each HF sample size are shown, regardless of architecture type or network complexity (right).}
    \label{fig:MSE_Lines_2d_equal}
\end{figure}

When comparing these multi-fidelity and single-fidelity networks results in Figure~\ref{fig:MSE_Lines_2d_equal}, we see that both MLP and KAN architectures are sufficiently flexible to learn the nonlinear correlation between LF and HF without help from a coordinate encoding. Therefore the benefit of having additional information from a low-fidelity model become almost negligible.

% ==================================================
\subsubsection{Two-dimensional model pair with dissimilar parameterization (2DU)}
% ==================================================

\begin{figure}[ht!]
\centering
\includegraphics[width=\linewidth]{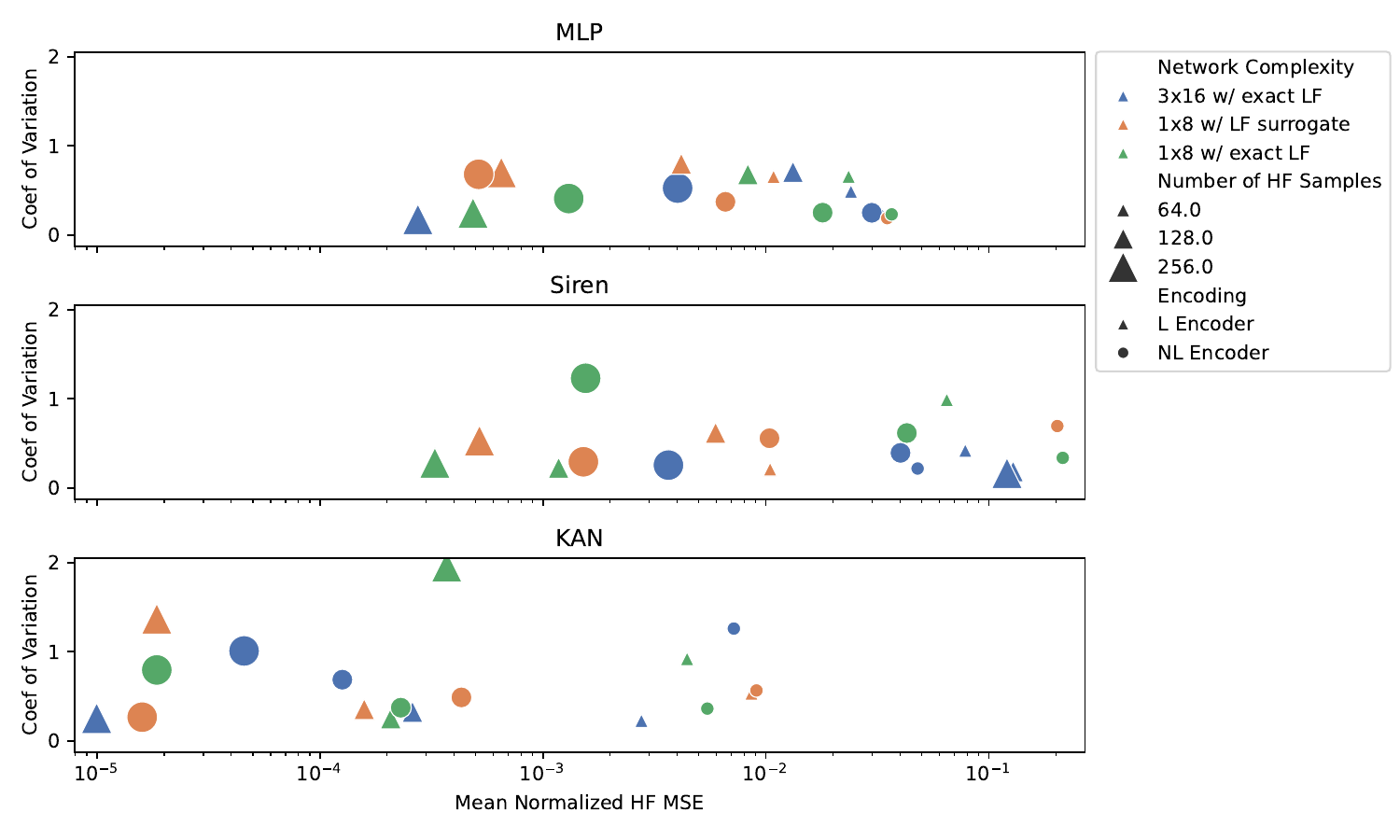}
\caption{Mean normalized HF MSE over 4 repetitions for example 2DU, using various network configurations.}
\label{fig:2d_unequal_scatter_plot}
\end{figure}

Consider the model pair defined by 
\begin{equation}
\begin{aligned}
y_{H}(x, y, z) =& \exp(0.7z + 0.3y) + 0.15 \sin(2 \pi z) + 0.5 x^3, & x,y,z \in [-1.5,1.5],
\\
y_{L}(x, y) =& \exp(0.01\,x + 0.99\,y) + 0.15\,\sin(3 \pi x), & x,y \in [-1.5,1.5].
\end{aligned}
\end{equation}
Note that the dimensionality of the inputs differ between models, with the LF model missing one of the HF parameters. Here, the standard MF architecture cannot deal with this difference in dimensionality, as the architecture necessitates the dimensions of all model inputs to be identical.
However, an encoder can handle the difference in dimensionality.
For this reason, in this section, we only show results for encoded architectures.
As shown in Figure \ref{fig:2d_unequal_scatter_plot}, KAN performs best on average using a linear or nonlinear encoding. Further, KAN combined with a learnable LF surrogate achieves lower MSE than other architectures using an exact LF model, with significantly improved results with respect to MLP and Siren networks.

\begin{figure}[ht!]
\centering
\includegraphics[width=0.32\linewidth]{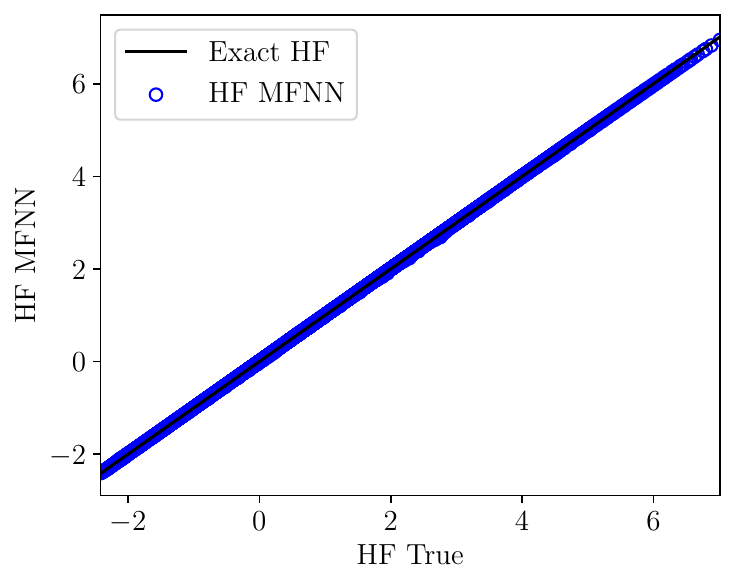}
\caption{Parity plot for best KAN emulator in 2DU test case.}
\label{fig:2d_unequal_fit}
\end{figure}

Figure~\ref{fig:2d_unequal_fit}, shows a parity plot resulting from the best KAN emulator, confirming excellent agreement between predictions and true model outputs.
 
\begin{figure}
    \centering
    \includegraphics[width=0.7\linewidth]{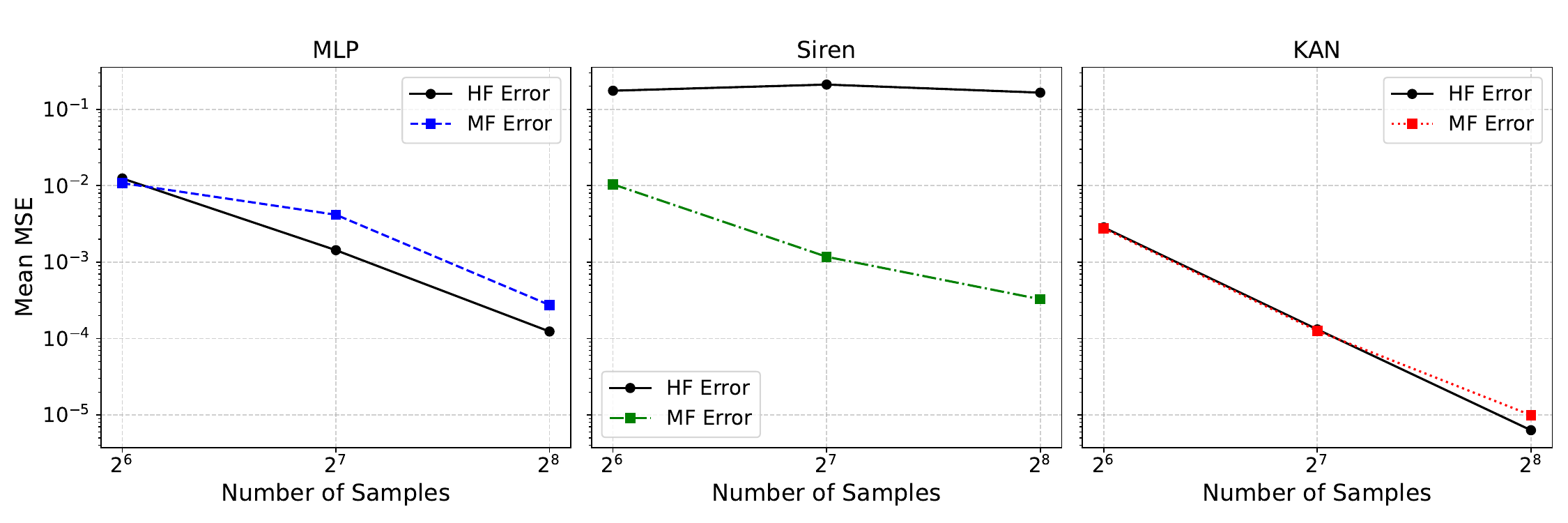}
    \includegraphics[width=0.25\linewidth, height=3.5cm]{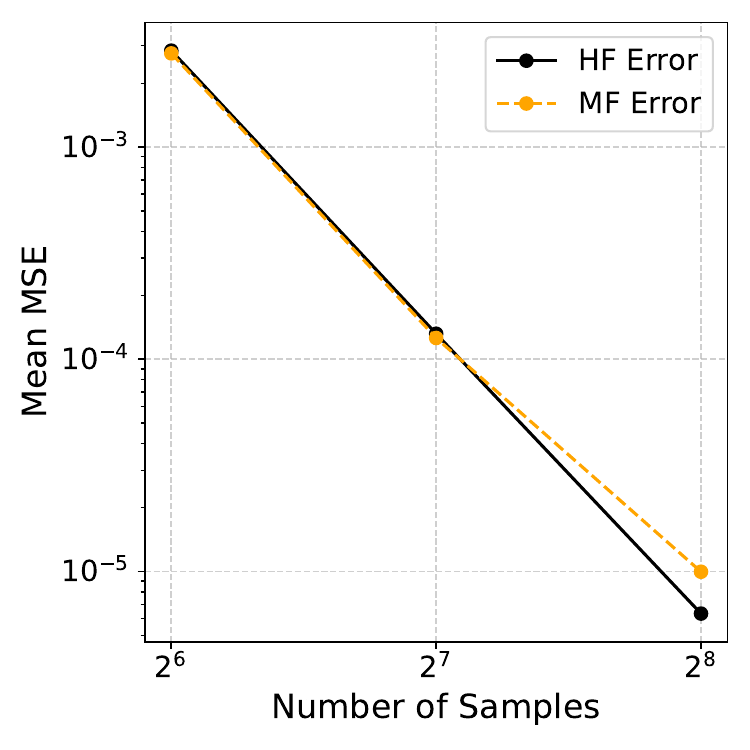}
    \caption{
    Mean MSE results of test case 2DU comparing the best-performing single high-fidelity (HF) network and the best multi-fidelity network: results separated by network architecture across different HF sample sizes (left) and the best results at each HF sample size are shown, regardless of architecture type or network complexity (right).}
    \label{fig:MSE_Lines_2d_unequal}
\end{figure}

Figure~\ref{fig:MSE_Lines_2d_unequal} compares single high-fidelity networks against multi-fidelity networks employing encoders to handle the mismatched input dimensionality between fidelity levels. Notably, the multi-fidelity encoder framework elevates the performance of architectures that typically struggle with single-fidelity learning. While Siren's HF-only performance remains the weakest at approximately $10^{-1}$, its MF approach with encoding achieves significantly better accuracy near $3 \times 10^{-4}$. The MLP and KAN architectures maintain their accuracy at all sampling levels with fully converged HF-only networks. The architecture-agnostic panel reveals closely tracking HF and MF curves across all sample sizes, indicating that the coordinate encoding required for dimensional reconciliation preserves the predictive capacity of the multi-fidelity approach relative to single-fidelity learning, with both pathways benefiting similarly from the encoding structure.

% ===================================================
\subsubsection{Reaction-diffusion PDE (RD)}
% ===================================================

%{\bf\color{red}UPDATE TO MAKE SURE THE SOLUTION IS NOT ZERO!!!}
%{\bf\color{red} Discussion in RD2.tex }

\begin{figure}[ht!]
\centering
\includegraphics[width=\linewidth]{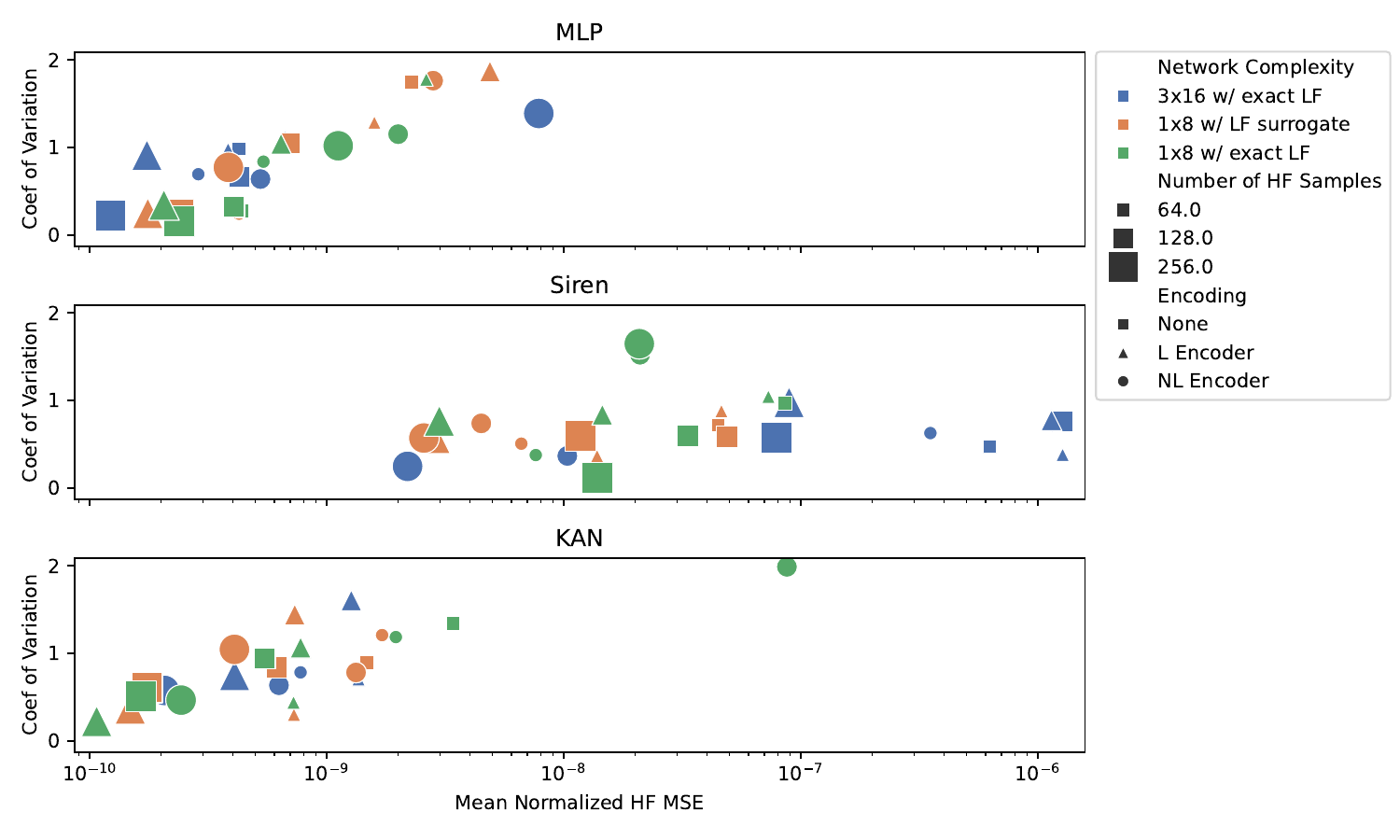}
\caption{Mean normalized HF MSE over 4 repetitions for example RD, using various network configurations.}
\label{fig:rd_scatter_plot}
\end{figure}

\begin{figure}[ht!]
\centering
\includegraphics[trim={0 0 0 2.5em},clip,width=0.32\linewidth]{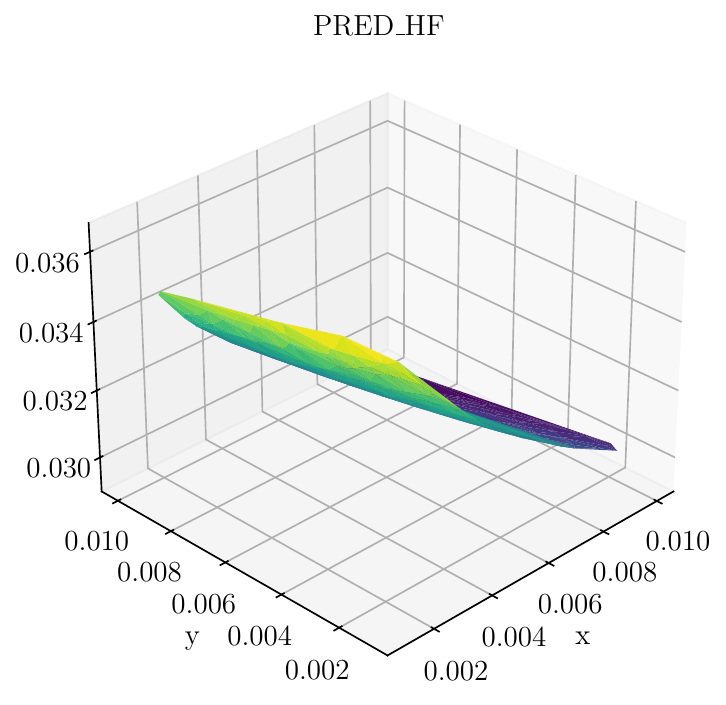}
\includegraphics[trim={0 0 0 0.65em},clip,width=0.32\linewidth]{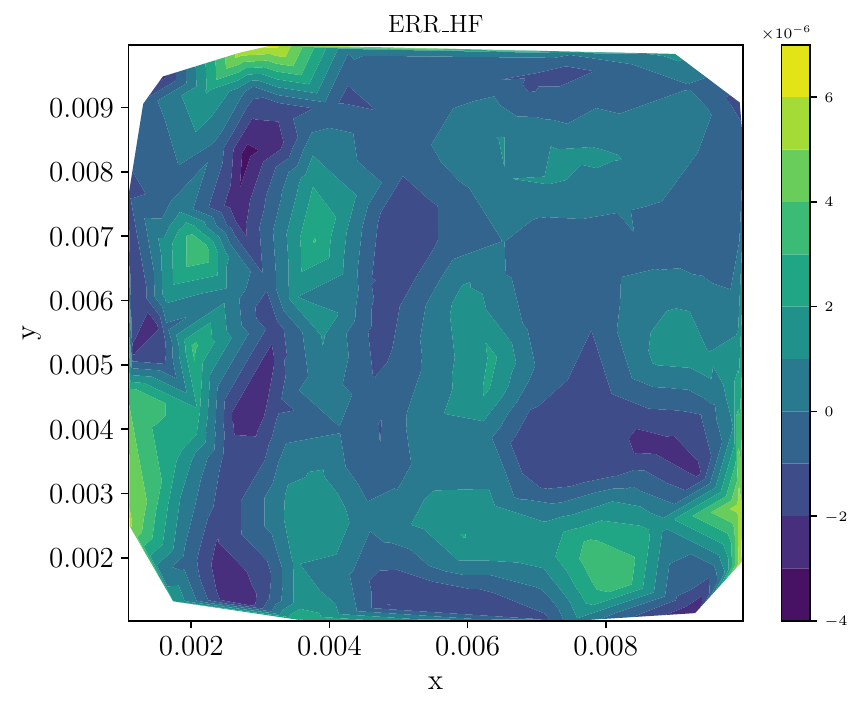}
\caption{Best model fitting results for RD test case obtained from a KAN with linear encoding (left), using 256/2048 HF/LF training samples. The figure also shows a contour of the approximation error (right).}
\label{fig:rd_fit}
\end{figure}

\noindent Consider the 2D reaction-diffusion equation defined in terms of the activator $u=u(t,x_1,x_2)$ and the inhibitor $v=v(t,x_1,x_2)$ on the domain $\Omega = [-1,1]^2$ satisfying a PDE of the form
\begin{equation}
\begin{cases}
    \partial_t u - D_u \Delta u = R_u(u,v), & (x_1,x_2) \in \Omega,
    \\
    \partial_t v - D_v \Delta v = R_v(u,v), & (x_1,x_2) \in \Omega,
    \\
    \bm{\nabla} u = \bm{\nabla} v = \bm{0}, & (x_1,x_2) \in \partial\Omega,
    \\
    u(0,x_1,x_2), v(0,x_1,x_2) \sim N(0,1), & (x_1,x_2) \in \Omega,
\end{cases}
\end{equation}
where $D_u$ and $D_v$ are the corresponding diffusion coefficients.
In particular, our reaction-diffusion equation will take the form of the Fitzhugh-Nagumo equations~\cite{fitzhugh1955mathematical}  with $R_u(u,v) = u - u^3 - k - v$ and $R_v(u,v) = u - v$, with $k = 5 \cdot 10^{-3}$.
Suppose we are interested in calculating a scalar quantity of interest $Q(D_u, D_v) = \int_{0}^{1} \int_{-1}^{0} (u(1,x_1,x_2))\,dx_1\,dx_2$ for $(D_u, D_v) \in [10^{-3}, 10^{-2}]^2$. In other words, $Q$ calculates the mean concentration of our inhibitor $u$ in the region $[0,1] \times [-1,0]$ at $t=1$.
Also suppose we have access to two different numerical approximations of $u$ discretized by time steps $\Delta t = 0.01$ but differing in discretization over the domain $[-1,1]^2$.
Specifically, we have a HF predictor using a uniformly discretized domain in both the $x_1$ and $x_2$ direction of size 128x128 and a down-sampled LF predictor using only a 64x64 spatial mesh.
Here, both approximations are constructed using the PDEBench library~\cite{takamoto2022pdebench}.
This example serves to display how the different network architectures perform when using low-fidelity data from a numerical PDE solution with coarse mesh.
Examining the magnitude of the normalized MSE in Figure \ref{fig:rd_scatter_plot}, we see that every single network achieves a normalized MSE of at least $10^{-5}$ with the best achieving a normalized MSE of $10^{-10}$, and with the Siren network underperforming with respect to MLP and KAN. 
Independent on coordinate encoding, it appears all strategies are equally capable to achieve accurate HF models.
Results from the best KAN network using linear encoding are shown in Figure~\ref{fig:rd_fit}, in addition to the predicted HF model for values over the entire domain $[-1,1]^2$. Further inspecting the error contour plot, we see the testing error on the domain in the order of $10^{-9}$.

\begin{figure}
    \centering
    \includegraphics[width=0.7\linewidth]{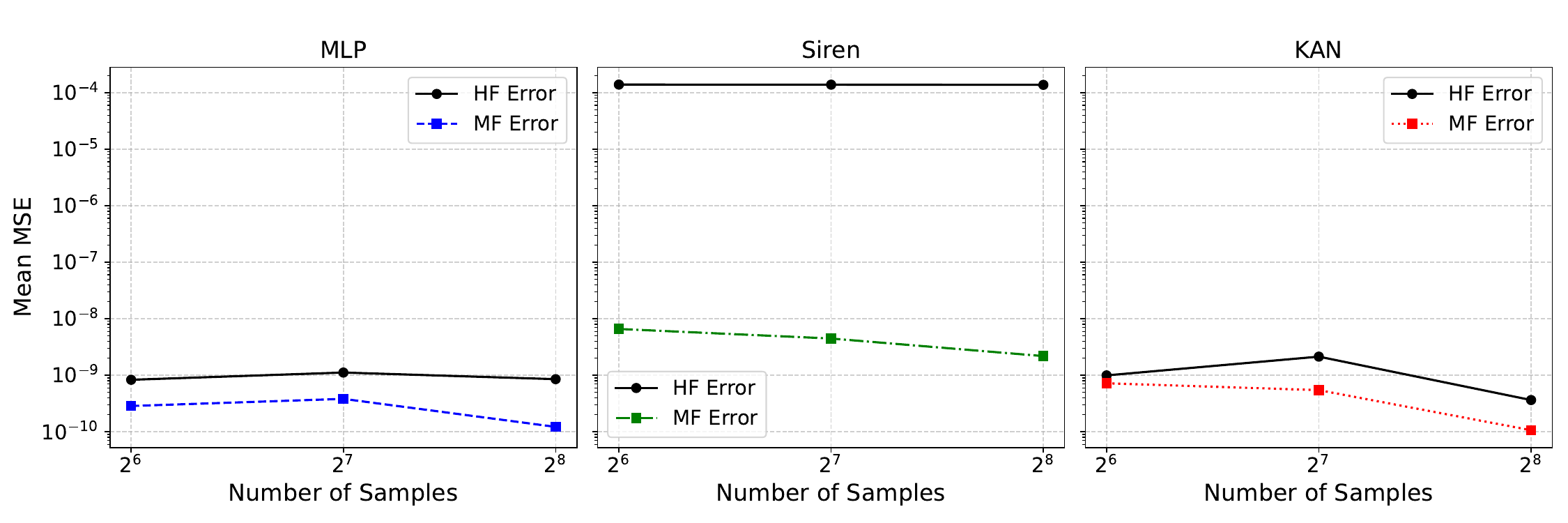}
    \includegraphics[width=0.25\linewidth, height=3.5cm]{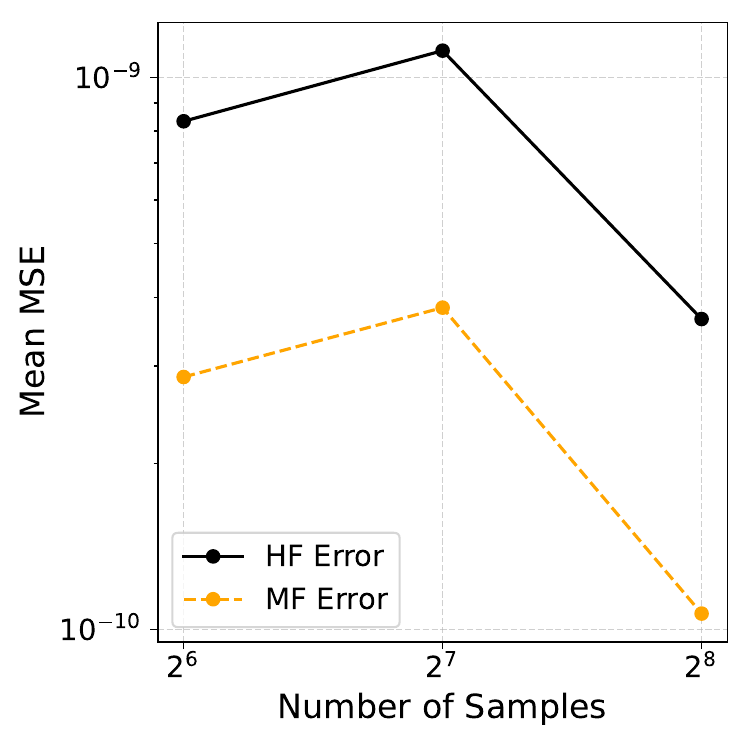}
    \caption{
    Mean MSE results of test case RD comparing the best-performing single high-fidelity (HF) network and the best multi-fidelity network: results separated by network architecture across different HF sample sizes (left) and the best results at each HF sample size are shown, regardless of architecture type or network complexity (right).}
    \label{fig:MSE_Lines_RD}
\end{figure}

Figure~\ref{fig:MSE_Lines_RD} presents results for the RD test case, revealing exceptional performance across all architectures with errors reaching near-machine-precision levels. MLP networks achieve remarkable accuracy with both HF and MF pathways maintaining stable errors near $10^{-9}$ across all sample sizes, with the curves essentially converging and showing no meaningful difference between approaches. Siren networks exhibit their characteristic HF failure mode with stagnant performance near $10^{-4}$, while MF approaches achieve significantly better accuracy around $10^{-8}$, demonstrating multi-fidelity learning remains beneficial for this architecture. KAN architectures display intriguing non-monotonic behavior in both approaches, with HF errors showing an inverted-U pattern peaking at $2^7$ samples before improving at $2^8$, while MF errors exhibit similar dynamics at slightly lower magnitudes. The architecture-agnostic panel reveals this non-monotonic pattern dominates the overall behavior, with both HF and MF curves tracking closely and showing performance degradation at intermediate sample counts before recovering at maximum sampling, suggesting the optimization landscape for this problem contains challenging regions at specific sample sizes that affect all high-capacity architectures.

% ============================================================
\subsection{Test cases with more than two low-fidelity models}
% ============================================================

% =================================================================
\subsubsection{Multiple noisy low-fidelity models (GJG9)}
% =================================================================

For $i_1, i_2 \in \{0,1\}$, consider the function~\cite{Gorodetsky2021}
\begin{equation}
    f_{i_1, i_2}(x_1, x_2)=(2 + 0.5 x_1 + 0.5 x_2 + 3 x_1 x_2) + i_1 \left( 2 x_1^5 + 2 x_2^5 \right) + i_2 \left( x_1^2 + x_2^2 + 5 x_1^2 x_2^2 \right),\,\,x_1, x_2 \in [-1.1]^2
\end{equation}
and the model array
\begin{equation}
\resizebox{0.94\linewidth}{!}{$
\begin{aligned}
    y_{L_1}(x) &=
    f_{0,0}(x) \cdot \mathcal{N}\left( 1, \frac{1}{5} \right),
    \quad&\quad
    y_{L_2}(x) &=
    f_{0,0}(x) \cdot \mathcal{N}\left( 1, \frac{1}{10} \right),
    \quad&\quad
    y_{L_3}(x) &=
    f_{0,0}(x) \cdot \mathcal{N}\left( 1, \frac{1}{100} \right),
    \\
    y_{L_4}(x) &=
    f_{0,1}(x) \cdot \mathcal{N}\left( 1, \frac{1}{5} \right),
    \quad&\quad
    y_{L_5}(x) &=
    f_{0,1}(x) \cdot \mathcal{N}\left( 1, \frac{1}{10} \right),
    \quad&\quad
    y_{L_6}(x) &=
    f_{0,1}(x) \cdot \mathcal{N}\left( 1, \frac{1}{100} \right),
    \\
    y_{L_7}(x) &=
    f_{1,1}(x) \cdot \mathcal{N}\left( 1, \frac{1}{5} \right),
    \quad&\quad
    y_{L_8}(x) &=
    f_{1,1}(x) \cdot \mathcal{N}\left( 1, \frac{1}{10} \right),
    \quad&\quad
    y_{H}(x) &=
    f_{1,1}(x) \cdot \mathcal{N}\left( 1, \frac{1}{100} \right).
\end{aligned}
$}
\end{equation}
where $\mathcal{N}\left( \mu, \sigma^2 \right)$ is single instance of a Gaussian random variable with mean $\mu$ and variance $\sigma^2$.
In this test case, we examine how different neural network architectures handle information fusion from multiple low-fidelity sources with varying degrees of accuracy and noise. The model array consists of nine functions (eight low-fidelity and one high-fidelity) that systematically vary across two key dimensions: (1) model complexity, represented by the inclusion of higher-order polynomial terms through the parameters $i_1$ and $i_2$, and (2) noise level, implemented through different variance values in the multiplicative Gaussian noise.
The high-fidelity function contains all higher-order terms and has the lowest noise level, while the low-fidelity functions represent various compromises between model complexity and noise.

\begin{figure}[ht!]
\centering
\includegraphics[width=\linewidth]{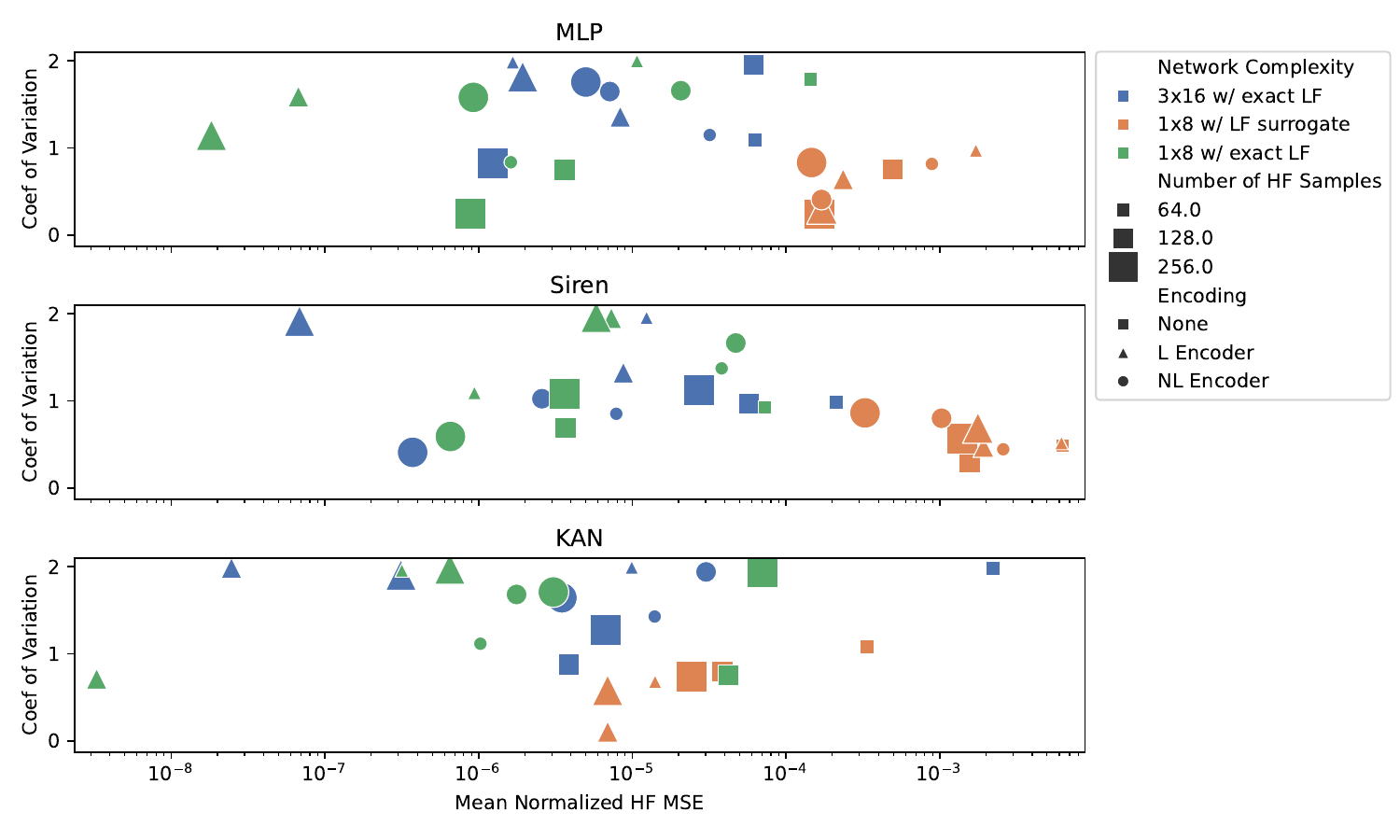}
\caption{Mean normalized HF MSE over 4 repetitions for example GJG9, using various network configurations.}
\label{fig:9_model_scatter_plot}
\end{figure}

\begin{figure}[ht!]
\centering
\includegraphics[trim={0 0 0 1.25cm},clip,width=0.32\linewidth]{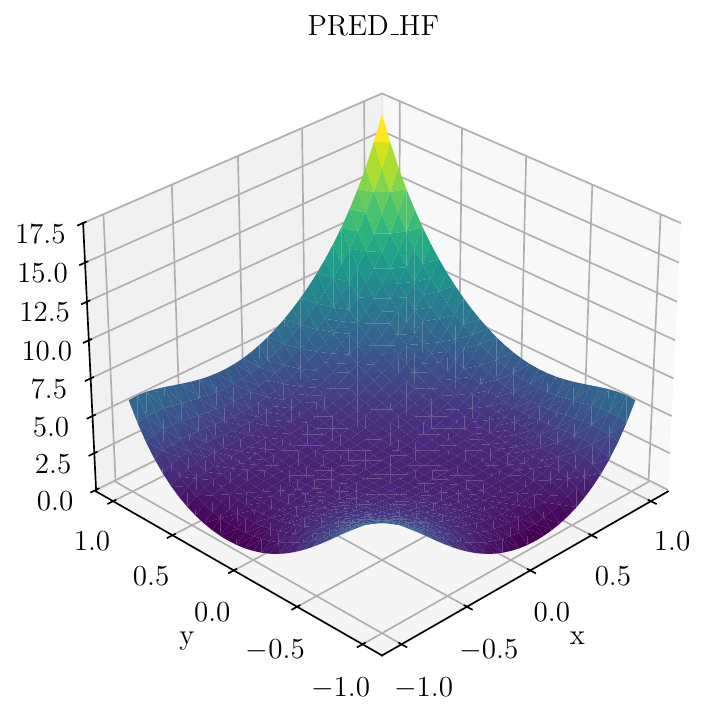}
\includegraphics[trim={0 0 0 1.25cm},clip,width=0.32\linewidth]{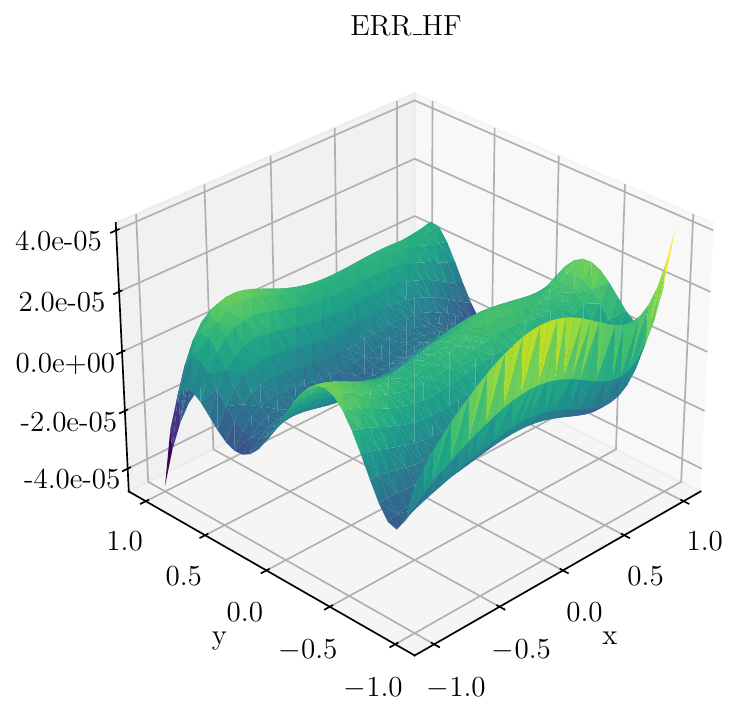}
\caption{Best surface fit and corresponding error plot result for test case GJG9 from a KAN architecture with linear encoding and 128 HF samples.}
\label{fig:9_model_l_en_fit}
\end{figure}

The results in Figure~\ref{fig:9_model_scatter_plot} show that linear encoding significantly outperforms both nonlinear and no encoding approaches across all network architectures, achieving improvements by orders of magnitude.
In addition, the results show that most network configurations are able to generate accurate emulators, and that the best results are not necessarily achieved using all 256 samples.
As an example, Figure~\ref{fig:9_model_l_en_fit} shows the best fit results achieved by the 2nd realization of a KAN network with linear encoding trained on 128 HF samples.

Figure~\ref{fig:9_model_l_en_fit} shows the predicted surface and corresponding error distribution for the best-performing KAN network with linear encoding. The smooth predicted show an accurate emulation of the high-fidelity function with error remaining below $10^{-5}$ across most of the domain

\begin{figure}
    \centering
    \includegraphics[width=0.7\linewidth]{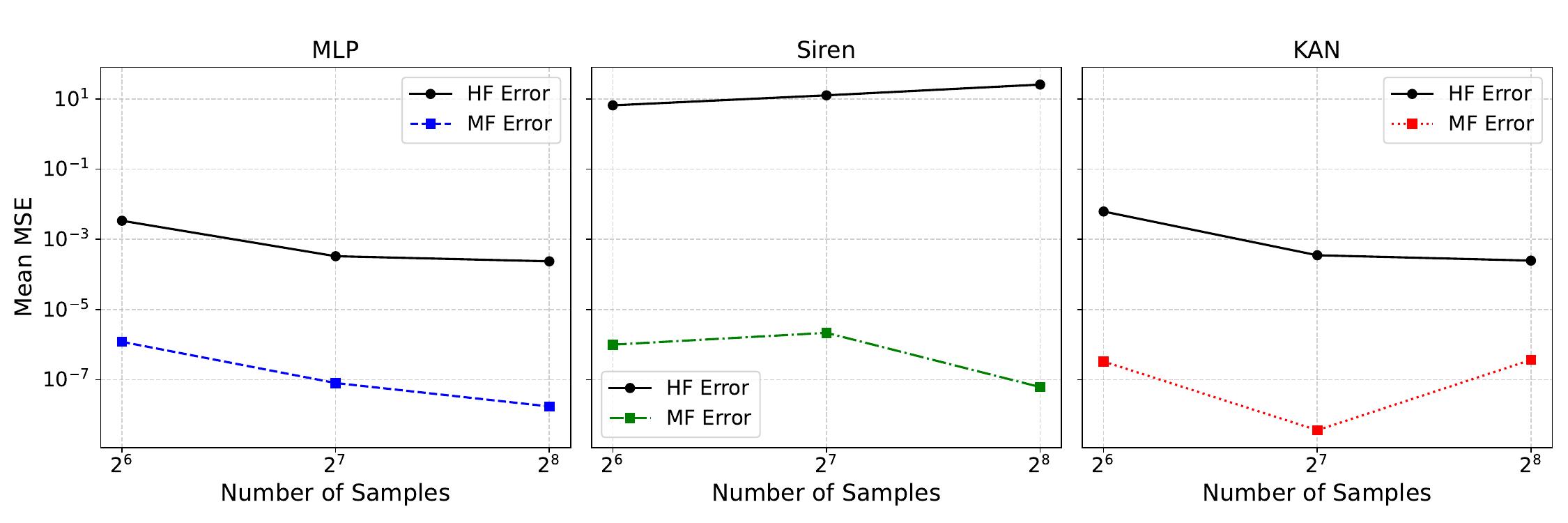}
    \includegraphics[width=0.25\linewidth, height=3.5cm]{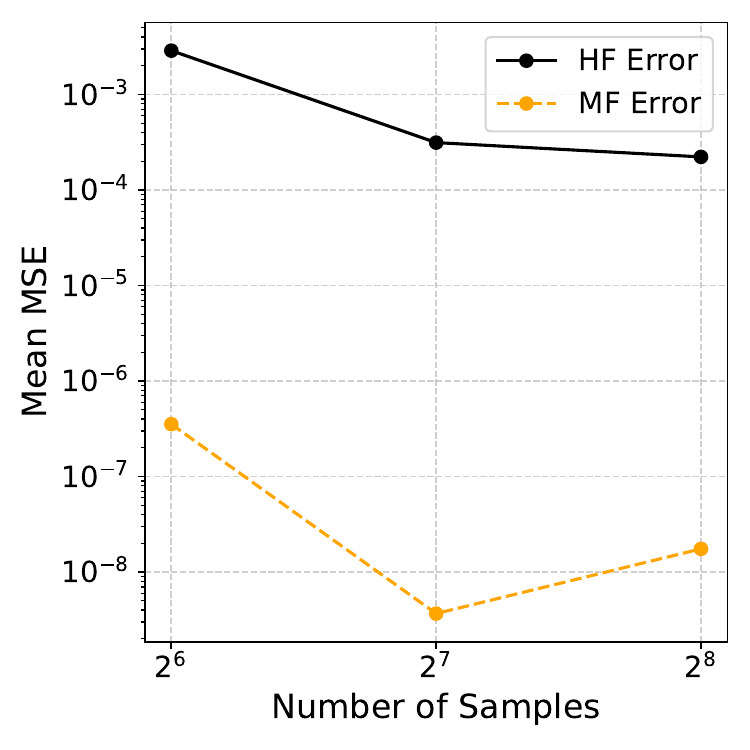}
    \caption{
    Mean MSE results of test case GJG9 comparing the best-performing single high-fidelity (HF) network and the best multi-fidelity network: results separated by network architecture across different HF sample sizes (left) and the best results at each HF sample size are shown, regardless of architecture type or network complexity (right).}
    \label{fig:MSE_Lines_gjg}
\end{figure}

Figure~\ref{fig:MSE_Lines_gjg} examines convergence behavior across architectures. The results reveal that this information fusion task fundamentally alters learning dynamics compared to single low-fidelity cases. MLP networks demonstrate the most dramatic multi-fidelity advantage, with HF-only errors plateauing near $10^{-3}$ while MF approaches leverage the diverse low-fidelity ensemble to achieve errors near $10^{-8}$. Siren networks fail catastrophically with HF-only training but recover to functional performance through multi-fidelity learning. KAN architectures show modest HF improvement but exhibit non-monotonic MF behavior, with performance peaking at $2^7$ samples near $2 \times 10^{-8}$ before degrading slightly at $2^8$. This non-monotonic pattern, observed across the architecture-agnostic panel as well, suggests that the complexity of fusing information from multiple heterogeneous sources may lead to overfitting or optimization challenges at higher sample counts, indicating an optimal training regime exists for this problem class.

% ========================================
\subsection{Higher-dimensional test cases}
% ========================================

% ==============================================================================
\subsubsection{High dimensional nonlinearly correlated models (K5)}
% ==============================================================================

Consider the model pair
\begin{comment}
    % Example 5 as written in Meng, Karniadakis paper (K5)
    \begin{equation}
    \begin{aligned}
        y_L(x) =&
        0.8 y_H(x) - \sum_{i=1}^{19} 0.4 x_i x_{i+1} - 50,
        \\
        y_H(x) =&
        (x_1 - 1)^2 + \sum_{i=2}^{20} ( 2x_i - x_{i-1} )^2,
        x_i \in [-3,3], i=1, \cdots, 20
    \end{aligned}
    \end{equation}
\end{comment}
% Re-arranged K5 example
\begin{equation}
\label{eq:K5}
\begin{aligned}
    y_L(x) =&
    0.8\,(x_1 - 1)^2 + 0.4 \sum_{i=1}^{19} (2x_{i+1} - x_{i})(x_{i+1} - 2x_{i}) - 50, & x_1, \cdots, x_{20} \in [-3,3]
    \\
    y_H(x) =&
    1.25\,y_L(x) + \sum_{i=1}^{19} 0.5\,x_i\,x_{i+1} + 62.5, & x_1, \cdots, x_{20} \in [-3,3]
    .
\end{aligned}
\end{equation}

\begin{figure}[ht!]
\centering
\includegraphics[width=\linewidth]{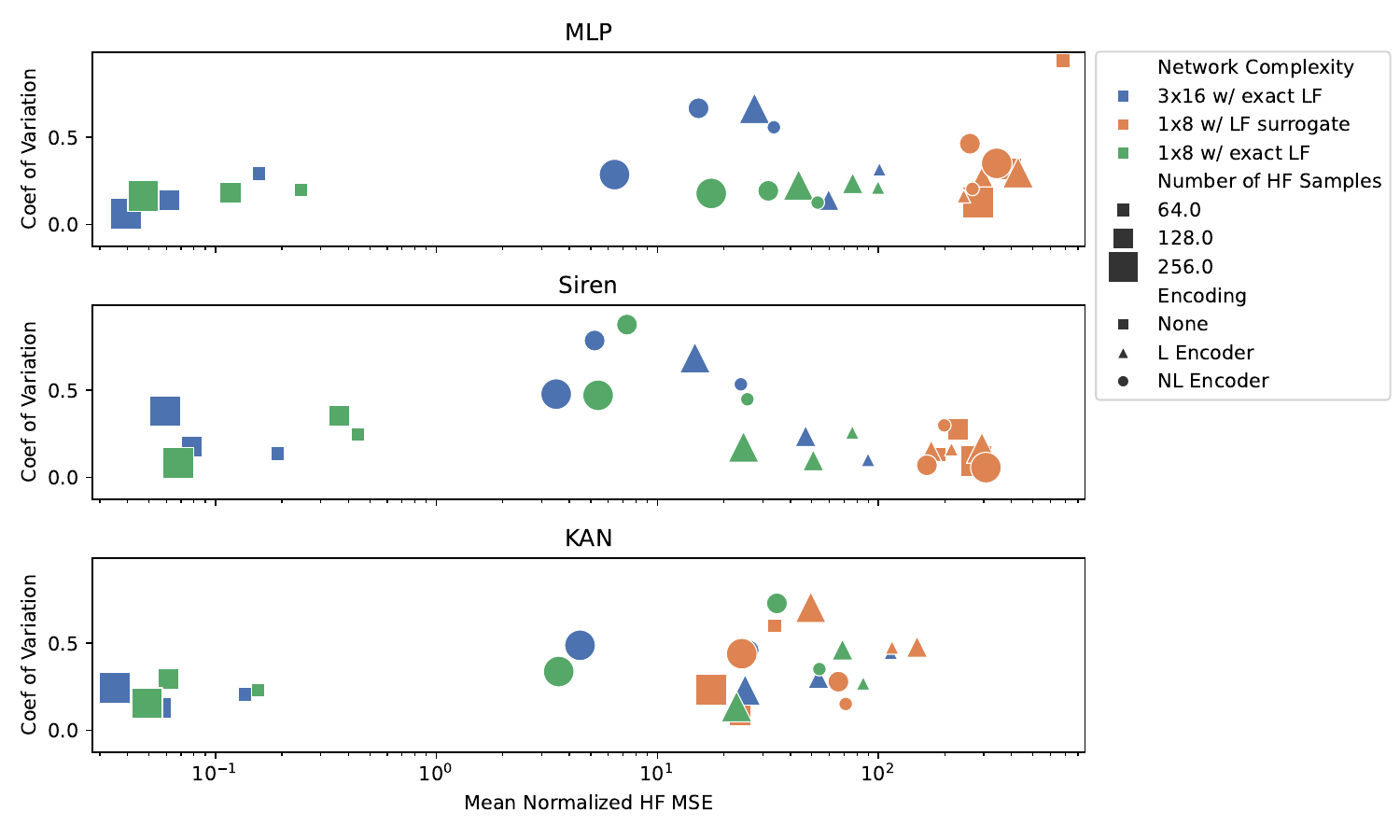}
\caption{Mean normalized HF MSEs of 4 repetitions for example K5 using various network configurations.}
\label{fig:k5_scatter_plot}
\end{figure}

This a high dimensional example, so even while leveraging a MF framework with a LF function that captures some portion of the HF response, without both sufficient HF data and a network with sufficient complexity, we expect the HF predictions to be poor.
% Has been shown to work for Karniadakis paper
In this example, we test how the network architectures deal with high dimensional models and whether any particular choice in architecture or inclusion of encoding block proves to be more effective in producing HF surrogates.
The results in Figure \ref{fig:k5_scatter_plot} show all the normalized MSEs for this example are relatively high compared to the normalized MSEs found in the other test cases, regardless of network complexity, encoding, architecture, and sample size.
Moreover, we see that the networks that perform best are the ones not using coordinate encoding, with nonlinear encoding and linear encoding following in that order.
For the network configurations using an exact LF, we see a consistent trend where the mean normalized MSEs reduces as the number of samples increases.
However, we do see very little change in accuracy as the number of HF samples increase. 
This may suggest the number of HF samples required to provide sufficient generalization has not been reached. For the networks using a LF surrogate, the trend is less clear as the MSEs are higher than when using an exact LF function and cluster together, regardless of the number of HF samples.

\begin{figure}[ht!]
\centering
\includegraphics[width=0.32\linewidth]{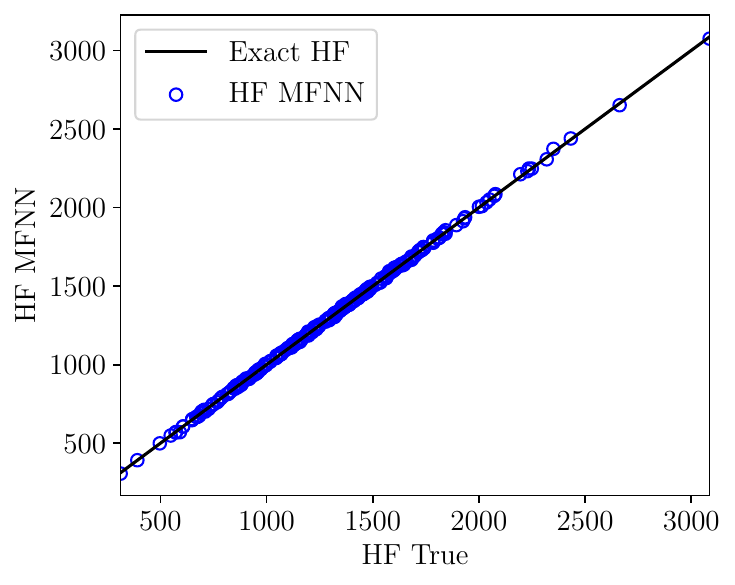}
\includegraphics[width=0.32\linewidth]{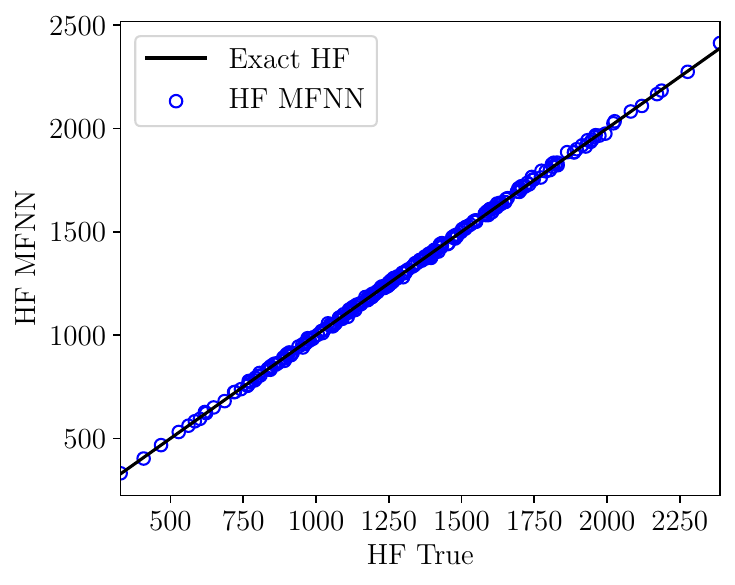}
\includegraphics[width=0.32\linewidth]{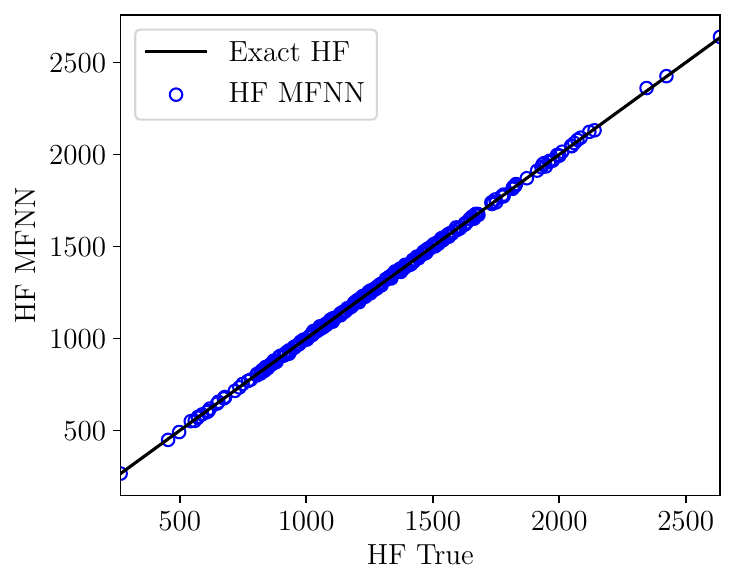}
\caption{512 random HF testing samples plotted against HF fit for MLP network using an exact LF model with no encoding (left), linear encoding (middle), and nonlinear encoding (right).}
\label{fig:k5_MLP_fit}
\end{figure}

Inspecting the model predictions of the best performing MLP networks using 256 HF samples and an exact LF for all network architecture in Figure~\ref{fig:k5_MLP_fit}, we see very comparable performances in the parity plots across all architectures.

\begin{figure}
    \centering
    \includegraphics[width=0.7\linewidth]{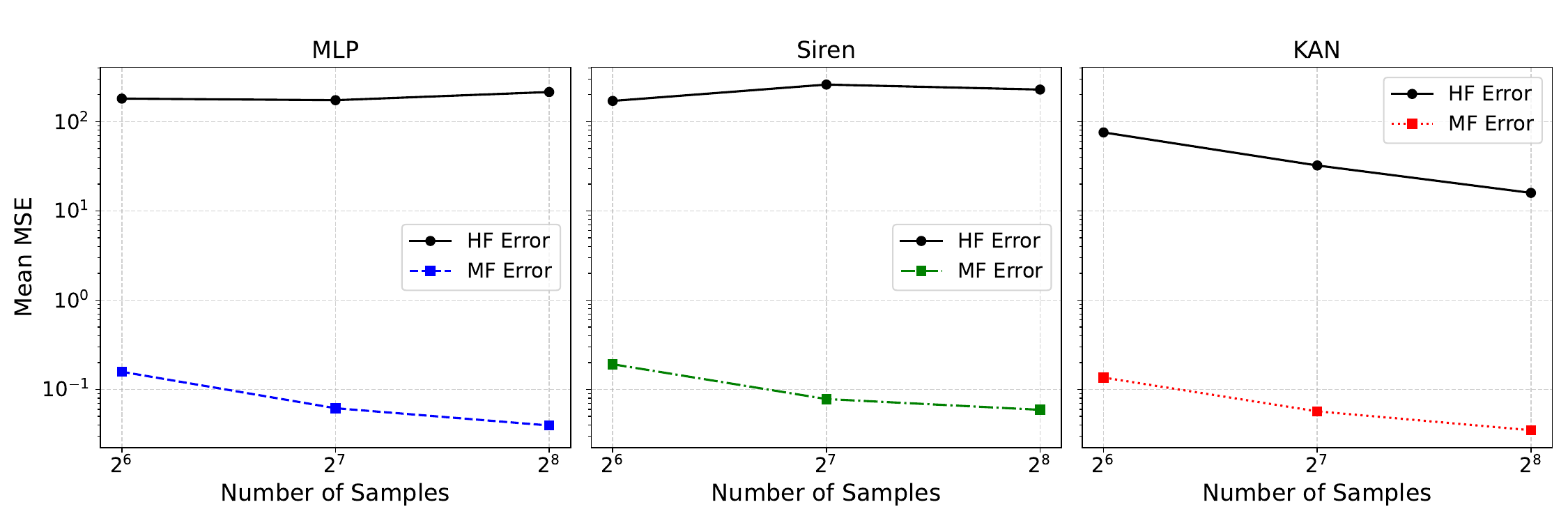}
    \includegraphics[width=0.25\linewidth, height=3.5cm]{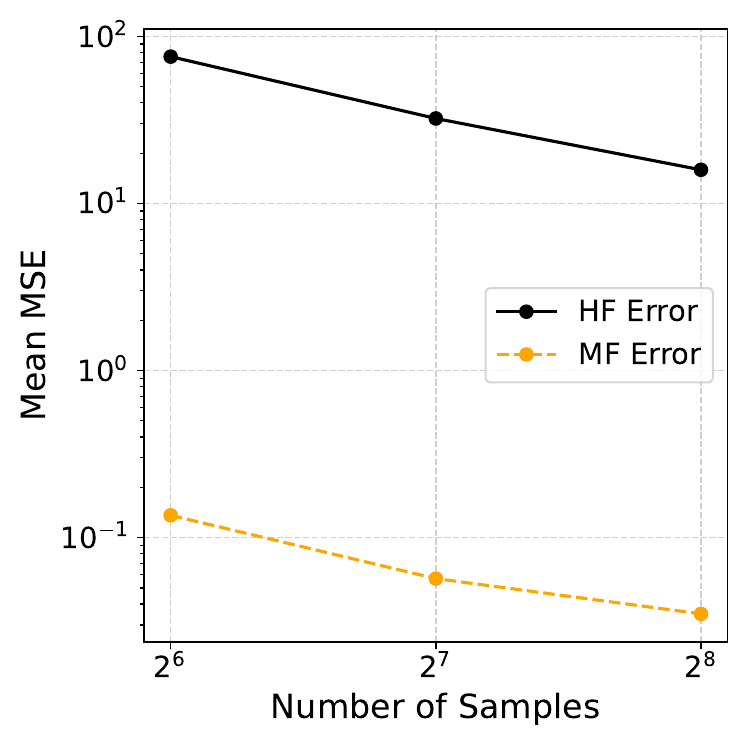}
    \caption{
    Mean MSE results of test case K5 comparing the best-performing single high-fidelity (HF) network and the best multi-fidelity network: results separated by network architecture across different HF sample sizes (left) and the best results at each HF sample size are shown, regardless of architecture type or network complexity (right).}
    \label{fig:MSE_Lines_k5}
\end{figure}

Examining Figure~\ref{fig:MSE_Lines_k5}, we see that multi-fidelity networks work better than single fidelity networks by over 2 order of magnitude in every cases and each multi-fidelity network improves in accuracy from increased HF samples. 
This is in contrast of single fidelity MLPs and Sirens as they show stagnant performance of mean MSE results above $10^2$ as the number of HF samples grows. Single fidelity KAN architectures, however, show fundamentally different scaling behavior where, like the multi-fidelity networks, they also improve with the number of available HF samples. This convergent trend suggests that KANs possess the architectural capacity to assimilate high-fidelity information more effectively, though the persistent gap between HF and MF errors across all sample sizes indicates that multi-fidelity learning still provides consistent value for high-dimensional data. 

% ========================================
\section{Discussion}\label{sec:Discussion}
% ========================================

The numerical experiments presented in Section~\ref{sec:results} provide several key insights into the performance of various neural network architectures in multi-fidelity modeling. KAN networks consistently outperform other architectures across most test cases, especially in problems involving complex correlations or high dimensionality. This superior performance is attributed to their ability to efficiently model complex functions as compositions of univariate functions. While MLPs perform competitively in simpler problems and demonstrate good robustness across different scenarios, they typically require more samples to achieve accuracy comparable to that of KAN networks. 
% Siren consistently underperforming
Siren networks, though effective for certain specialized applications, generally underperform relative to KAN and MLP architectures in multi-fidelity settings. Their periodic activations promote overfitting that typically results in limited accuracy for training sets with relatively small size, as those commonly encountered in multi-fidelity problems.
So in the spectral domain, the three architectures exhibit complementary strengths: KAN demonstrates superior performance across all frequency ranges when training data is limited, MLP architectures show preferential accuracy in capturing low-frequency components, while Siren networks, leveraging their periodic activation functions, prove particularly effective at representing high-frequency oscillatory behavior.

% Mention encoding may only help when correlations are complex enough
The effectiveness of coordinate encoding varies considerably depending on the characteristics of the test case. For cases with dominant linear correlations, such as test case K1, coordinate encoding provides little benefit and may harm performance by introducing unnecessary complexity to the models. In these scenarios, standard architectures without encoding perform well. However, both linear and nonlinear encoding show substantial advantages in problems involving phase shifts (test case K4) or misaligned parameterizations (test case 2DU), where the encoding networks successfully learn coordinate transformations that align LF data with HF features.

% Benefit for encoders to capture shocks
Additionally, in problems with discontinuities, as demonstrated in test cases K2 and K2 shift, the choice of encoding becomes more critical. When discontinuities in the models are co-located, simpler architectures without encoding perform best. On the other hand, when discontinuities are offset, nonlinear encoding provides significant benefits by learning the appropriate domain transformation. This emphasizes the importance of selecting the network architecture that matches the specific characteristics of the problem.

The experiments with multiple low-fidelity sources (test cases K4D, K4U, K4P, and GJG9) offer valuable insights into how different architectures handle information fusion. All networks tested were able to sensibly process multiple information sources. Specifically, they demonstrated the ability to leverage redundant information constructively, ignore irrelevant data, select appropriate sources for different regions, and effectively combine multiple noisy sources.

The K5 test case reveals an important distinction between coordinate encoding and multi-fidelity learning itself. While multi-fidelity approaches consistently deliver substantial improvements over single high-fidelity training, coordinate encoding provides minimal advantage for this problem. This contrasts with cases like K4 and 2DU where encoding proves essential for capturing phase shifts or parameterization misalignments. The K5 results indicate that when fidelity levels share consistent parameterizations but the function is intrinsically difficult to learn, standard multi-fidelity architectures without encoding suffice to achieve exceptional accuracy.

% ==================================
\section{Conclusion and Future Work}\label{sec:Conclusion and Future Work}
% ==================================

In this comprehensive study, we evaluate the performance of different neural network architectures (MLP, Siren, and KAN) with and without the addition of coordinate encoding in a multi-fidelity setting across a diverse range of test cases.
The results show that while each architecture has its strengths, KAN networks generally deliver superior performance, particularly for complex problems involving nonlinear correlations or high dimensionality, while MLPs remain a robust option good for most test cases. Additionally, coordinate encoding proved valuable in scenarios involving phase shifts between low- and high-fidelity models, misaligned parameterizations, multiple potentially noisy low-fidelity sources, and discontinuities.
However, the benefits of encoding must be weighed against the increased computational complexity and potential optimization challenges, especially in simpler problems where standard architectures perform sufficiently well.

Future research should focus on several key areas. These may include developing adaptive architecture selection methods, improve computational efficiency, particularly by optimizing the training process for large-scale problems, and developing more effective strategies for hyperparameter optimization.
These advancements would significantly enhance the practical utility of multi-fidelity neural emulators in scientific computing and engineering design applications.

% =========================
\section*{Acknowledgments}
% =========================

\noindent This work is supported by a NSF CAREER award \#1942662 (PI DES), a NSF CDS\&E award \#2104831 (University of Notre Dame PI DES) and used computational resources provided through the Center for Research Computing at the University of Notre Dame.  
CV also acknowledges the support from the Department of Energy through the Computational Science Graduate Fellowship (DOE CSGF).
Sandia National Laboratories is a multi-mission laboratory managed and operated by National Technology \& Engineering Solutions of Sandia, LLC (NTESS), a wholly owned subsidiary of Honeywell International Inc., for the U.S. Department of Energy’s National Nuclear Security Administration (DOE/NNSA) under contract DE-NA0003525. This written work is authored by an employee of NTESS. The employee, not NTESS, owns the right, title and interest in and to the written work and is responsible for its contents. Any subjective views or opinions that might be expressed in the written work do not necessarily represent the views of the U.S. Government. The publisher acknowledges that the U.S. Government retains a non-exclusive, paid-up, irrevocable, world-wide license to publish or reproduce the published form of this written work or allow others to do so, for U.S. Government purposes. The DOE will provide public access to results of federally sponsored research in accordance with the DOE Public Access Plan.

% Bibliography
\printbibliography

% Leaving this here in case we add more section like Loss Table with printed loss statistics
\appendix

% Discussion of effect of normalization on correlation
% Important for sensitivity analysis calculations
\section{Encoded predictions on normalized data}

In this section, we discuss how data normalization interacts with coordinate encoding and affect model correlation. Suppose we are given a multi-fidelity
dataset in the form of a high-fidelity dataset $\{\bm{x}_{H}^{(j)}, y_{H}^{(j)} \}_{j = 1}^{N_H} \subset \boldsymbol{\mathcal{X}}_{H} \times \boldsymbol{\mathcal{Y}}$ and $n$ sets of low-fidelity data $\{ \bm{x}_{L_{i}}^{(j)}, y_{L_{i}}^{(j)}\}_{j = 1}^{N_{L_i}} \subset \boldsymbol{\mathcal{X}}_{L_{i}} \times \boldsymbol{\mathcal{Y}}$ for $i=1,\ldots,n$.
Also suppose we wish to normalize (or standardize) our data before
training with invertible normalizing functions
\begin{equation}
\begin{aligned}
    \widetilde{x}_{L_i} = W_{ \boldsymbol{\mathcal{X}}_{L_i} } (x_{L_i}),
    \quad
    \widetilde{x}_{H} = W_{\boldsymbol{\mathcal{X}}_{H}}(x_{H}),
    \quad
    \widetilde{y}_{L_i} = W_{ \boldsymbol{\mathcal{Y}}_{L_i} } (y_{L_i}),
    \quad
    \widetilde{y}_{H} = W_{\boldsymbol{\mathcal{Y}}_{H}}(y_{H})
    .
\end{aligned}
\end{equation}
Also let $T_{i}: \boldsymbol{\mathcal{X}}_{H} \to \boldsymbol{\mathcal{X}}_{L_{i}}$ be a coordinate encoder on unnormalized data with $\widetilde{T}_{i} : W_{\boldsymbol{\mathcal{X}}_{H}}\left( \boldsymbol{\mathcal{X}}_{H} \right) \to W_{ \boldsymbol{\mathcal{X}}_{L_i} } \left( \boldsymbol{\mathcal{X}}_{L_{i}} \right)$ be the corresponding encoder on normalized data.
It follows $\widetilde{T}_{i} = W_{ \boldsymbol{\mathcal{X}}_{L_i} } \circ T_{i} \circ W_{\boldsymbol{\mathcal{X}}_{H}}^{-1}$. For the same reason, the normalized predictions satisfy $\widetilde{y}_{H} (\widetilde{x}_{H}) = (W_{\boldsymbol{\mathcal{Y}}_{H}} \circ \widehat{y}_{H}) (x_{H})$ and $\widetilde{y}_{L_i} (\widetilde{x}_{L_i}) = (W_{\boldsymbol{\mathcal{Y}}_{L_i}} \circ \widehat{y}_{L_i}) (x_{L_i})$ and $(\widetilde{y}_{L_i} \circ \widetilde{T}_i) (\widetilde{x}_{H}) = (W_{\boldsymbol{\mathcal{Y}}_{L_i}} \circ \widehat{y}_{L_i} \circ T_{i}) (x_{H})$.
Now consider the general form of the normalized HF correlation function
\begin{equation}
\begin{aligned}
    \widetilde{y}_{H} (\widetilde{x}_{H})
    =&
    \sum_{i = 1}^{n}
    \left(
    \widetilde{A}_{i} (\widetilde{y}_{L_{i}} \circ \widetilde{T}_{i}) (\widetilde{x}_{H})
    \right)
    +
    \widetilde{B} \widetilde{x}_{H}
    +
    \widetilde{C}
    +
    \widetilde{\Delta}(\widetilde{x}_{H}, \widetilde{y}_{L_{1}} \circ \widetilde{T}_{1}, \cdots, \widetilde{y}_{L_{n}} \circ \widetilde{T}_{n}).
\end{aligned}
\end{equation}
Here, $\widetilde{A}_{i}$, $\widetilde{B}$,$\widetilde{C}$ and $\widetilde{\Delta}$ 
are the contributions of the normalized low-fidelity models, the direct linear normalized HF input term, a constant offset, and a nonlinear residual term, respectively.
Without assumptions on the forms of the normalization, it is difficult to express the original coefficients directly.
However, if the transformation is affine such that $W_{k}(z) = \alpha_{k}^{-1} (z - \beta_{k})$ and $W_{k}^{-1}(z) = \alpha_{k} z + \beta_{k}$ for all $k$, then the original, unnormalized coefficients can be explicitly recovered from the normalized coefficients:
\begin{enumerate}
    \item 
    $A_i = \alpha_{\boldsymbol{\mathcal{Y}}_{H}} \widetilde{A}_{i} \alpha_{ \boldsymbol{\mathcal{Y}}_{L_i} }^{-1}$
    \item 
    $B = \alpha_{\boldsymbol{\mathcal{Y}}_{H}} \widetilde{B} \alpha_{\boldsymbol{\mathcal{X}}_{H}}^{-1}$
    \item 
    $C = \alpha_{\boldsymbol{\mathcal{Y}}_{H}}
    \widetilde{C}
    -
    \left(
    \sum_{i = 1}^{n}
    \alpha_{\boldsymbol{\mathcal{Y}}_{H}}
    \widetilde{A}_{i} \alpha_{\boldsymbol{\mathcal{Y}}_{L_i}}^{-1} \beta_{\boldsymbol{\mathcal{Y}}_{L_i}}
    \right)
    -
    \alpha_{\boldsymbol{\mathcal{Y}}_{H}}
    \widetilde{B} \alpha_{\boldsymbol{\mathcal{X}}_{H}}^{-1} \beta_{\boldsymbol{\mathcal{X}}_{H}}
    +
    \beta_{\boldsymbol{\mathcal{Y}}_{H}}$
    \item 
    $\Delta = \alpha_{\boldsymbol{\mathcal{Y}}_{H}}
    \widetilde{\Delta}(W_{\boldsymbol{\mathcal{X}}_{H}} (x_{H}), (W_{\boldsymbol{\mathcal{Y}}_{L_1}} \circ \widehat{y}_{L_1} \circ T_{1}) (x_{L_1}), \cdots, (W_{\boldsymbol{\mathcal{Y}}_{L_n}} \circ \widehat{y}_{L_n} \circ T_{n})(x_{L_n}))$
\end{enumerate}
This allows us to interpret the model in the original data units while still training on normalized inputs and outputs.

\begin{comment}
For general invertible normalizing functions, we can express the unnormalized predictions as
\begin{equation}
\begin{aligned}
    \widehat{y}_{H} (x_{H})
    =&
    W_{\boldsymbol{\mathcal{Y}}_{H}}^{-1}
    \left(
    \sum_{i = 1}^{n}
    \left(
    \widetilde{A}_{i} (W_{\boldsymbol{\mathcal{Y}}_{L_i}} \circ \widehat{y}_{L_i} \circ T_{i}) (x_{H})
    \right)
    +
    \widetilde{B} W_{\boldsymbol{\mathcal{X}}_{H}} (x_{H})
    +
    \widetilde{C}
    +
    \Delta_1
    \right)
\end{aligned}
\end{equation}
where $\Delta_1 = \widetilde{\Delta}(W_{\boldsymbol{\mathcal{X}}_{H}} (x_{H}), (W_{\boldsymbol{\mathcal{Y}}_{L_1}} \circ \widehat{y}_{L_1} \circ T_{1}) (x_{L_1}), \cdots, (W_{\boldsymbol{\mathcal{Y}}_{L_n}} \circ \widehat{y}_{L_n} \circ T_{n})(x_{L_n}))$.
\end{comment}

\end{document}